\documentclass{article}

\PassOptionsToPackage{numbers, compress}{natbib}  %

\PassOptionsToPackage{numbers, compress}{natbib}
\usepackage[preprint]{neurips_2026}

\usepackage[utf8]{inputenc} %
\usepackage[T1]{fontenc}    %
\usepackage{hyperref}       %
\usepackage{url}            %
\usepackage{booktabs}       %
\usepackage{amsfonts}       %
\usepackage{nicefrac}       %
\usepackage{microtype}      %
\usepackage{xcolor}         %
\usepackage{amsmath}
\usepackage{graphicx}
\usepackage[utf8]{inputenc}

\usepackage{booktabs}
\usepackage{tabularx}
\usepackage{tcolorbox}
\usepackage{enumitem}
\usepackage{amssymb}
\usepackage{makecell}
\usepackage{multirow} 
\usepackage{booktabs}
\usepackage{graphicx}
\usepackage{float}
\title{Can MLLMs "Read" What is Missing? }

\author{%
  \textbf{Jindi Guo} \\
  DP Technology \\
  \texttt{guojindi@dp.tech} \\
  \And
  \textbf{Chaozheng Huang} \\
  DP Technology \\
  \texttt{huangchaozheng@dp.tech} \\
  \And
  \textbf{Xi Fang} \\
  DP Technology \\
  \texttt{fangxi@dp.tech} \\
}

\begin{document}

\maketitle

\begin{abstract}
We introduce MMTR-Bench, a benchmark designed to evaluate the intrinsic ability of Multimodal Large Language Models (MLLMs) to reconstruct masked text directly from visual context. Unlike conventional question-answering tasks, MMTR-Bench eliminates explicit prompts, requiring models to recover masked text from single- or multi-page inputs across real-world domains such as documents and webpages. This design isolates the reconstruction task from instruction-following abilities, enabling a direct assessment of a model's layout understanding, visual grounding, and knowledge integration. MMTR-Bench comprises 2,771 test samples spanning multiple languages and varying target lengths. To account for this diversity, we propose a level-aware evaluation protocol. Experiments on representative MLLMs show that the benchmark poses a significant challenge, especially for sentence- and paragraph-level reconstruction. The homepage is available at \href{https://mmtr-bench-dataset.github.io/MMTR-Bench/}{MMTR-Bench}.

\end{abstract}

\section{Introduction}
\label{sec:intro}

\begin{figure*}[h!]
    \centering
    \includegraphics[width=1\textwidth]{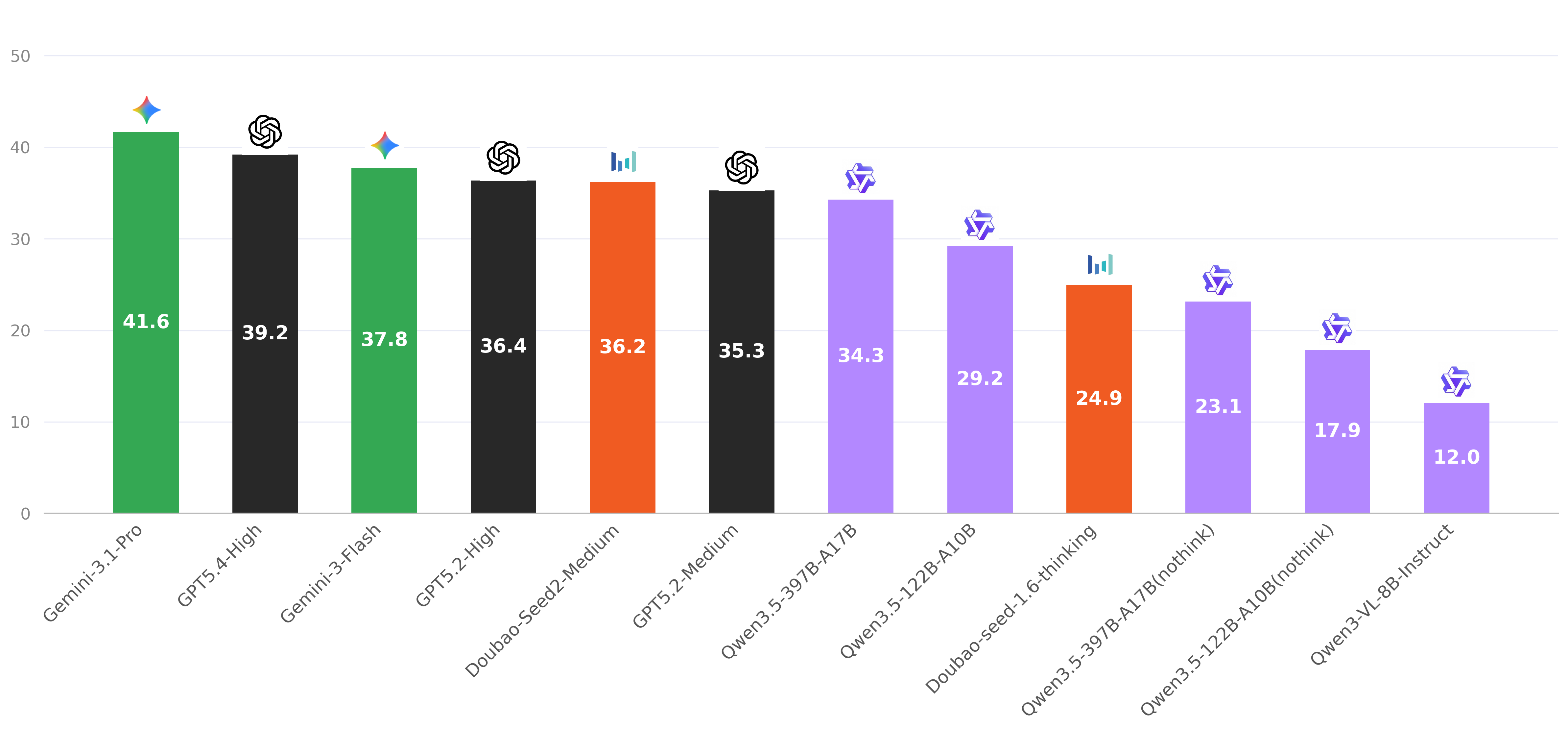}
    \caption{Overall performance of representative models on MMTR-Bench. Models from the same provider share the same color. Strong closed-source models achieve the best results, while smaller open-source vision-language models remain clearly behind.}
    \label{fig:maskar_main_bar}
\end{figure*}

In recent years, multimodal large language models (MLLMs) have made significant progress in understanding documents, charts, and webpages. However, most existing benchmarks still rely heavily on explicit question answering (QA). In these tests, models are given an image along with a clear question that tells them exactly what to look for.

While the QA format is useful, it does not fully reflect how models process real-world visual data. In practical scenarios---such as reading papers, analyzing multi-page reports, or parsing complex webpages---inputs do not come with guiding prompts. Instead, information is naturally distributed across layouts, tables, figures, and cross-page text. To truly understand this content, an MLLM must identify structural gaps and recover missing information by combining the surrounding visual context with its own world knowledge.

Although masking is widely used for training, we still lack a benchmark to test this native recovery ability. To fill this gap, we introduce \textbf{MMTR-Bench} (Multimodal Masked Text Reconstruction Benchmark). Instead of asking explicit questions, we give models masked single- or multi-page inputs. The task is to recover the hidden text using the remaining visual and structural context---such as titles, charts, table layouts, and cross-page clues---or by integrating this context with world knowledge. MMTR-Bench includes 2,771 test samples across multiple languages. It covers diverse sources, ranging from academic documents and webpage screenshots to natural scene text. The masked targets also vary in length, from short strings (like years or numbers) to full sentences and explanatory paragraphs.

Since targets of different lengths need different evaluation criteria, we design a \textbf{level-aware evaluation pipeline}. We divide the samples into four levels. For short targets, we focus on exact matching. For longer ones, we measure semantic similarity and factual consistency. To ensure high-quality scoring for complex targets, we also introduce an LLM-based factuality gate. This approach brings the automated metrics much closer to human judgment.

Finally, we evaluate several representative closed-source and open-source models,as illustrated
in Figure~\ref{fig:maskar_main_bar}. Our results show that MMTR-Bench is still highly challenging. Stronger closed-source models achieve the best results, but smaller open-source vision-language models clearly lag behind. Overall, there is still plenty of room for improvement, especially for sentence- and paragraph-level recovery under native visual inputs.

Our main contributions are as follows:
\begin{itemize}
    \item We introduce \textbf{MMTR-Bench}, a new benchmark designed to evaluate native multimodal perception and reasoning through masked visual context recovery, moving away from explicit question-based guidance.
    \item We build a diverse test set of 2,771 samples. It covers single- and multi-page inputs, multiple languages, and various real-world sources.
    \item We propose a \textbf{level-aware evaluation pipeline}, integrating lexical matching, semantic similarity, and an LLM-based factuality gate, to fairly judge recovery targets of different lengths and difficulties.
\end{itemize}

\section{Related work}

Existing multimodal benchmarks have substantially improved document, webpage, and chart understanding, but most of them are still framed as question answering. Representative examples include DocVQA~\cite{mathew2021docvqa} for document image question answering, InfographicVQA~\cite{mathew2022infographicvqa} for infographic understanding, ChartQA~\cite{masry2022chartqa} for chart reasoning, and WebSRC~\cite{chen2021websrc} for webpage reading comprehension over screenshots and HTML structure. More recent benchmarks such as DUDE~\cite{van2023document}, MMLongBench-Doc~\cite{ma2024mmlongbench}, LongDocURL~\cite{deng2025longdocurl}, and M-LongDoc~\cite{chia2025m} further extend evaluation to multi-page or long-context settings, showing that current models still struggle when evidence is distributed across pages and layout regions. In addition, WorldVQA~\cite{zhou2026worldvqa} focuses on measuring atomic visual world knowledge in MLLMs, emphasizing whether models can correctly ground and recognize real-world entities rather than perform only task-local reasoning. This perspective is also relevant to our setting, since some masked targets in MMTR-Bench cannot be recovered solely from local string matching and instead require implicit world knowledge together with surrounding visual and structural context. However, unlike WorldVQA~\cite{zhou2026worldvqa} and most prior benchmarks, our work is centered on reconstructing masked content directly from multimodal context rather than answering explicit questions.

Masking and reconstruction have also been widely used in document and webpage modeling, but mostly as training objectives rather than standalone evaluation tasks. LayoutLMv3~\cite{huang2022layoutlmv3} uses unified text and image masking for document pre-training, UDOP~\cite{tang2023unifying} combines vision, text, and layout modeling with reconstruction-style objectives, and Pix2Struct~\cite{lee2023pix2struct} treats masked screenshot parsing as a pretraining signal for visually situated language understanding. These works suggest that reconstruction-based learning is useful for structured multimodal inputs, but they do not directly provide a benchmark centered on masked contextual reconstruction. In terms of evaluation, prior work on generative tasks often combines lexical metrics and semantic metrics, and recent LLM-based evaluators further indicate that no single metric is sufficient for all answer lengths and granularities~\cite{gu2024survey}. This motivates our level-aware evaluation design for MMTR-Bench.

To make the position of our benchmark clearer, Table~\ref{tab:benchmark_comparison} summarizes the main differences between MMTR-Bench and several representative benchmarks. While existing work has substantially improved evaluation across documents, charts, webpages, and world knowledge, most are still built around explicit question answering. In contrast, MMTR-Bench abandons explicit questions, requiring models to recover masked targets from surrounding visual, structural, and semantic context. Furthermore, unlike benchmarks such as WorldVQA that focus primarily on visual world knowledge, MMTR-Bench integrates this as just one component of a broader contextual reconstruction framework, uniquely combining multi-page inputs, diverse source coverage, long-context samples, and a level-aware evaluation protocol.

\begin{table*}[t]
    \centering
    \caption{Comparison with related benchmarks. MMTR-Bench differs from prior work mainly in task format and evaluation design: it focuses on masked context reconstruction rather than explicit question answering, while also supporting broad visual-text sources, world-knowledge-dependent cases, and level-aware evaluation.}
    \label{tab:benchmark_comparison}
    \scriptsize
    \setlength{\tabcolsep}{3.4pt}
    \begin{tabular}{lccccccc}
        \toprule
        Benchmark 
        & \makecell{Explicit\\QA} 
        & \makecell{Multi-page /\\multi-page} 
        & \makecell{Web / Chart /\\Scene\\Coverage} 
        & \makecell{Masked\\Reconstruction} 
        & \makecell{Long-context /\\Long-form} 
        & \makecell{World\\Knowledge} 
        & \makecell{Level-aware\\Evaluation} \\
        \midrule
        DocVQA~\cite{mathew2021docvqa} & \checkmark & -- & -- & -- & -- & -- & -- \\
        ChartQA~\cite{masry2022chartqa} & \checkmark & -- & \checkmark & -- & -- & -- & -- \\
        WebSRC~\cite{chen2021websrc} & \checkmark & -- & \checkmark & -- & -- & -- & -- \\
        DUDE~\cite{van2023document} & \checkmark & \checkmark & -- & -- & -- & -- & -- \\
        MMLongBench-Doc~\cite{ma2024mmlongbench} & \checkmark & \checkmark & -- & -- & \checkmark & -- & -- \\
        LongDocURL~\cite{deng2025longdocurl} & \checkmark & \checkmark & -- & -- & \checkmark & -- & -- \\
        M-LongDoc~\cite{chia2025m} & \checkmark & \checkmark & -- & -- & \checkmark & -- & -- \\
        WorldVQA~\cite{zhou2026worldvqa} & \checkmark & -- & \checkmark & -- & -- & \checkmark & -- \\
        MMTR-Bench & -- & \checkmark & \checkmark & \checkmark & \checkmark & \checkmark & \checkmark \\
        \bottomrule
    \end{tabular}
\end{table*}

\section{Task and Benchmark}
\label{sec:task_benchmark}
\subsection{Task Definition}
\label{ssec:task_definition}

MMTR-Bench studies a masked visual context reconstruction task. Given one or more images with a locally masked region, the model is asked to recover the target text hidden by the mask from the remaining text, layout structure, chart elements, and other contextual cues in the input.

\begin{figure*}[t]
    \centering
    \includegraphics[width=0.97\textwidth]{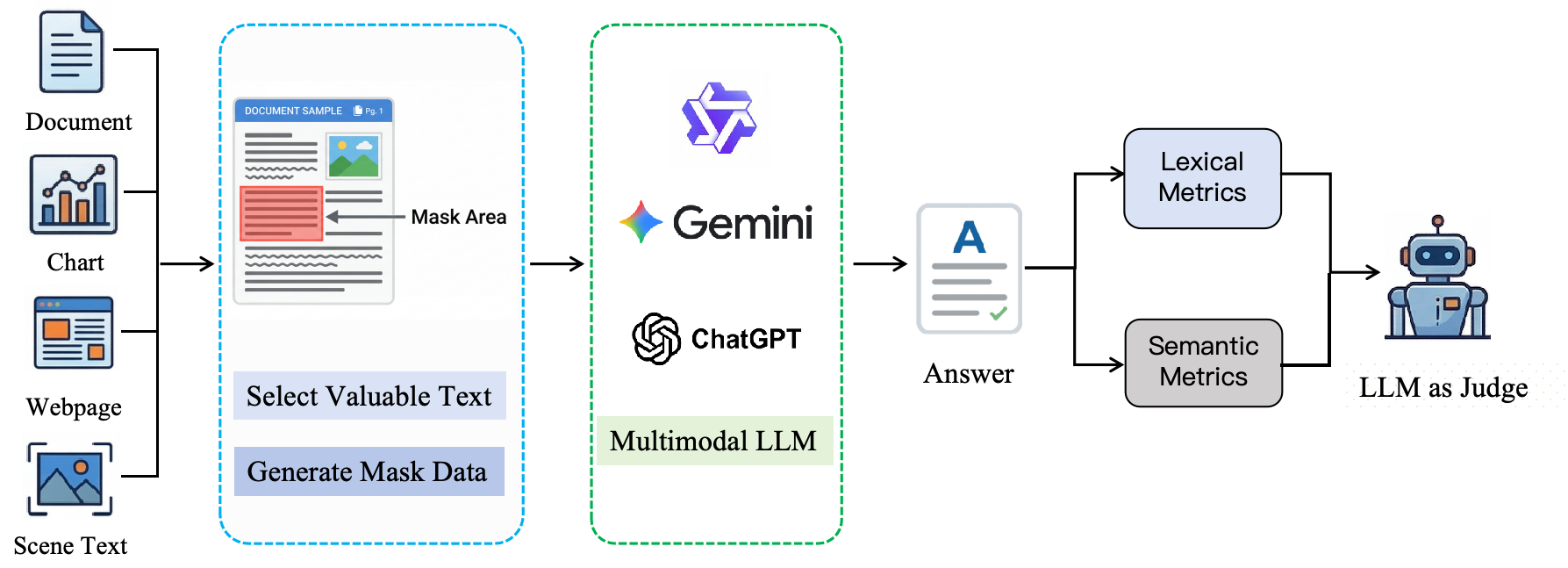}
    \caption{The pipeline consists of four main stages: (1) Data Preparation, where valuable text is selected and masked; (2) Inference, involving various multimodal LLMs (e.g., Gemini and ChatGPT); (3) Metric Calculation, focusing on lexical and semantic features; and (4) Automated Assessment, which uses an LLM-as-Judge to determine the final score.}
    \label{fig:maskar_pipe}
\end{figure*}
This task is different from traditional visual question answering. In standard VQA, the model is usually given both an image and a question, and the question already tells the model what to look for. In our task, there is no extra question. The model must first identify which regions are related to the masked target, and then use these clues to recover the missing content. Because of this, the task depends more on global understanding of the input rather than response to a local prompt.

MMTR-Bench includes both single-page and multi-page samples. single-page samples mainly test local relation modeling and layout understanding within one page or one image. multi-page samples require the model to combine evidence across pages or images in order to recover the target content. This makes the task more challenging than simple local completion.

The masked target can also vary a lot in length. It may be a short item such as a year, a number, or an entity name. It may also be a full sentence or a short explanatory paragraph. For this reason, MMTR-Bench is not just an OCR completion task. It is designed to test whether a model can recover meaning from visual context.

\subsection{Benchmark Construction}
\label{ssec:benchmark_construction}

The samples in MMTR-Bench are drawn from several common but challenging visual-text settings, including academic documents, webpage screenshots, charts and diagrams, natural scene text, and multi-page long documents. We chose these sources because they typically feature high information density, clear layout structures, and rich visual interference, making them ideal for testing context-based recovery in realistic scenarios.

To ensure the integrity of the benchmark and reduce the risk of data contamination (data leakage) from the pre-training corpora of current MLLMs, we implemented a strict temporal cutoff. All newly collected academic papers (e.g., from arXiv), public literature, and webpage screenshots are strictly dated after June 2025. For samples drawn from existing high-quality OCR benchmarks like OmniDocBench, the novel task of targeted masking effectively forces models to rely on zero-shot visual reasoning rather than memorized sequences.

During construction, we completely discarded automated random masking. Instead, human experts carefully selected and masked text regions based on four strict principles: 
First, the masked content must have a clear relationship with its surrounding context, rather than being an isolated fragment. 
Second, the remaining regions must provide enough evidence for recovery, ensuring the task tests contextual reasoning rather than pure guessing. 
Third, the sample must force the model to use visual content, structural cues, or cross-region relations, instead of relying solely on language priors. 
Finally, every sample underwent a secondary human review to permanently exclude cases with strong ambiguity or without a unique, deterministic answer.

These principles help make the benchmark highly focused. The goal is not to hide random text and test whether the model can guess it. The goal is to test whether the model can effectively use layout, image-text relations, and surrounding evidence to recover missing content in a reliable way. This fundamental requirement clearly separates MMTR-Bench from standard text masking or simple OCR completion tasks.

To further improve coverage, MMTR-Bench includes multilingual samples (spanning 22 languages, including Chinese, English, Japanese, and Korean) and multi-page inputs that require cross-page evidence integration. The structural distribution also reflects our design choice: while most samples are visually structured and text-rich, we intentionally keep a smaller set of noisier, open-layout cases to avoid making the benchmark too narrow or artificially clean.

\subsection{Dataset Statistics}
\label{ssec:dataset_statistics}

At present, MMTR-Bench contains 2771 test samples in total. Among them, 2268 are single-page samples, accounting for 81.85\%, and 503 are multi-page samples, accounting for 18.15\%. Overall, the benchmark is still dominated by single-page tasks, while a meaningful portion of multi-page samples is kept to test cross-page evidence integration.

By evaluation level, the dataset contains 401 Level 1 samples (14.47\%), 1377 Level 2 samples (49.69\%), 893 Level 3 samples (32.23\%), and 100 Level 4 samples (3.61\%). With this updated distribution, the vast majority of the benchmark focuses on medium-difficulty Level 2 and Level 3 tasks. Short, rigid targets (Level 1) make up a baseline portion, while highly complex paragraph-level samples (Level 4) remain the fewest but most challenging.

\begin{figure*}[t]
    \centering
    \includegraphics[width=0.96\textwidth]{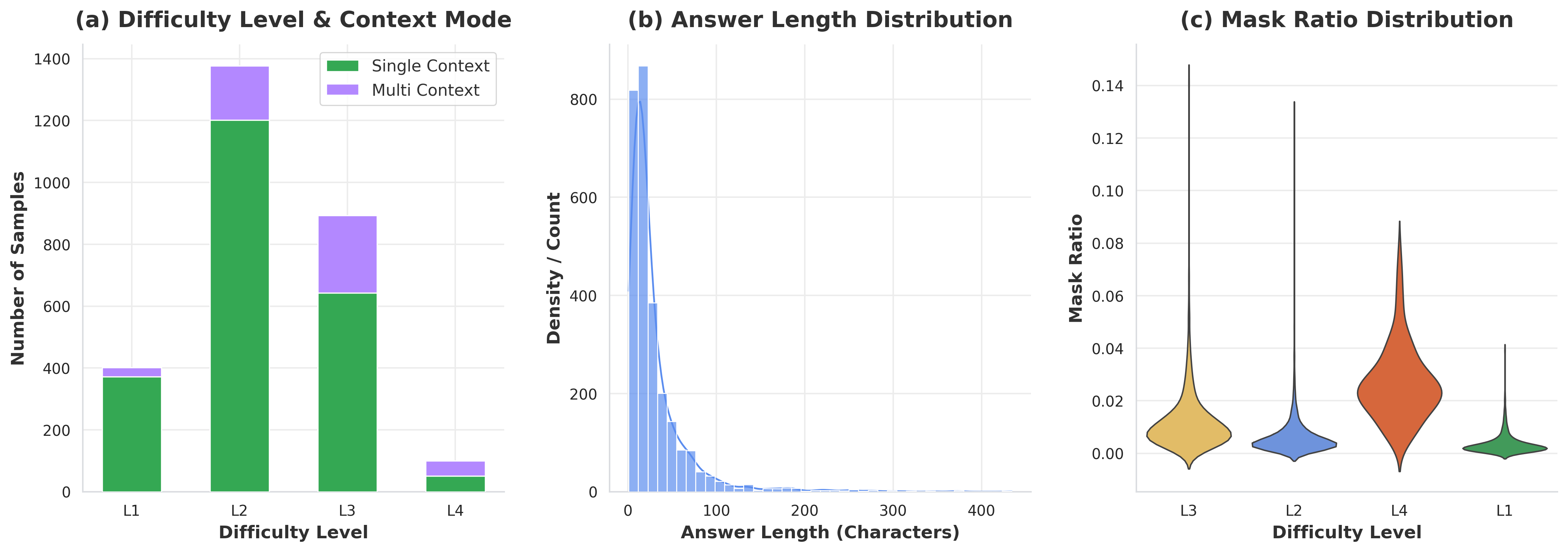}
    \caption{Overview statistics of MMTR-Bench, including level distribution, single-page versus multi-page composition, answer length distribution, and mask ratio distribution. The benchmark is centered on medium-length reconstruction targets and uses mainly local masking rather than large-area masking.}
    \label{fig:benchmark_overview}
\end{figure*}

This distribution is kept on purpose. If too many easy short-text samples are included, models may obtain high scores mainly by local recognition or template-like completion. In that case, the benchmark would be less useful for distinguishing context reconstruction ability. By placing the most weight on Level 2 and Level 3 cases, the current distribution better reflects differences in contextual modeling and semantic recovery without skewing the evaluation toward extreme lengths.

Figure~\ref{fig:benchmark_overview} provides a more detailed view of the dataset. The answer length distribution shows a clear long-tail pattern, with a median length of 18 characters and a maximum reaching 434 characters. Shorter targets are more common, but the benchmark still keeps a non-trivial number of longer sentence-level and paragraph-level cases. This is consistent with the level distribution above. The mask ratio distribution is also concentrated at relatively low values, featuring a global median of just 0.0052. This means that the benchmark mainly uses local target masking instead of very large masked areas, making the task closer to fine-grained recovery from surrounding evidence.

Besides level and input mode, MMTR-Bench also has diverse data sources. The samples cover academic documents, textbooks, slides, webpage screenshots, natural scene text, charts and diagrams, and other open-domain content. These sources differ a lot in layout structure, text density, and visual noise, which also makes the benchmark more diverse and more challenging.

The semantic composition of the benchmark is also broad. Our dataset does not only cover plain body text. It also contains many chart-related targets, table structure, and a smaller number of formula and code-related targets. If we further look at layout elements, the benchmark includes titles, captions, main body text, table cells, floating badges, and header or footer elements. This means the task is not restricted to one fixed text region.

The context scope distribution gives another useful view. Many samples can be recovered from local context, but the benchmark also includes cross-modal cases and a smaller set of samples that require broader page-level or external knowledge cues. This makes MMTR-Bench more suitable for studying native multimodal perception and reasoning, rather than only short-range text completion.

\section{Evaluation and Experiments}
\label{sec:eval_experiment}

\subsection{Level-Aware Dynamic Evaluation}
\label{ssec:dynamic_eval}

The answers in MMTR-Bench vary greatly in length. Short targets might be a single year or an entity. Longer targets can span full paragraphs. Using one scoring rule for everything is unfair. To fix this, we group the samples into four levels based on length and complexity. 

For Level 1 (short and factual targets), we focus on strict accuracy. We use Exact Match (EM) to check if the prediction is completely right. We also allow minor spelling errors (like one wrong letter) using a string similarity metric. 

For Levels 2 to 4 (longer targets like phrases and paragraphs), strict word matching is too harsh. Instead, we evaluate semantic consistency. We combine Rouge-L to check the text structure and embedding similarity to check the actual meaning. As the target text gets longer, we rely more on semantic similarity, because long texts allow more flexible ways to express the same idea. All exact formulas, hyperparameter weights, and metric definitions are detailed in Appendix~\ref{app:metrics}.

\subsection{Factuality Gating}
\label{ssec:factuality_gate}

Standard semantic metrics have a hidden flaw. They can give high scores to an answer that sounds right but contains critical factual errors (such as a wrong date or name). 

To solve this, we introduce an LLM-based factuality gate for Levels 2 to 4. We use a strong open-source LLM (Qwen3.5) as a judge. Instead of asking the LLM to give a continuous score (like 1 to 10), we force it to make a simple binary choice: yes or no. The judge only checks for key factual errors. If the core facts match, it passes the prediction, and the model keeps its base score. If there is a critical error, it rejects the prediction, and we heavily penalize the final score. 

This 0/1 binary decision is simple and avoids common LLM hallucination issues. In our human evaluation of 100 random samples, this binary gate reached a 91.0\% agreement rate with human judges. The specific penalty math and full LLM prompts are provided in Appendix~\ref{app:factuality}.

\subsection{Experimental Setup}
\label{ssec:exp_setup}

We evaluate several mainstream multimodal models on MMTR-Bench, including strong closed-source models and smaller open-source baselines. Our goal is to see if the benchmark can clearly expose the performance gaps in visual context reconstruction.

We test all models directly on the benchmark without extra training. For each sample, the model takes the masked input and outputs the hidden text. We apply our level-aware scoring to each sample and aggregate the final results. We report both the overall score and the individual scores for Levels 1 to 4. This clearly shows how different models handle varying text lengths.

\begin{table*}[t]
\caption{Main results on MMTR-Bench. We report overall scores on single-page and multi-page samples, together with performance across four difficulty levels. ``Think'' marks models with explicit reasoning, except for variants explicitly marked as ``nothink'' or ``Instruct''. All numbers are reported as percentages.}
\label{tab:main_results_maskar_bench}
\centering
\small
\setlength{\tabcolsep}{4.6pt}
\begin{tabular}{lc|cc|cccc|c}
\toprule
\multirow{2}{*}{Models} & \multirow{2}{*}{Think}
& \multicolumn{2}{c|}{Page Type}
& \multicolumn{4}{c|}{Difficulty}
& \multirow{2}{*}{Final} \\
\cmidrule(lr){3-4} \cmidrule(lr){5-8}
& & Single-page & Multi-page & L1 & L2 & L3 & L4 & \\
\midrule
Gemini-3.1-Pro                 & $\checkmark$ & 42.57 & 38.70 & 64.17 & 44.64 & 37.50 & 31.86 & 41.87 \\
GPT5.4-High                    & $\checkmark$ & 41.00 & 30.98 & 57.46 & 41.20 & 35.72 & 30.92 & 39.18 \\
Gemini-3-Flash                 & $\checkmark$ & 38.49 & 34.90 & 56.75 & 38.51 & 34.86 & 29.46 & 37.84 \\
GPT5.2-High                    & $\checkmark$ & 36.64 & 37.62 & 51.49 & 38.61 & 34.02 & 29.42 & 36.81 \\
Doubao-Seed2-Medium            & $\checkmark$ & 37.06 & 31.96 & 52.46 & 36.10 & 33.63 & 31.28 & 36.13 \\
GPT5.2-Medium                  & $\checkmark$ & 35.39 & 36.61 & 50.27 & 37.22 & 32.72 & 30.51 & 35.61 \\
Qwen3.5-397B-A17B              & $\checkmark$ & 34.67 & 30.10 & 48.39 & 34.67 & 31.46 & 26.68 & 33.84 \\
Qwen3.5-122B-A10B              & $\checkmark$ & 30.37 & 23.94 & 43.91 & 27.23 & 27.84 & 23.92 & 29.20 \\
Doubao-Seed1.6-Thinking        & $\checkmark$ & 25.50 & 23.01 & 33.81 & 22.10 & 24.74 & 25.02 & 25.04 \\
Qwen3.5-397B-A17B              &              & 24.25 & 18.96 & 31.94 & 20.75 & 22.91 & 22.37 & 23.29 \\
Qwen3.5-122B-A10B              &              & 18.56 & 15.47 & 18.79 & 13.62 & 19.31 & 23.40 & 18.00 \\
Qwen3-VL-8B-Instruct           &              & 12.16 & 11.38 & 7.94  & 7.12  & 14.19 & 20.11 & 12.02 \\
\bottomrule
\end{tabular}
\end{table*}

\subsection{Main Results}
\label{ssec:main_results}

Table~\ref{tab:main_results_maskar_bench} reports the overall and level-wise results of representative models on MMTR-Bench. In general, there is a clear performance gap across models, which suggests that the benchmark can effectively distinguish their ability on visual context reconstruction.

Looking at the Overall score, the strongest closed-source models achieve the best results, while smaller open-source vision-language models remain much weaker. This suggests that context reconstruction in complex visual-text inputs still places high demands on visual recognition, evidence integration, and semantic generation.

\begin{figure*}[t]
    \centering
    \includegraphics[width=0.96\textwidth]{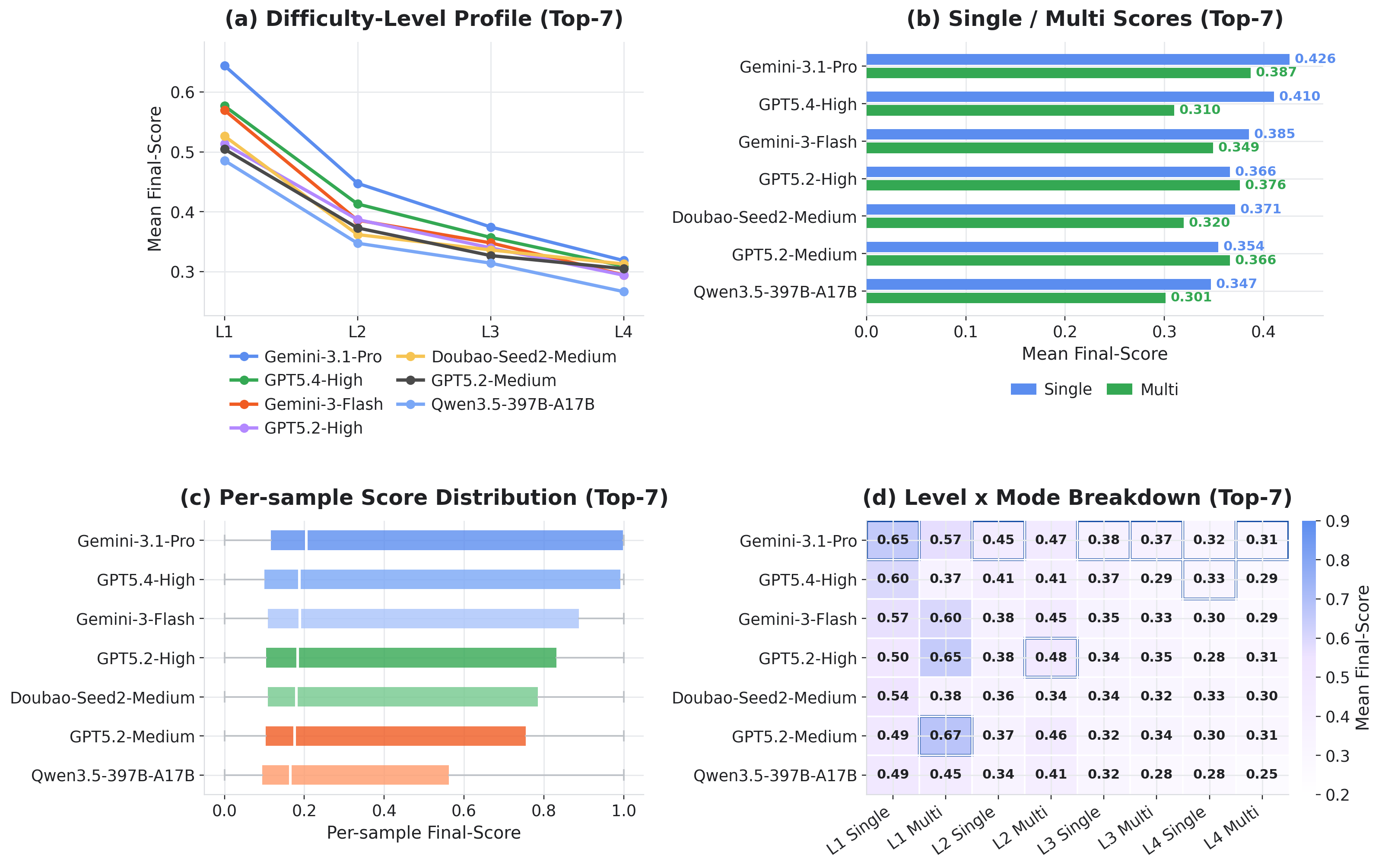}
    \caption{Compact overview of model behavior on MMTR-Bench, including difficulty-level trends, single-page versus multi-page performance, per-sample score distributions, and a detailed Level x Mode score breakdown for the top-7 models.}
    \label{fig:compact_overview}
\end{figure*}

Figure~\ref{fig:compact_overview} gives a compact view of model behavior across difficulty levels, input modes, and per-sample score distributions for the top-7 models. A clear trend appears in the level profile: all strong models perform best on Level 1, and the largest drop happens when moving from Level 1 to Level 2. This suggests that models are much less reliable once the target goes beyond short rigid spans and requires phrase-level or sentence-level recovery. Performance then continues to decrease from Level 2 to Level 4, but the drop is more gradual.

The single-page versus multi-page comparison shows that multi-page inputs are generally harder, but the gap is model-dependent. Some models show a clear drop when moving from single-pa ge to multi-page settings, while others are relatively more stable. This indicates that cross-page or cross-image evidence integration remains difficult, and current models do not handle it equally well. 

In addition, the Level x Mode breakdown in Figure~\ref{fig:compact_overview}(d) provides a more granular view of how difficulty and input mode interact. For most models, multi-page settings consistently yield lower scores than their single-page counterparts within the same difficulty level (e.g., L1 Single vs. L1 Multi). This confirms that integrating evidence across multiple images introduces an orthogonal challenge to the target length and complexity. Furthermore, the per-sample score distributions show that even strong models still face a wide spread of sample difficulty, ensuring that MMTR-Bench is not dominated by one narrow case type.

\subsection{Fine-grained Analysis by Semantics, Layout, and Visual Conditions}
\label{ssec:fine_grained_analysis}

\begin{figure*}[t]
    \centering
    \includegraphics[width=0.96\textwidth]{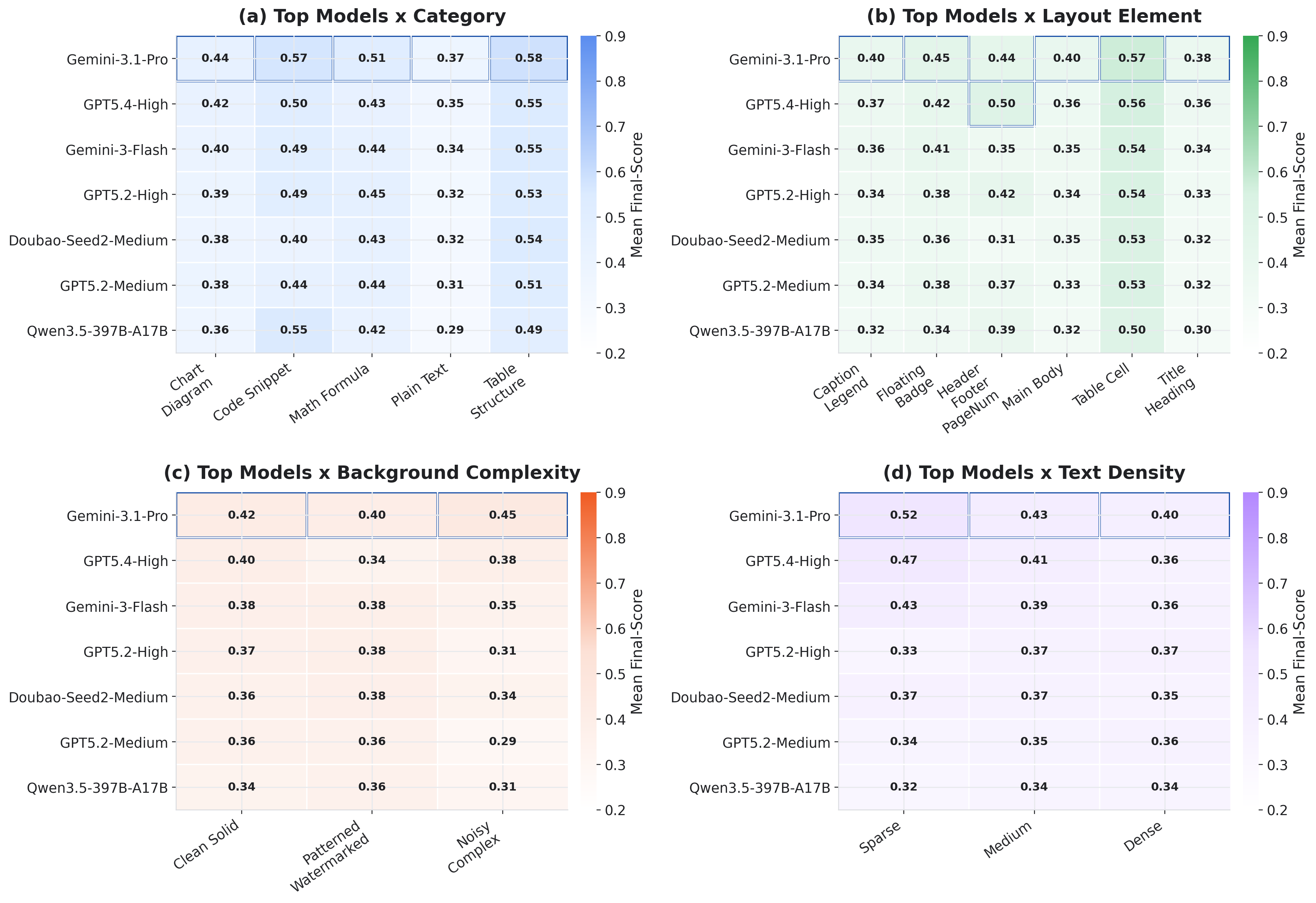}
    \caption{Model performance breakdown across semantic categories, layout elements, background complexity, and text density.}
    \label{fig:compact_analysis}
\end{figure*}

Figure~\ref{fig:compact_analysis} shows a comprehensive breakdown of model performance across different semantic categories, layout elements, and visual conditions. One interesting observation from the semantic and layout analyses is that plain-text recovery and main-body content are not always the easiest cases. In fact, highly structured targets, such as code snippets, table structure, and table cells, often obtain higher scores than plain text for top-performing models. A possible reason is that freer text spans allow more variation in wording and require stronger semantic control, while structured targets have clearer local constraints and predictable formats. Title headings and main-body content remain consistently difficult, suggesting that the benchmark effectively tests content reconstruction that depends on broader layout and semantic relations rather than just local token recognition.

The visual condition plots (Figure~\ref{fig:compact_analysis}c and d) reveal consistent robustness trends. Denser text inputs remain more difficult than sparse ones across all evaluated models, which aligns with the benchmark's focus on information-rich visual-text inputs. Similarly, background complexity significantly affects performance; clean solid backgrounds generally yield higher scores, while noisy, complex, or heavily watermarked backgrounds degrade the reconstruction ability of most systems. 

Taken together, these results demonstrate that benchmark difficulty is shaped not only by target length and semantic structure but also by visual crowdedness and background interference. Higher scores on some structured elements do not mean the benchmark is easy; instead, they highlight the varying behaviors of current models across different target types and visual conditions, underscoring the necessity of this multi-dimensional evaluation.

\subsection{Qualitative Analysis}
\label{subsec:case_study_diagram}

\begin{figure*}[!h]
    \centering
    \includegraphics[width=0.96\textwidth]{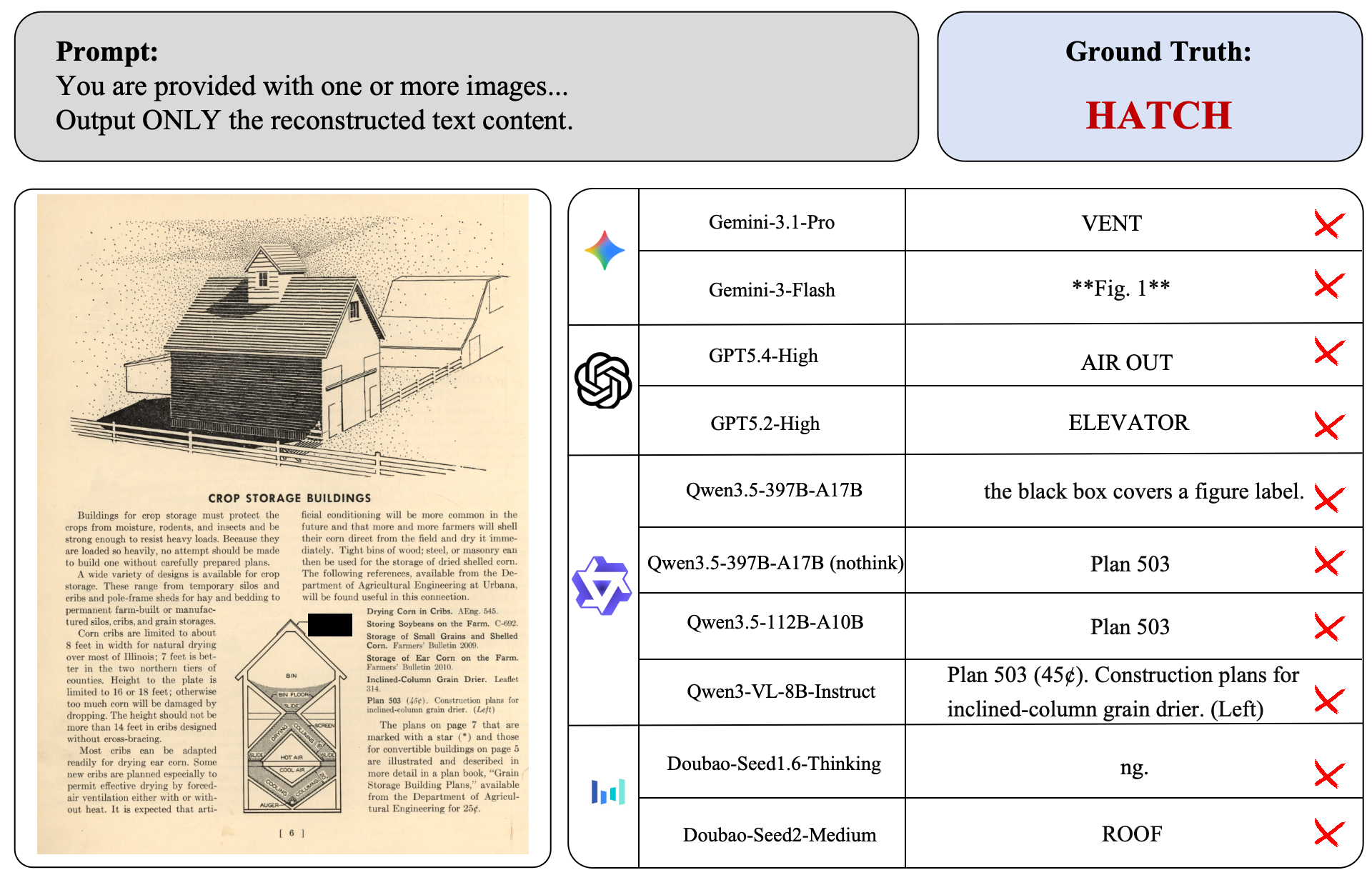}
    \caption{A challenging case of MMTR-Bench.}
    \label{fig:challengin}
\end{figure*}

To empirically demonstrate the high quality, complexity, and carefully curated nature of the samples in our proposed dataset, this section conducts a qualitative analysis of various Vision-Language Models (VLMs) responses. We select a highly representative and challenging example—an agricultural engineering diagram with masked text, alongside the respective model predictions, as illustrated in Figure \ref{fig:challengin}. 

We deliberately highlight this specific case because it perfectly encapsulates the core design philosophy of our benchmark: success extends far beyond basic Optical Character Recognition (OCR) or general image captioning. Instead, it demands a deep synthesis of spatial localization, contextual understanding, physical logic, and domain-specific knowledge. It is worth noting that a broader range of qualitative examples—including instances where most models easily succeed and extremely difficult cases where all models uniformly fail—are detailed and further analyzed in Appendix \ref{app:Result_Analysis}.

\subsubsection{Ground Truth Analysis}
The Ground Truth (GT) for the masked region is \textbf{HATCH}.
The masked bounding box is located at a protruding structure on the roof of the building. The lower sections of the schematic are explicitly labeled, including a ``BIN'' (storage bin) and areas indicating ``HOT AIR'' and ``COOL AIR''. In the context of agricultural storage facilities, grain is typically loaded from the top via elevators or conveyors through an opening. Therefore, identifying this top opening as a ``HATCH'' is highly accurate from an engineering perspective.

\subsubsection{Model Performance Categorization}
We categorize the model responses into four distinct behavioral patterns, highlighting the current capabilities and limitations of VLMs in complex reasoning tasks:

\paragraph{1. Logically Sound but Lacking Domain Prior (Gemini-3.1-Pro, GPT5.4-High)}
These models predicted \textit{VENT} and \textit{AIR OUT}. They demonstrate robust physical logic and cross-modal reasoning. Given the explicit text ``HOT AIR'', ``COOL AIR'', and ``DRYING'' in the lower structure, inferring that a roof opening serves as a ventilation system is a highly logical deduction based on the thermodynamics of rising heat. However, they fail to predict the exact GT due to a lack of domain-specific prior knowledge regarding agricultural loading procedures.

\paragraph{2. Correct Orientation but Overly Broad or Associative (Doubao-Seed2-Medium, GPT5.2-High)}
These models predicted \textit{ROOF} and \textit{ELEVATOR}. Doubao successfully grounds the location but provides a coarse-grained physical description rather than identifying the specific architectural component. Conversely, GPT5.2-High exhibits associative hallucination. While an elevator is indeed the external equipment used to transport grain to the hatch, the model mistakenly substitutes the external machinery for the building's structural label.

\paragraph{3. Layout Hallucination and Textual Distraction (Gemini-3-Flash, Qwen3.5-397B-A17B (nothink), Qwen3.5-122B-A10B, Qwen3-VL-8B-Instruct, Doubao-Seed1.6-Thinking)}
These models predicted \textit{**Fig. 1**}, fragmented strings like \textit{ng.}, or text such as \textit{Plan 503}. This group exhibits a complete failure in visual-spatial reasoning. Instead of tracing the indicator line from the masked box to the roof structure, they succumb to layout hallucinations common in academic papers or textbooks. The adjacent text explicitly states ``Plan 503... (Left)'' and contains nearby words like ``Engineering'' or ``drying'' (which likely triggered the fragmented ``ng.'' prediction). These models demonstrate a severe vulnerability in their visual attention mechanisms: they are overwhelmingly biased by adjacent dense text, treating the bounding box as a figure caption or title, and entirely ignoring fine-grained geometric cues.

\paragraph{4. Instruction Misalignment (Qwen3.5-397B-A17B)}
This model predicted \textit{the black box covers a figure label.} It fails to follow the zero-shot reasoning instruction. Instead of performing the masked text prediction, it falls back to a generic image captioning objective, merely describing the visual state of the image without attempting the underlying reasoning task.

\subsubsection{Discussion}
This case study illuminates two critical bottlenecks in current VLM architectures. First, there is a pronounced tension between general commonsense reasoning (e.g., heat rises $\rightarrow$ vent) and the necessity for specialized domain knowledge. Second, visual grounding remains fragile; models are heavily susceptible to textual distraction within document-like images. Instead of tracing fine-grained geometric cues like indicator lines, they prioritize nearby salient text blocks. In essence, these models are merely \textbf{``reading''} the textual layout rather than truly \textbf{``seeing''} the visual and geometric relationships within the image.Furthermore, a fundamental limitation in current evaluation is the difficulty of decoupling the root causes of model failures. When an incorrect prediction occurs, it is challenging to distinguish genuine deficits in visual reasoning from sub-optimal prompt engineering or poor instruction following (e.g., the ``Instruction Misalignment'' cases). Disentangling these confounding factors to provide a purer measurement of multimodal reasoning remains an open challenge for future work.

\section{Future Work}
\label{sec:future_work}

The results on MMTR-Bench show that current MLLMs still struggle to reconstruct missing visual semantics through long-range reasoning. To solve this, our next step is to evolve this benchmark task into a large-scale pre-training paradigm.

We believe that Multimodal Masked Text Reconstruction (MMTR) can serve as a unified objective for multimodal pre-training. Traditional methods treat pure text, interleaved data, and image-text pairs as separate data streams. Instead, we propose rendering all training corpora into native image sequences~\cite{wei2025deepseek}. By unifying these formats under a single ``Masked Image to Text'' objective, we can naturally combine language modeling with visual alignment, creating a stronger foundation for general document intelligence.

A key advantage of this evolution is learning world knowledge directly from professional literature, such as scientific papers, technical manuals, and legal archives. By masking and recovering text in these complex documents, models can learn logical reasoning straight from the original visual layouts~\cite{guoend}. This approach completely bypasses traditional parsing pipelines---such as OCR, layout analysis, and reading order restoration---and avoids their cascading errors. The model learns spatial and semantic correlations directly from raw pixels. Ultimately, by training models to recover missing text in complex global contexts, we can push them beyond simple visual perception toward deep, logic-driven comprehension.

\bibliographystyle{plainnat} 
\bibliography{references}   

\appendix

\section{Detailed Evaluation Metrics}
\label{app:metrics}

This section provides the exact formulas and hyperparameters for our level-aware evaluation pipeline. Table~\ref{tab:eval_levels} summarizes the metric weights and decay factors across all four levels.

\begin{table}[h!]
\centering
\caption{Overview of the level-aware dynamic evaluation strategy. $w$ and $\tau$ represent the semantic weight and factuality decay factor, respectively.}
\label{tab:eval_levels}
\begin{tabular}{llccc}
\toprule
\textbf{Level} & \textbf{Target Characteristics} & \textbf{Base Metrics} & \textbf{Weight ($w$)} & \textbf{Decay ($\tau$)} \\
\midrule
Level 1 & Short rigid strings (e.g., years, entities) & EM + ANLS & - & - \\
Level 2 & Short phrases or brief spans & Rouge-L + EmbedSim & 0.30 & 0.20 \\
Level 3 & Full sentences & Rouge-L + EmbedSim & 0.60 & 0.30 \\
Level 4 & Paragraphs or explanatory texts & Rouge-L + EmbedSim & 0.80 & 0.35 \\
\bottomrule
\end{tabular}
\end{table}

\subsection{Level 1 Scoring}
Level 1 combines Exact Match (EM) and Average Normalized Levenshtein Similarity (ANLS). 
\begin{equation}
EM(P, G) =
\begin{cases}
1, & \text{if } P = G \\
0, & \text{otherwise}
\end{cases}
\end{equation}
where $P$ is the model prediction and $G$ is the ground truth. 

For ANLS, let $dist(P, G)$ be the edit distance between $P$ and $G$. The normalized similarity is calculated as $Sim_{lev}(P, G) = 1 - \frac{dist(P, G)}{\max(|P|, |G|)}$. We only keep the ANLS score if it surpasses a threshold of 0.5:
\begin{equation}
ANLS(P, G) =
\begin{cases}
Sim_{lev}(P, G), & \text{if } Sim_{lev}(P, G) \ge 0.5 \\
0, & \text{otherwise}
\end{cases}
\end{equation}

The final score for Level 1 is a weighted sum:
\begin{equation}
Score_{L1}(P, G) = 0.7 \cdot EM(P, G) + 0.3 \cdot ANLS(P, G)
\end{equation}

\subsection{Levels 2 to 4 Scoring}
For longer texts, we combine Rouge-L and cosine semantic similarity ($EmbedSim$). Given the embeddings $v_p$ and $v_g$ of the prediction and ground truth, the base score is defined as:
\begin{equation}
EmbedSim(P, G) = \max\left(0, \min\left(1, \frac{v_p \cdot v_g}{\|v_p\| \|v_g\|}\right)\right)
\end{equation}

Using the semantic weight $w$ from Table~\ref{tab:eval_levels}, the base score is:
\begin{equation}
Score_{base}(P, G) = (1-w) \cdot RougeL(P, G) + w \cdot EmbedSim(P, G)
\end{equation}

\subsection{Factuality Gating Details}
\label{app:factuality}

For Levels 2 to 4, we use a binary LLM judge ($Judge_{output} \in \{0, 1\}$). The final score incorporates a decay factor $\tau$ (from Table~\ref{tab:eval_levels}) to heavily penalize critical factual errors:
\begin{equation}
Score_{final}(P, G) = Score_{base}(P, G) \cdot [\tau + (1-\tau) \cdot Judge_{output}]
\end{equation}

If the prediction passes the check ($Judge_{output}=1$), it keeps its original score. If it fails ($Judge_{output}=0$), the score drops significantly to $Score_{base} \cdot \tau$.

\newpage

\section{Evaluation Prompt}
\label{app:eval_prompt}

For zero-shot evaluation on MMTR-Bench, we use the following unified prompt template for all tested models.

\begin{tcolorbox}[
    colback=white,
    colframe=black!70,
    title=\textbf{\texttt{Evaluation Prompt (Masked Text Reconstruction)}},
    arc=3mm,
    boxrule=1pt,
    fonttitle=\bfseries,
    left=3mm, right=3mm, top=3mm, bottom=3mm
]

You are an advanced multimodal document understanding assistant specialized in content restoration. Your capability involves identifying obscured or masked information in visually rich documents and reconstructing it accurately using available context.

\vspace{0.5em}

You are provided with one or more images representing a document context. In one of the images, some text regions have been masked out (covered by black boxes). Please visually locate these masked regions and reconstruct the original text content based on the surrounding context and other provided pages. Output ONLY the reconstructed text content.

\end{tcolorbox}

\newpage

\section{Model Result Analysis}
\label{app:Result_Analysis}
\subsection{High-scoring Case Analysis}

In this chapter, we have selected representative cases from various categories where almost all models achieved nearly perfect scores.
\subsubsection{High-scoring Case 1}
This sample is a screenshot of a webpage where the occluded area features one of the lead actors from the Titanic. Solving this question requires the model to possess world knowledge for analysis or to perform reasoning by combining the primary visual elements in the image with its internal knowledge base. Since all models answered this question correctly, it demonstrates that even small-scale models possess a certain degree of world knowledge and reasoning capabilities.

\begin{figure*}[h]
    \centering
    \includegraphics[width=0.97\textwidth]{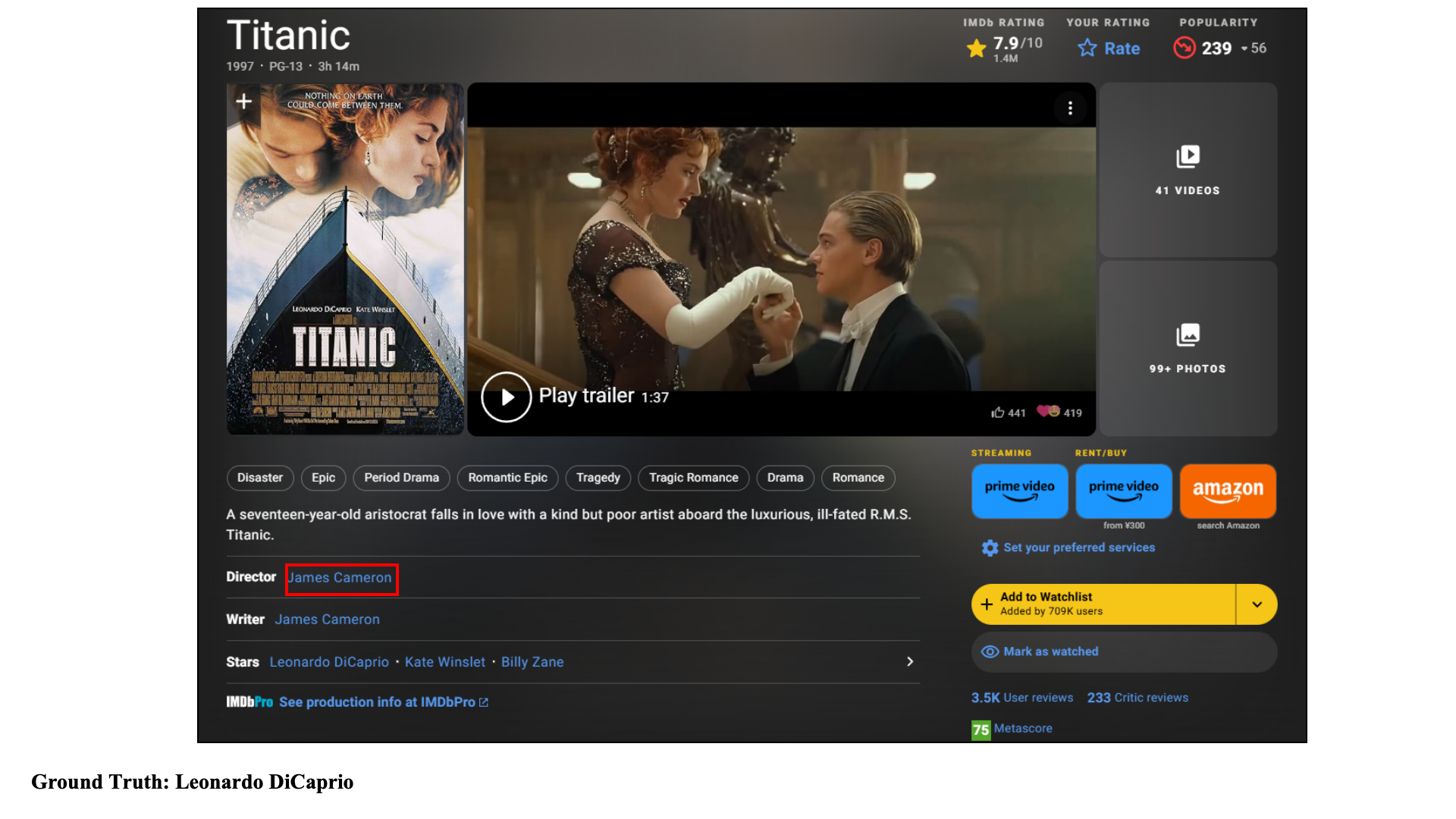}
    \caption{High-scoring Case 1}
    \label{fig:hsc1}
\end{figure*}

\subsubsection{High-scoring Case 2}
This sample is a webpage screenshot designed to test the model's world knowledge regarding gaming. The model can infer the answer from other category tags or by reasoning through the Xbox console news already visible in the image.

\begin{figure*}[h]
    \centering
    \includegraphics[width=0.8\textwidth]{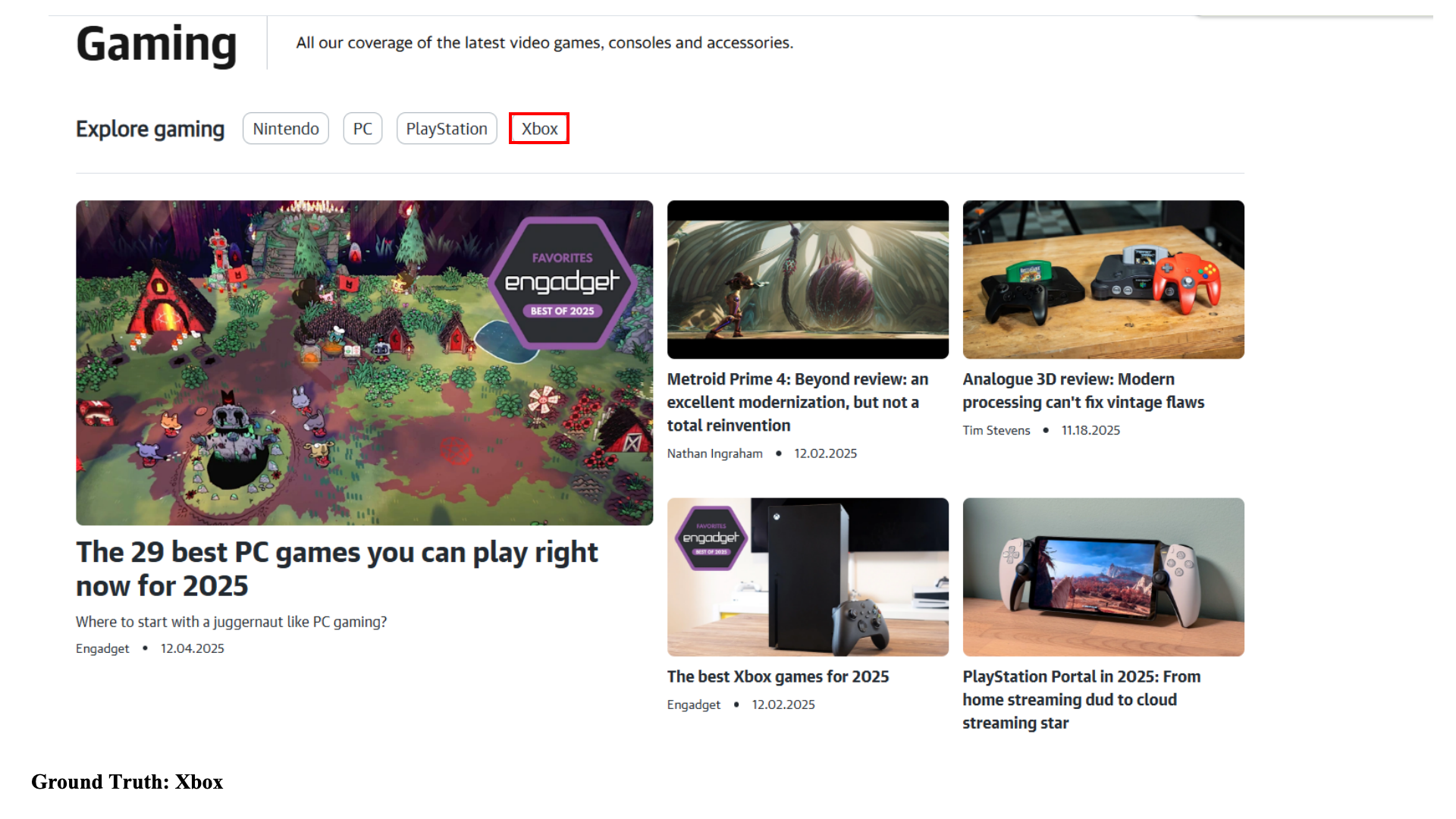}
    \caption{High-scoring Case 2}
    \label{fig:hsc2}
\end{figure*}
\newpage

\subsubsection{High-scoring Case 3}
This sample is an information-rich illustration from an academic publication, containing only the image and its title. All models performed the reasoning correctly for this sample, proving that even models with 8B parameters possess a certain level of proficiency in paper reading.

\begin{figure*}[h]
    \centering
    \includegraphics[width=0.97\textwidth]{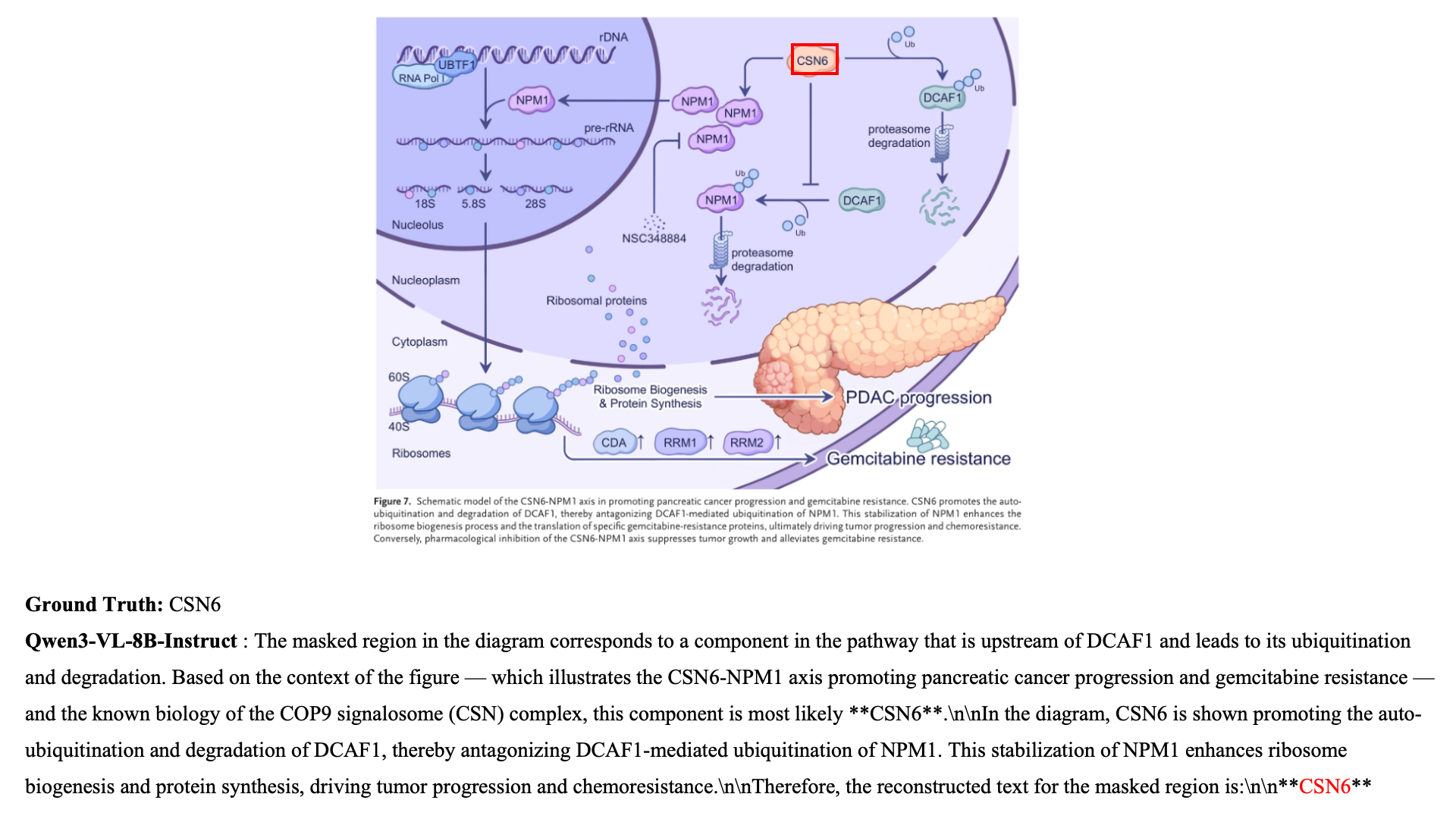}
    \caption{High-scoring Case 3}
    \label{fig:hsc3}
\end{figure*}
\newpage

\subsubsection{High-scoring Case 4}
This sample features an academic illustration with a brief, non-descriptive title. Consequently, the model must rely entirely on the visual content for its analysis. Solving this task requires the model to comprehend the flowchart and perform reasoning based on knowledge of geological hazards. The results indicate that current mainstream models possess extensive academic knowledge across various fields, and even 8B small-scale models are capable of understanding simple flowcharts.

\begin{figure*}[h]
    \centering
    \includegraphics[width=0.97\textwidth]{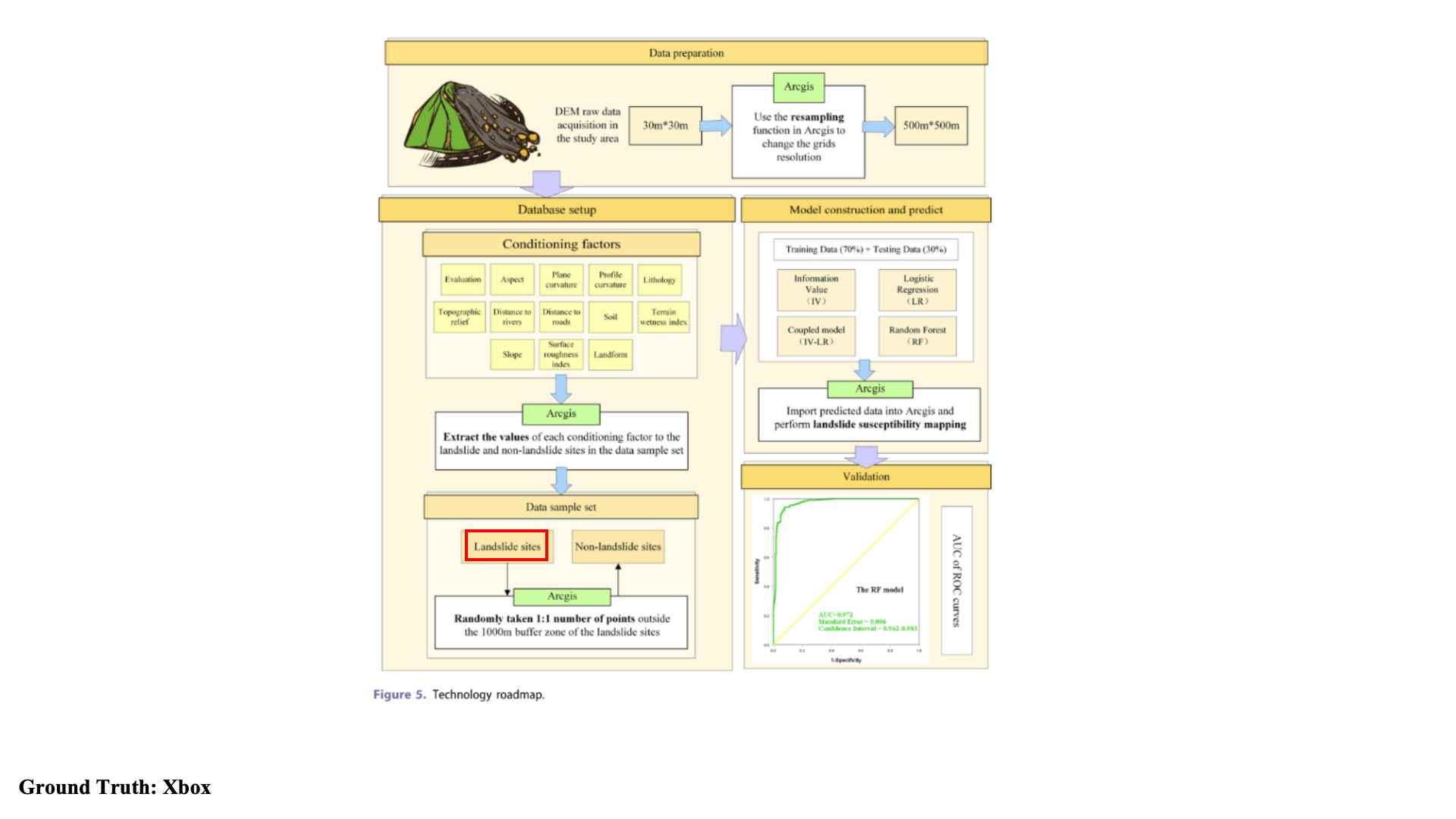}
    \caption{High-scoring Case 4}
    \label{fig:hsc4}
\end{figure*}

\subsubsection{High-scoring Case 5}
This sample tests the model's image analysis capabilities and world knowledge. The model must locate the position of tooth \#2 and perform reasoning based on that placement. Additionally, some models may infer the answer by identifying which specific tooth type is missing from the existing set.

\begin{figure*}[h]
    \centering
    \includegraphics[width=0.8\textwidth]{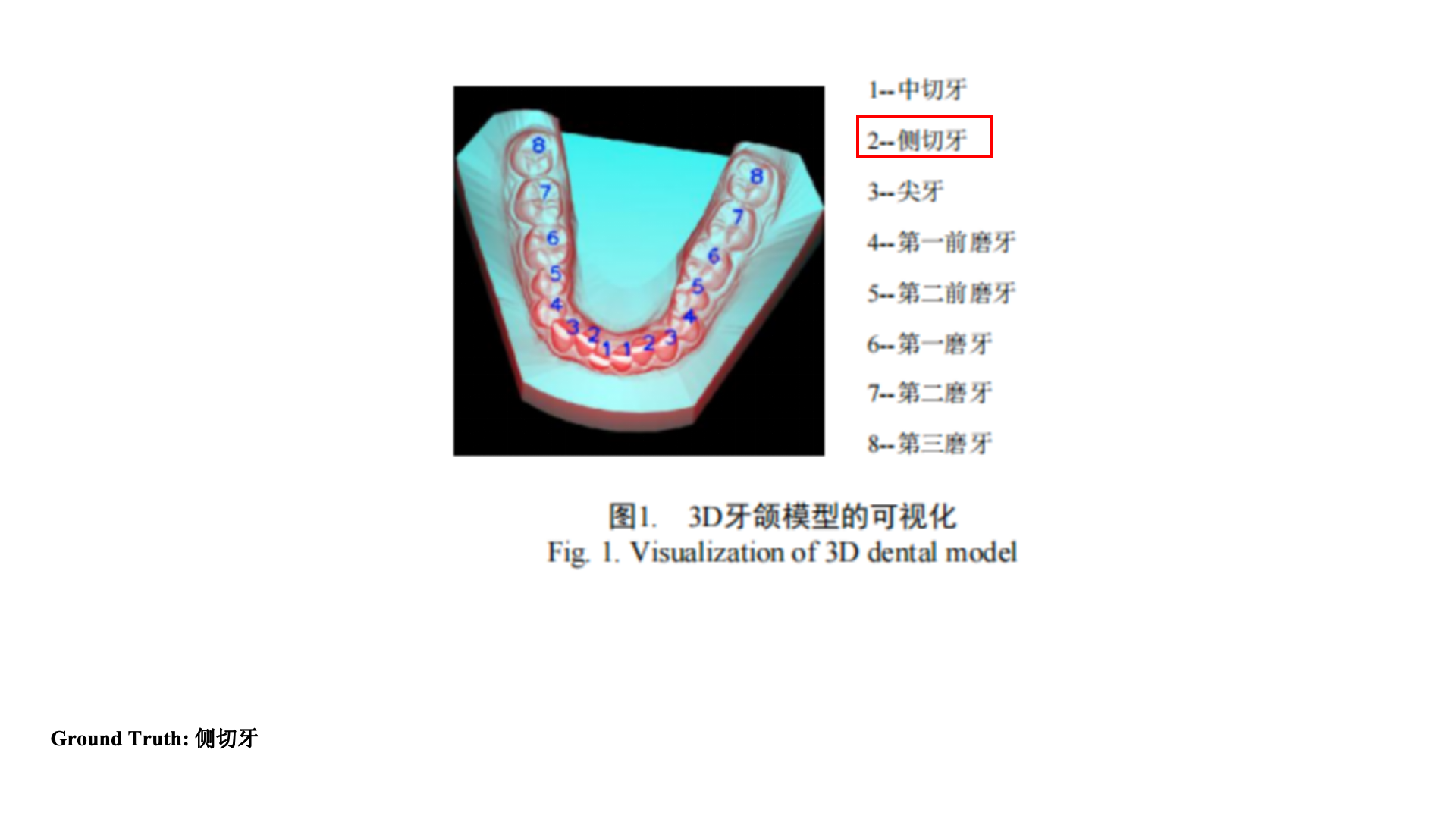}
    \caption{High-scoring Case 5}
    \label{fig:hsc5}
\end{figure*}
\newpage

\subsubsection{High-scoring Case 6}
This document serves as a pedagogical resource for model construction using TensorFlow. By occluding an intermediate code block, we evaluate the models' programmatic logic. The findings demonstrate that most models now possess advanced capabilities in code synthesis and contextual script analysis.

\begin{figure*}[h]
    \centering
    \includegraphics[width=0.97\textwidth]{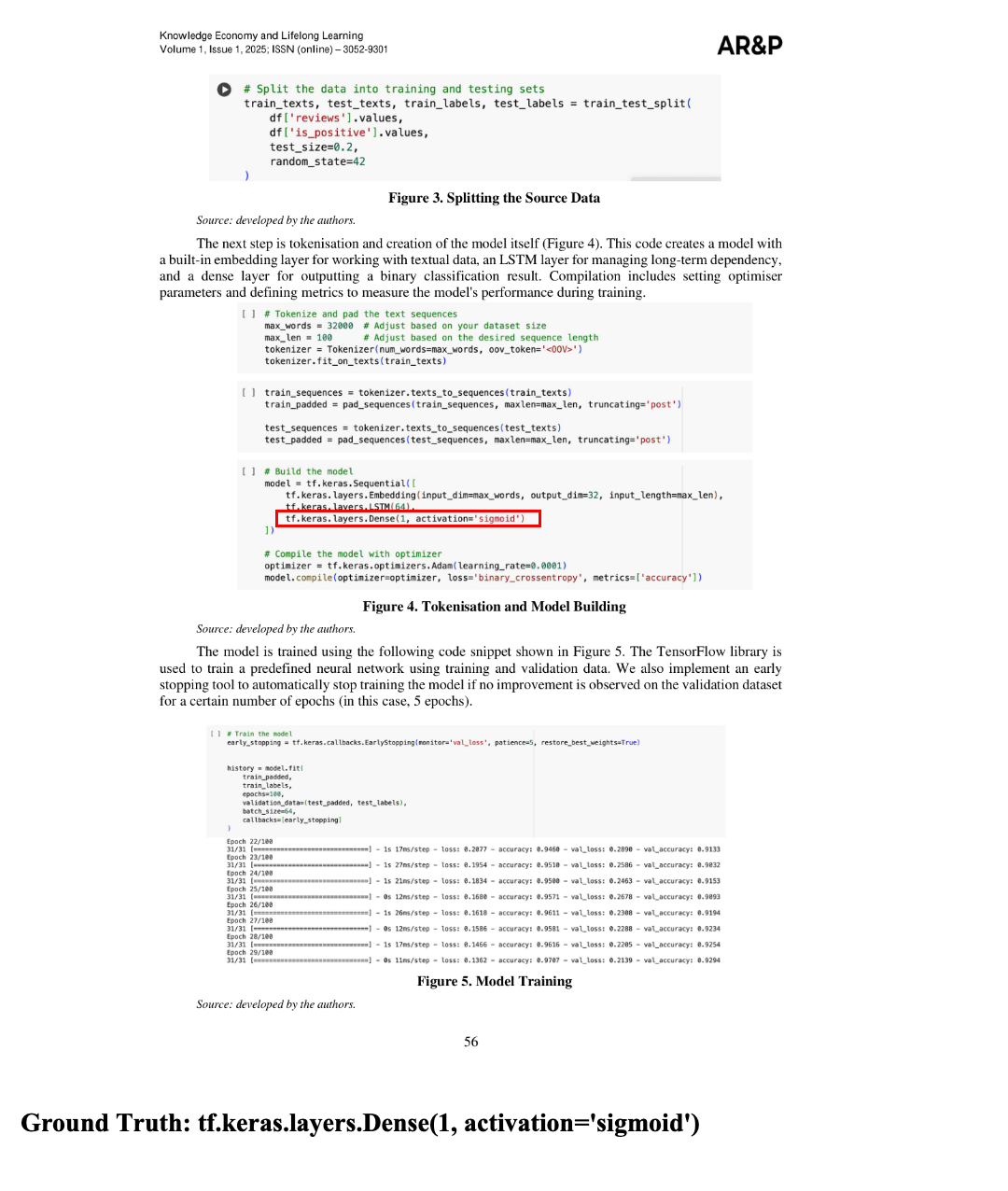}
    \caption{High-scoring Case 7}
    \label{fig:hsc7}
\end{figure*}
\newpage

\subsubsection{High-scoring Case 7}
This sample involves a multi-page query where the current page lacks a direct reference to the occluded title. However, cross-references exist on subsequent pages, which also contain textual descriptions of the image in question. The results indicate that contemporary models have begun to exhibit a nascent capability for cross-page contextual integration and the organization of multi-modal information.

\begin{figure*}[h]
    \centering
    \includegraphics[width=0.97\textwidth]{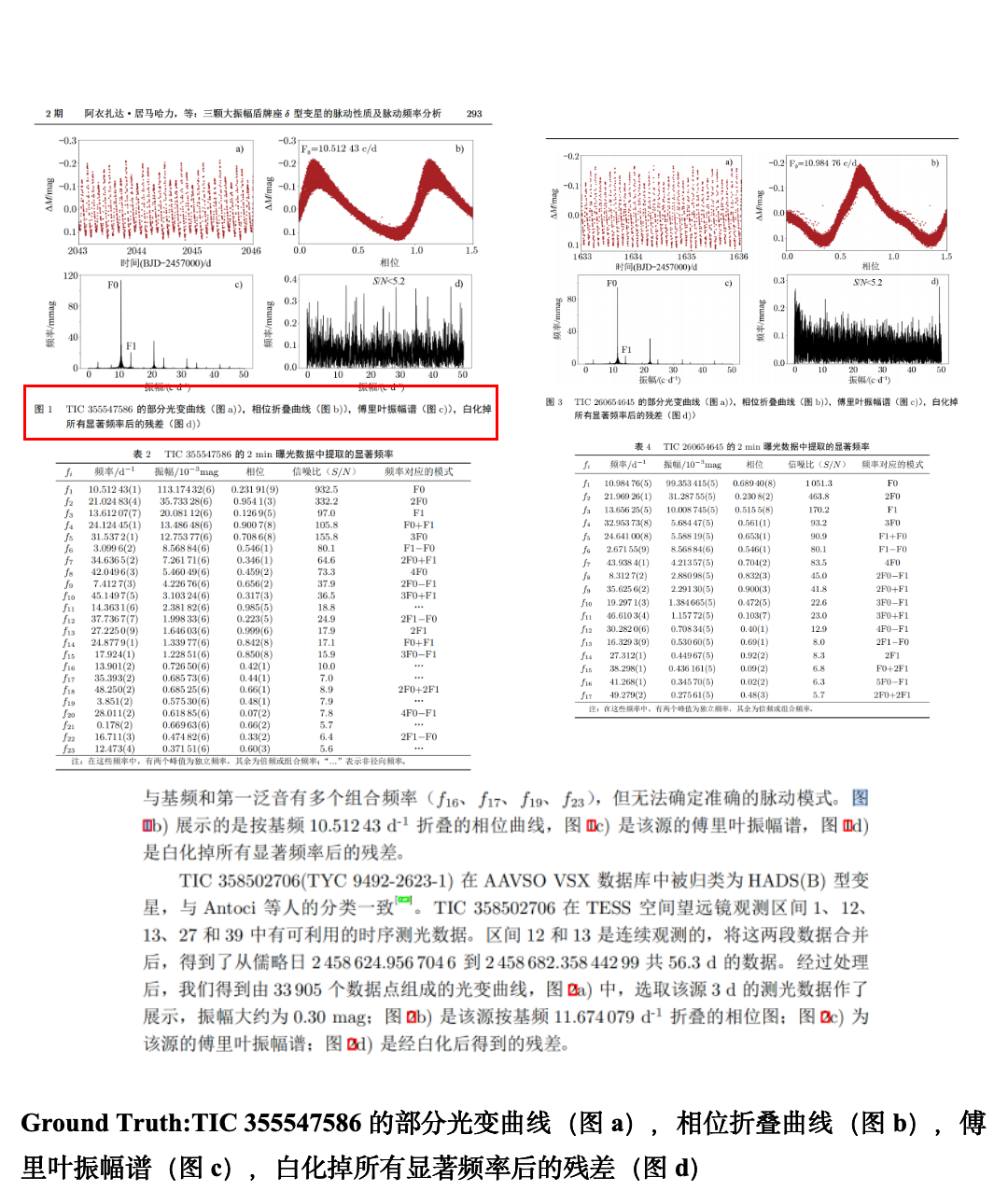}
    \caption{High-scoring Case 8}
    \label{fig:hsc8}
\end{figure*}
\newpage

\subsection{Low-scoring Case Analysis}

In this subsection, we present representative failure cases across various domains. These samples proved consistently difficult for the tested models, regardless of their specific architecture or training scale.

\subsubsection{Low-scoring Case 1}

\begin{figure}[t]
    \centering
    \includegraphics[width=0.9\linewidth]{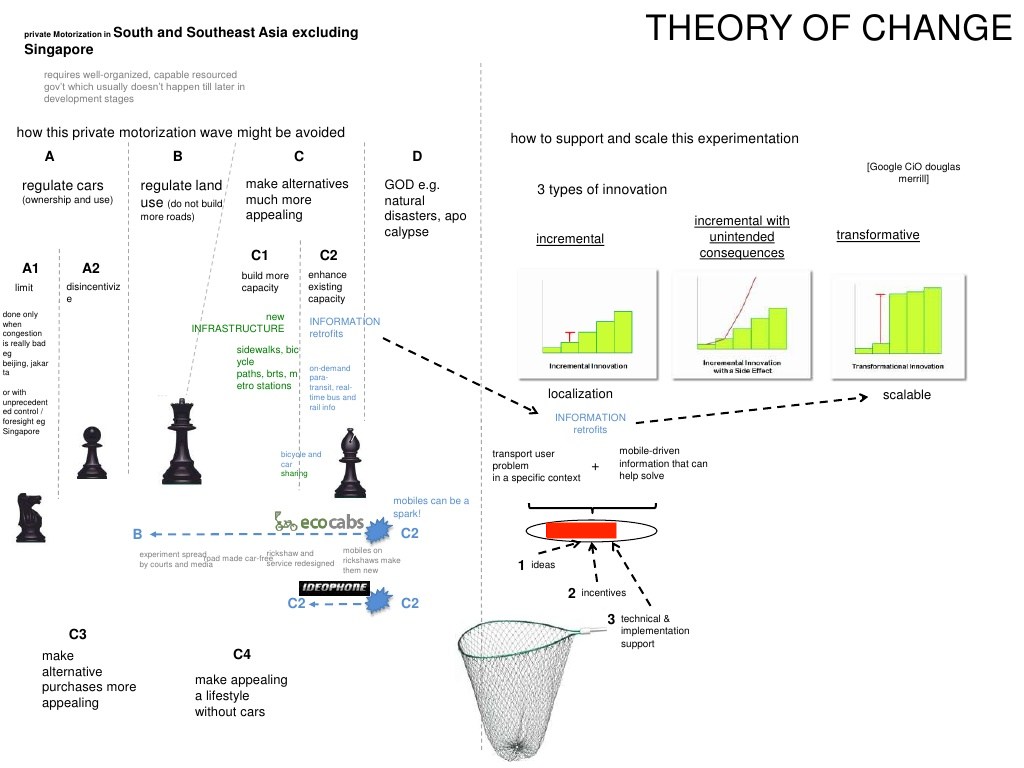}
    \caption{A representative failure case from MMTR-Bench. The masked target is \textit{entrepreneur}, but all models fail to recover the correct text and instead produce semantically related yet incorrect guesses.}
    \label{fig:bad_demo_1}
\end{figure}

\paragraph{Failure Case 1: Semantic drift under dense conceptual context.}
Figure~\ref{fig:bad_demo_1} presents a representative failure case where the ground-truth target is a concrete role noun, but all evaluated models fail to reconstruct it correctly.

\textbf{Ground truth:} \texttt{entrepreneur}

\begin{table}[H]
\centering
\small
\begin{tabular}{lp{0.60\linewidth}}
\toprule
\textbf{Model} & \textbf{Prediction} \\
\midrule
Doubao-Seed1.6-Thinking & \texttt{catalyst} \\
Doubao-Seed2-Medium & \texttt{viable local pilot solutions} \\
GPT5.2-High & \texttt{local experiments} \\
GPT5.2-Medium & \texttt{experiments} \\
GPT5.4-High & \texttt{innovation} \\
Gemini-3-Flash & \texttt{filter} \\
Gemini-3.1-Pro & \texttt{experimentation} \\
Qwen3.5-122B-A10B (nothink) & \texttt{1 ideas 2 incentives 3 technical \& implementation support} \\
Qwen3.5-122B-A10B & \texttt{localization} \\
Qwen3.5-397B-A17B (nothink) & \texttt{IDEOPHONE} \\
Qwen3.5-397B-A17B & \texttt{it} \\
\bottomrule
\end{tabular}
\caption{Model predictions for the failure case shown in Figure~\ref{fig:bad_demo_1}.}
\label{tab:failure_case_1}
\end{table}

\textbf{Analysis.}
This example is challenging because the masked target \textit{entrepreneur} is surrounded by a visually dense conceptual diagram containing many semantically related terms, such as \textit{innovation}, \textit{experimentation}, \textit{localization}, and \textit{information retrofits}. Instead of recovering the exact hidden word, most models drift toward high-level topic descriptors that are globally consistent with the figure but locally incorrect.

A clear pattern is that the models capture the \emph{theme} of the infographic but fail to identify the \emph{specific lexical item} required by the masked region. Several models generate abstract summary words, such as \textit{innovation}, \textit{experiments}, or \textit{experimentation}, suggesting reliance on global semantic gist rather than precise visual grounding. Other models are distracted by nearby visible text and directly copy salient surrounding elements, such as \textit{localization}, \textit{IDEOPHONE}, or even the enumerated support items.

This case therefore highlights a core failure mode measured by MMTR-Bench: under semantically rich yet structurally crowded multimodal context, current MLLMs often produce contextually plausible hallucinations instead of exact reconstruction. In other words, they can infer what the figure is broadly about, but still fail to determine what text is actually missing.

\newpage

\subsubsection{Low-scoring Case 2}

\begin{figure}[t]
    \centering
    \includegraphics[width=0.72\linewidth]{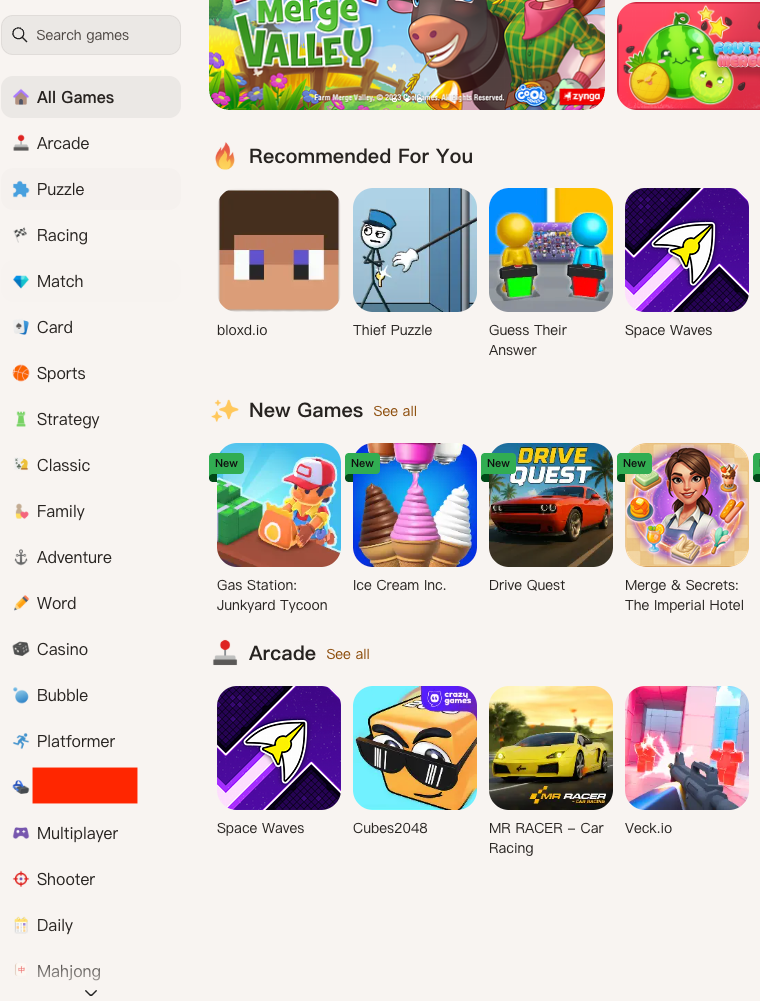}
    \caption{A representative failure case from MMTR-Bench in a UI browsing scenario. The masked target is \textit{Simulation}, but all models fail to recover the correct category label and instead predict nearby genre names or visually related concepts.}
    \label{fig:bad_demo_2}
\end{figure}

\paragraph{Failure Case 2: Neighbor-category confusion in structured UI layouts.}
Figure~\ref{fig:bad_demo_2} shows a representative failure case where the masked text corresponds to a game category label in a left-side navigation menu. Although the target is a short and common word, none of the evaluated models reconstruct it correctly.

\textbf{Ground truth:} \texttt{Simulation}

\begin{table}[H]
\centering
\small
\begin{tabular}{lp{0.58\linewidth}}
\toprule
\textbf{Model} & \textbf{Prediction} \\
\midrule
Doubao-Seed1.6-Thinking & \texttt{Action} \\
Doubao-Seed2-Medium & \texttt{Driving} \\
GPT5.2-High & \texttt{.io} \\
GPT5.2-Medium & \texttt{IO} \\
GPT5.4-High & \texttt{2 Player} \\
Gemini-3-Flash & \texttt{Driving} \\
Gemini-3.1-Pro & \texttt{Clicker} \\
Qwen3-VL-8B-Instruct & \texttt{Platformer} \\
Qwen3.5-122B-A10B (nothink) & \texttt{Racing} \\
Qwen3.5-122B-A10B & \texttt{Driving} \\
Qwen3.5-397B-A17B (nothink) & \texttt{Racing} \\
Qwen3.5-397B-A17B & \texttt{Driving} \\
\bottomrule
\end{tabular}
\caption{Model predictions for the failure case shown in Figure~\ref{fig:bad_demo_2}.}
\label{tab:failure_case_2}
\end{table}

\textbf{Analysis.}
This case differs from infographic-style failures in that the surrounding layout is highly regular and the masked text appears inside a structured navigation menu. However, the models still fail systematically. Most predictions are not random strings, but plausible category labels such as \textit{Driving}, \textit{Racing}, \textit{Platformer}, \textit{Action}, \textit{Clicker}, or \textit{2 Player}. This indicates that the models correctly identify the masked region as a game genre label, yet fail to recover the exact category name.

A notable pattern is \emph{neighbor-category confusion}. Since the masked entry is positioned between other visible menu items and is accompanied by a small car-like icon, many models are attracted to semantically nearby labels such as \textit{Driving} or \textit{Racing}. Others instead copy adjacent visible categories such as \textit{Platformer}, or generate genre terms common in game portals, such as \textit{Action} and \textit{Clicker}. The two GPT variants producing \textit{.io} and \textit{IO} further suggest that visible game titles in the main content area can interfere with label reconstruction, even when the target belongs to a different UI region.

This example highlights that MMTR-Bench is not only challenging for dense documents and conceptual diagrams, but also for seemingly simple interface screenshots. Even in clean menu layouts, current MLLMs may rely on coarse semantic association, nearby lexical copying, or icon-triggered guessing instead of precise local recovery. The failure therefore reflects a limitation in fine-grained grounding within structured UI environments, where the model must distinguish among multiple visually and semantically similar candidate labels.

\newpage

\subsubsection{Low-scoring Case 3}

\begin{figure}[t]
    \centering
    \includegraphics[width=0.88\linewidth]{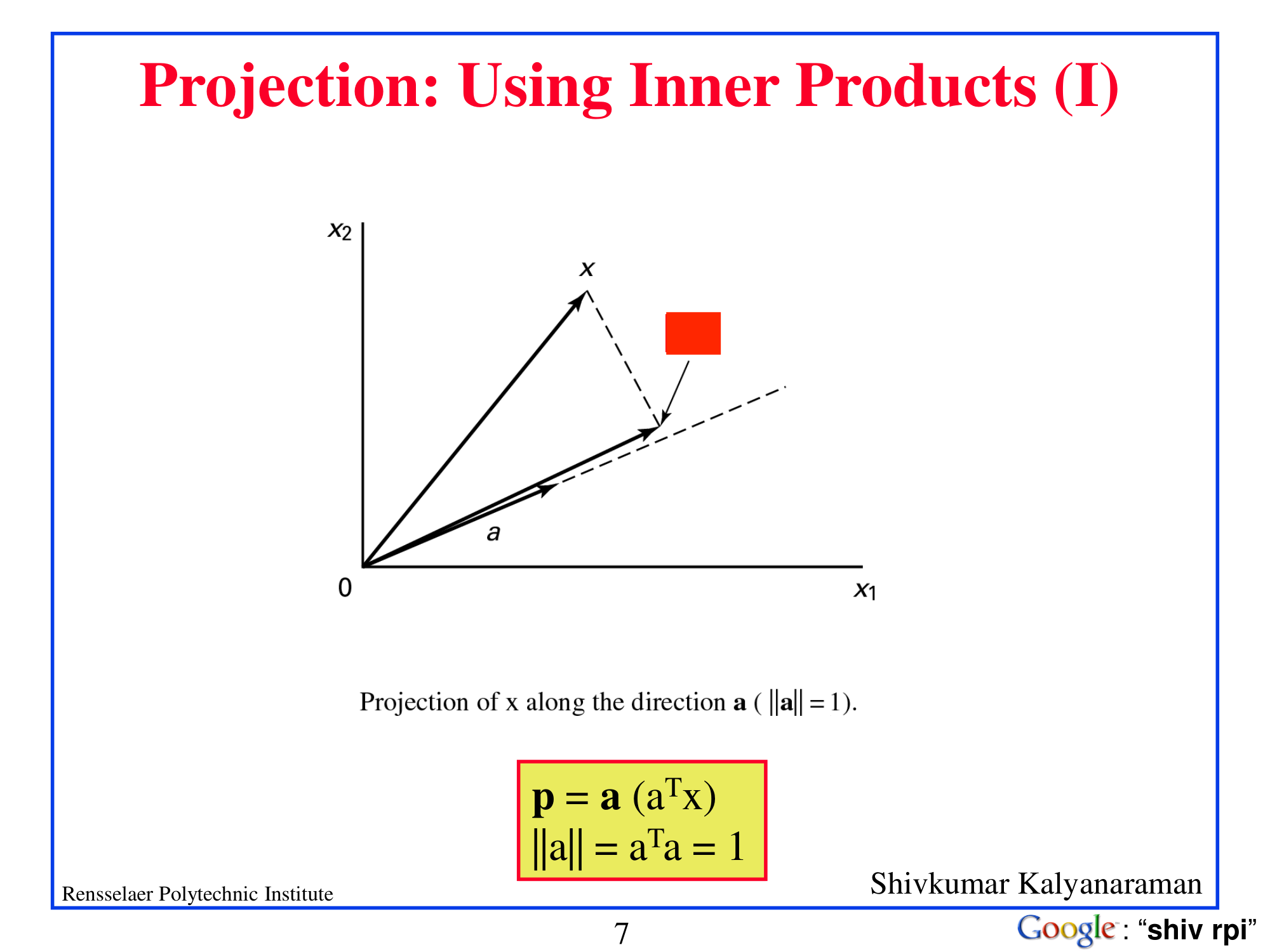}
    \caption{A representative failure case from MMTR-Bench in an educational slide containing geometric and algebraic notation. The masked target is \mbox{$x^{T}a$}, but existing models are consistently distracted by nearby visible symbols and equations, and fail to recover the correct inner-product term.}
    \label{fig:bad_demo_3}
\end{figure}

\paragraph{Failure Case 3: Symbol anchoring failure in mathematical diagrams.}
Figure~\ref{fig:bad_demo_3} presents a representative failure case where the masked target is the mathematical expression \mbox{$x^{T}a$}. Although the overall slide content clearly concerns vector projection and inner products, none of the evaluated models reconstruct the hidden formula correctly.

\textbf{Ground truth:} \mbox{$x^{T}a$}

\begin{table}[H]
\centering
\small
\begin{tabular}{lp{0.60\linewidth}}
\toprule
\textbf{Model} & \textbf{Prediction} \\
\midrule
Doubao-Seed1.6-Thinking & \texttt{p} \\
Doubao-Seed2-Medium & \texttt{p} (or the projection vector of $\mathbf{x}$ onto direction $\mathbf{a}$) \\
GPT5.2-High & \texttt{p} \\
GPT5.2-Medium & \texttt{p} \\
GPT5.4-High & \texttt{p} \\
Gemini-3-Flash & \texttt{p} \\
Gemini-3.1-Pro & \texttt{p} \\
Qwen3-VL-8B-Instruct & \texttt{Projection of x along the direction a (||a|| = 1).} \\
Qwen3.5-122B-A10B (nothink) & $p = a(a^{T}x),\ \|a\| = a^{T}a = 1$ \\
Qwen3.5-122B-A10B & \texttt{p} \\
Qwen3.5-397B-A17B (nothink) & \texttt{p} \\
Qwen3.5-397B-A17B & \texttt{p} \\
\bottomrule
\end{tabular}
\caption{Model predictions for the failure case shown in Figure~\ref{fig:bad_demo_3}.}
\label{tab:failure_case_3}
\end{table}

\textbf{Analysis.}
This case reveals a distinct failure mode in mathematically structured visual content. Rather than recovering the hidden inner-product term \mbox{$x^{T}a$}, most models collapse to the nearby visible symbol \textit{p}, which appears prominently in the boxed equation and serves as the projection vector in the slide. This suggests that the models identify the general topic of the figure, but fail to resolve which specific mathematical token is missing from the masked region.

A notable pattern is \emph{symbol anchoring to salient visible variables}. Because the slide repeatedly emphasizes \textit{p} through both the geometric diagram and the highlighted formula box, many models overfit to this highly visible symbol and substitute it for the masked expression. Other models produce even coarser responses, such as paraphrasing the slide caption or copying the full projection equation, indicating that they understand the surrounding semantics of projection but cannot isolate the exact algebraic component required by the masked location.

More broadly, this example highlights that mathematical reconstruction is not merely a special case of OCR. The correct answer requires jointly understanding the diagram, the projection formula, and the role of the hidden annotation in the figure. Current MLLMs appear able to infer the overall concept of vector projection, yet still fail at fine-grained symbolic recovery when multiple semantically related notations co-occur. MMTR-Bench therefore exposes an important gap between topic-level mathematical understanding and exact formula-level reconstruction.
\newpage

\subsubsection{Low-scoring Case 4}

\begin{figure}[t]
    \centering
    \includegraphics[width=0.82\linewidth]{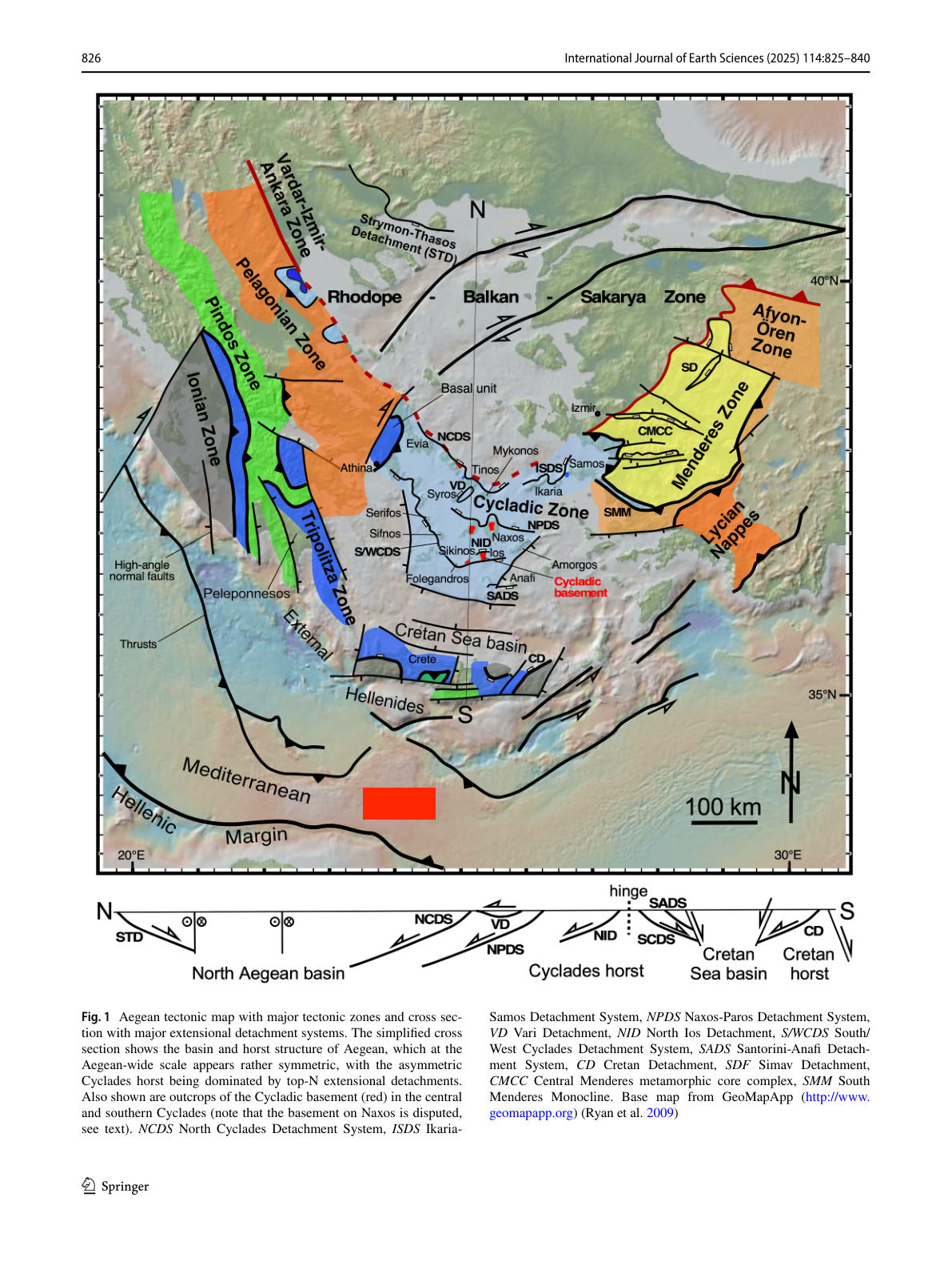}
    \caption{A representative failure case from MMTR-Bench in a scientific geologic map. The masked target is \textit{Ridge}, but current models are distracted by nearby place names, geographic entities, and document-level text, leading to systematic reconstruction errors.}
    \label{fig:bad_demo_4}
\end{figure}

\paragraph{Failure Case 4: Geographic mislocalization in dense scientific maps.}
Figure~\ref{fig:bad_demo_4} presents a representative failure case where the masked target is the short geographic term \textit{Ridge}. Despite the target being a single common word, none of the evaluated models recovers it correctly.

\textbf{Ground truth:} \texttt{Ridge}

\begin{table}[H]
\centering
\small
\begin{tabular}{lp{0.60\linewidth}}
\toprule
\textbf{Model} & \textbf{Prediction} \\
\midrule
Doubao-Seed1.6-Thinking & \texttt{No masked text regions were identified in the provided document images.} \\
Doubao-Seed2-Medium & \texttt{Africa} \\
GPT5.2-High & \texttt{25 E} \\
GPT5.2-Medium & \texttt{Libyan Sea} \\
GPT5.4-High & \texttt{Crete} \\
Gemini-3-Flash & \texttt{Libyan Sea} \\
Gemini-3.1-Pro & \texttt{Pelagonian Zone} \\
Qwen3-VL-8B-Instruct & \texttt{The provided document does not contain any images where text regions are masked out with black boxes.} \\
Qwen3.5-122B-A10B (nothink) & \texttt{[long document-level extraction unrelated to the masked region]} \\
\bottomrule
\end{tabular}
\caption{Model predictions for the failure case shown in Figure~\ref{fig:bad_demo_4}.}
\label{tab:failure_case_4}
\end{table}

\textbf{Analysis.}
This case highlights a characteristic failure mode in scientific maps: although the masked target is local and relatively short, the surrounding visual field is crowded with many competing labels, including tectonic zones, place names, seas, coordinates, and figure-caption text. As a result, the models do not recover the missing word \textit{Ridge}, but instead output other geographically plausible strings such as \textit{Libyan Sea}, \textit{Crete}, \textit{Pelagonian Zone}, or even a coordinate marker such as \textit{25 E}. This indicates that the models roughly recognize the input as a geographic map, but fail to localize the exact missing label.

A notable pattern here is \emph{geographic mislocalization}. Rather than grounding prediction on the masked area itself, several models appear to select nearby or globally salient map entities. In other words, they retrieve a plausible \emph{type} of answer---a place name, tectonic unit, or map annotation---but not the correct one. This suggests that the models are influenced more by regional semantic context than by the precise local evidence needed for exact reconstruction.

An additional failure mode is \emph{document-level override}. Some models do not even attempt local reconstruction, instead claiming that no mask is present or producing long excerpts from the accompanying paper. This behavior is especially revealing because it shows that the model may completely lose track of the masked target once the page contains both a complex figure and dense surrounding academic text. The error is therefore not only about wrong lexical choice, but also about a breakdown in region selection and task focus.

More broadly, this example shows that map-based reconstruction is difficult for current MLLMs even when the hidden string is short. Accurate recovery requires distinguishing among many neighboring labels of the same semantic type, while also ignoring irrelevant but highly salient text elsewhere on the page. MMTR-Bench therefore captures an important gap between coarse scene understanding of scientific figures and precise localized text recovery in map-heavy document images.
\newpage
\subsubsection{Low-scoring Case 5}

\begin{figure}[t]
    \centering
    \includegraphics[width=0.88\linewidth]{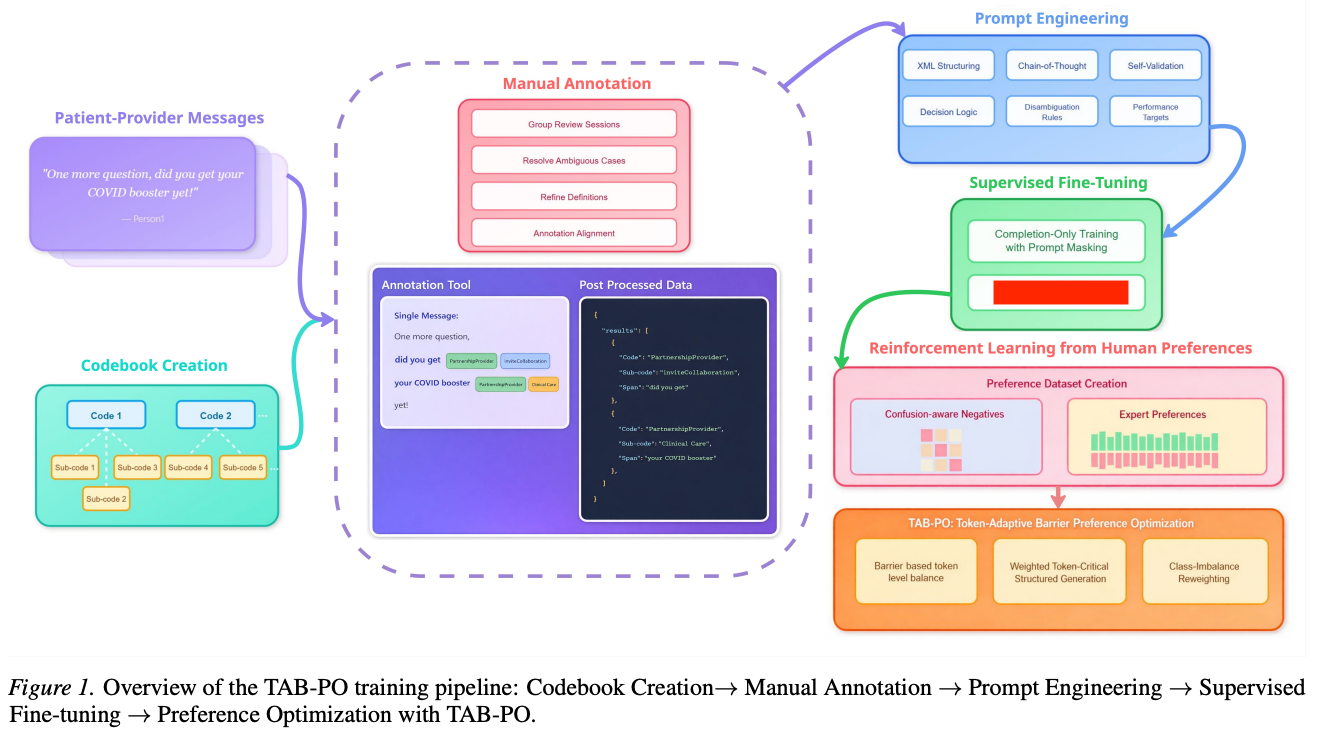}
    \caption{A representative failure case from MMTR-Bench in a research pipeline figure. The masked target is \textit{Length Normalization}, but current models systematically favor nearby visible phrases and high-level training terminology instead of the exact hidden module name.}
    \label{fig:bad_demo_5}
\end{figure}

\paragraph{Failure Case 5: Visible-text copying in modular pipeline diagrams.}
Figure~\ref{fig:bad_demo_5} shows a representative failure case where the masked target is the phrase \textit{Length Normalization}. Although the masked region lies inside a clean and highly structured pipeline figure, none of the evaluated models reconstructs the correct text.

\textbf{Ground truth:} \texttt{Length Normalization}

\begin{table}[H]
\centering
\small
\begin{tabular}{lp{0.60\linewidth}}
\toprule
\textbf{Model} & \textbf{Prediction} \\
\midrule
Doubao-Seed1.6-Thinking & \texttt{Instruction Fine-Tuning} \\
Doubao-Seed2-Medium & \texttt{Structured Generation Fine-Tuning} \\
GPT5.2-High & \texttt{Parameter-Efficient Fine-Tuning with QLoRA} \\
GPT5.2-Medium & \texttt{Parameter-Efficient Fine-Tuning (QLoRA)} \\
GPT5.4-High & \texttt{Low-Rank Adaptation (LoRA)} \\
Gemini-3-Flash & \texttt{SFT Model} \\
Gemini-3.1-Pro & \texttt{Structured JSON Generation} \\
Qwen3-VL-8B-Instruct & \texttt{Completion-Only Training with Prompt Masking} \\
Qwen3.5-122B-A10B (nothink) & \texttt{Completion-Only Training with Prompt Masking} \\
Qwen3.5-122B-A10B & \texttt{Completion-Only Training with Prompt Masking} \\
Qwen3.5-397B-A17B (nothink) & \texttt{Completion-Only Training with Prompt Masking} \\
Qwen3.5-397B-A17B & \texttt{Completion-Only Training with Prompt Masking} \\
\bottomrule
\end{tabular}
\caption{Model predictions for the failure case shown in Figure~\ref{fig:bad_demo_5}.}
\label{tab:failure_case_5}
\end{table}

\textbf{Analysis.}
This case highlights a different error pattern from maps or mathematical slides. The surrounding figure is visually clean, modular, and semantically well organized, yet the models still fail completely. Rather than recovering the hidden phrase \textit{Length Normalization}, most predictions collapse to other training-related expressions that are either explicitly visible in the same green module or strongly associated with supervised fine-tuning, such as \textit{Completion-Only Training with Prompt Masking}, \textit{Instruction Fine-Tuning}, \textit{LoRA}, or \textit{QLoRA}.

The dominant failure mode here is \emph{visible-text copying plus semantic substitution}. Several models directly copy the most salient nearby phrase in the same panel, namely \textit{Completion-Only Training with Prompt Masking}, while others generate plausible fine-tuning terminology that fits the topic of the figure but is not grounded in the masked region itself. This indicates that the models identify the correct semantic domain---LLM training pipelines and supervised fine-tuning---but fail to resolve which specific subcomponent is being occluded.

More broadly, this example shows that structured infographic layouts do not necessarily make reconstruction easy. Even when the figure is neatly partitioned into modules, the presence of multiple semantically compatible labels can cause models to over-rely on topical consistency instead of exact local recovery. MMTR-Bench therefore exposes a gap between understanding the overall pipeline and reconstructing the precise hidden module name.

\newpage

\subsection{Effect of Explicit Reasoning on Masked Text Reconstruction}
\label{sec:thinking_vs_nonthinking}

To better understand whether explicit reasoning improves masked text reconstruction, we further compare thinking and non-thinking variants on representative MMTR-Bench examples. We find that reasoning does not lead to a uniform gain. In some cases, it helps the model exploit local structure or integrate distributed semantic cues, while in other cases it may encourage broader but less grounded inference. These observations suggest that the value of explicit reasoning is highly dependent on the type of evidence required for recovering the masked content.

\begin{figure}[t]
    \centering
    \includegraphics[width=0.82\linewidth]{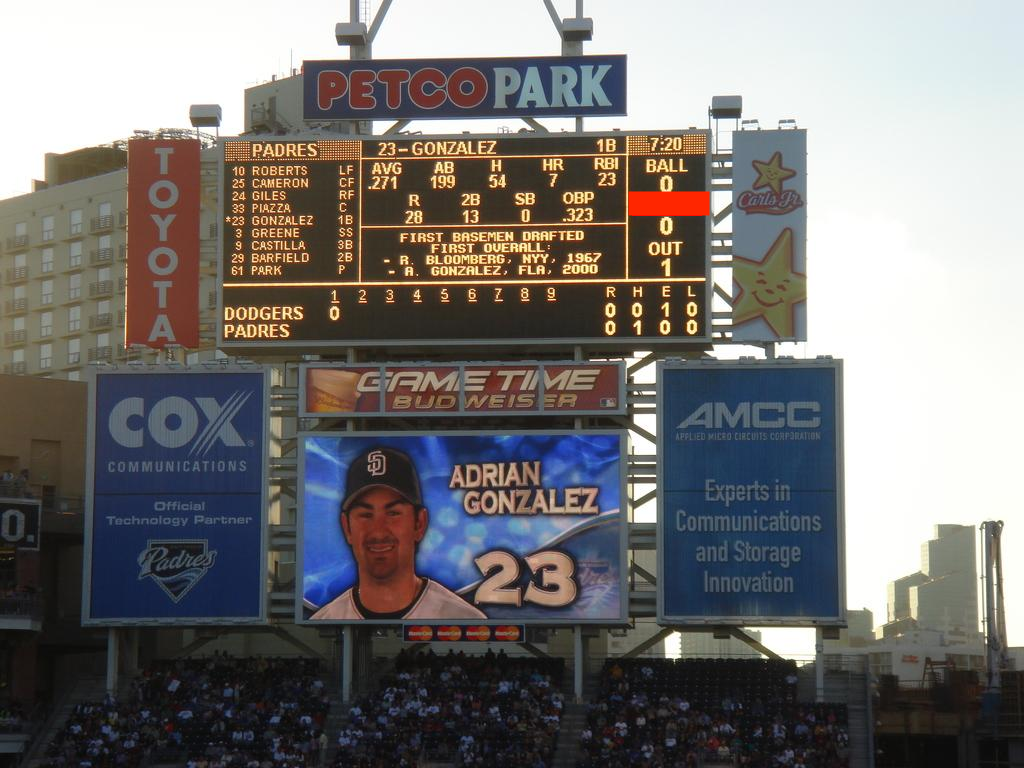}
    \caption{A comparison case for thinking and non-thinking variants. The masked target is \textit{STRIKE}. The surrounding scoreboard provides a strong structural template (\textit{BALL--STRIKE--OUT}), making the hidden label recoverable through local relational reasoning.}
    \label{fig:thinking_demo_1}
\end{figure}

\paragraph{Case 1: Reasoning helps when the target is structurally constrained.}
Figure~\ref{fig:thinking_demo_1} compares the thinking and non-thinking variants on a scoreboard example where the masked target is \textit{STRIKE}.

\textbf{Ground truth:} \texttt{STRIKE}

\begin{table}[H]
\centering
\small
\begin{tabular}{lll}
\toprule
\textbf{Model Family} & \textbf{Non-thinking} & \textbf{Thinking} \\
\midrule
Qwen3.5-122B-A10B & \texttt{0} & \texttt{STRIKE} \\
\bottomrule
\end{tabular}
\caption{Comparison between thinking and non-thinking variants for the case shown in Figure~\ref{fig:thinking_demo_1}.}
\label{tab:thinking_case_1}
\end{table}

\textbf{Analysis.}
This example shows a case where explicit reasoning is genuinely beneficial. The non-thinking variant appears to anchor on the most immediate local token and outputs the nearby count value \textit{0}, indicating shallow pattern matching without resolving the functional role of the masked text. By contrast, the thinking variant successfully infers the latent scoreboard schema, namely \textit{BALL--STRIKE--OUT}, and reconstructs the missing label correctly.

This case suggests that reasoning can improve exact recovery when the surrounding visual context provides a compact and low-ambiguity structural template. In such settings, the advantage of thinking does not come from broader semantic extrapolation, but from identifying a stable local relation and completing it correctly.

\textbf{Abbreviated reasoning trace (thinking variant).}
The model first identifies the masked region as part of the scoreboard count panel. It then infers that the surrounding labels form the conventional baseball structure \textit{BALL--STRIKE--OUT}. Based on this local structural template, it reconstructs the hidden text as \textit{STRIKE}.

\begin{figure}[t]
    \centering
    \includegraphics[width=0.72\linewidth]{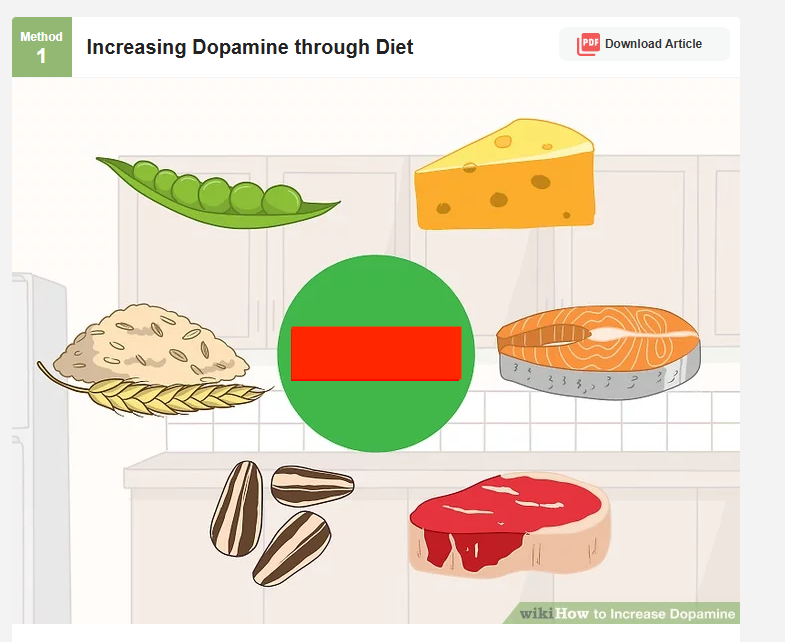}
    \caption{A comparison case for thinking and non-thinking variants on a wikiHow-style infographic. The masked target is \textit{Tyrosine}. The non-thinking variant copies a salient nearby visible label (\textit{Method 1}), whereas the thinking variant correctly infers the hidden concept from the semantic consistency of the depicted food items.}
    \label{fig:thinking_demo_2}
\end{figure}

\paragraph{Case 2: Reasoning helps when semantic integration is required.}
Figure~\ref{fig:thinking_demo_2} compares the thinking and non-thinking variants on an infographic where the masked target is \textit{Tyrosine}.

\textbf{Ground truth:} \texttt{Tyrosine}

\begin{table}[H]
\centering
\small
\begin{tabular}{lll}
\toprule
\textbf{Model Family} & \textbf{Non-thinking} & \textbf{Thinking} \\
\midrule
Qwen3.5-397B-A17B & \texttt{Method 1} & \texttt{Tyrosine} \\
\bottomrule
\end{tabular}
\caption{Comparison between thinking and non-thinking variants for the case shown in Figure~\ref{fig:thinking_demo_2}.}
\label{tab:thinking_case_2}
\end{table}

\textbf{Analysis.}
This example shows a second setting in which explicit reasoning is beneficial, but for a different reason from the scoreboard case. The non-thinking variant produces \textit{Method 1}, which is a highly visible label located in the upper-left corner of the image. This suggests a shallow strategy based on copying a salient visible token without identifying the semantic role of the masked central region. By contrast, the thinking variant correctly recovers \textit{Tyrosine}, indicating that it is able to integrate the broader semantic context of the infographic rather than relying only on the most visually prominent text.

The key difference is that this case is not governed by a rigid local template such as \textit{BALL--STRIKE--OUT}. Instead, successful reconstruction requires semantic aggregation: the model must connect the title \textit{Increasing Dopamine through Diet} with the depicted foods, including cheese, salmon, seeds, peas, grains, and meat, and infer that the hidden concept is the dopamine-related nutrient shared by these examples. In this sense, the thinking variant succeeds by synthesizing multiple weak contextual cues into a coherent concept, whereas the non-thinking variant fails by anchoring on a single superficial visible label.

\textbf{Abbreviated reasoning trace (thinking variant).}
The model recognizes that the figure is about increasing dopamine through diet and notes that the depicted foods are commonly associated with tyrosine-rich diets. It integrates these distributed semantic cues and infers that the masked central concept is \textit{Tyrosine}. Unlike the non-thinking variant, it does not simply copy the most salient visible label.

\textbf{Discussion.}
Taken together, these examples suggest that explicit reasoning can help masked text reconstruction in at least two distinct regimes. In one regime, it helps by exploiting a strong local structural schema; in another, it helps by combining distributed semantic evidence across the whole image. This indicates that the effect of thinking is neither uniformly positive nor uniformly negative. Rather, its usefulness depends on whether the masked content can be recovered through stable local relations or through coherent multi-cue semantic integration.

\begin{figure}[t]
    \centering
    \includegraphics[width=0.88\linewidth]{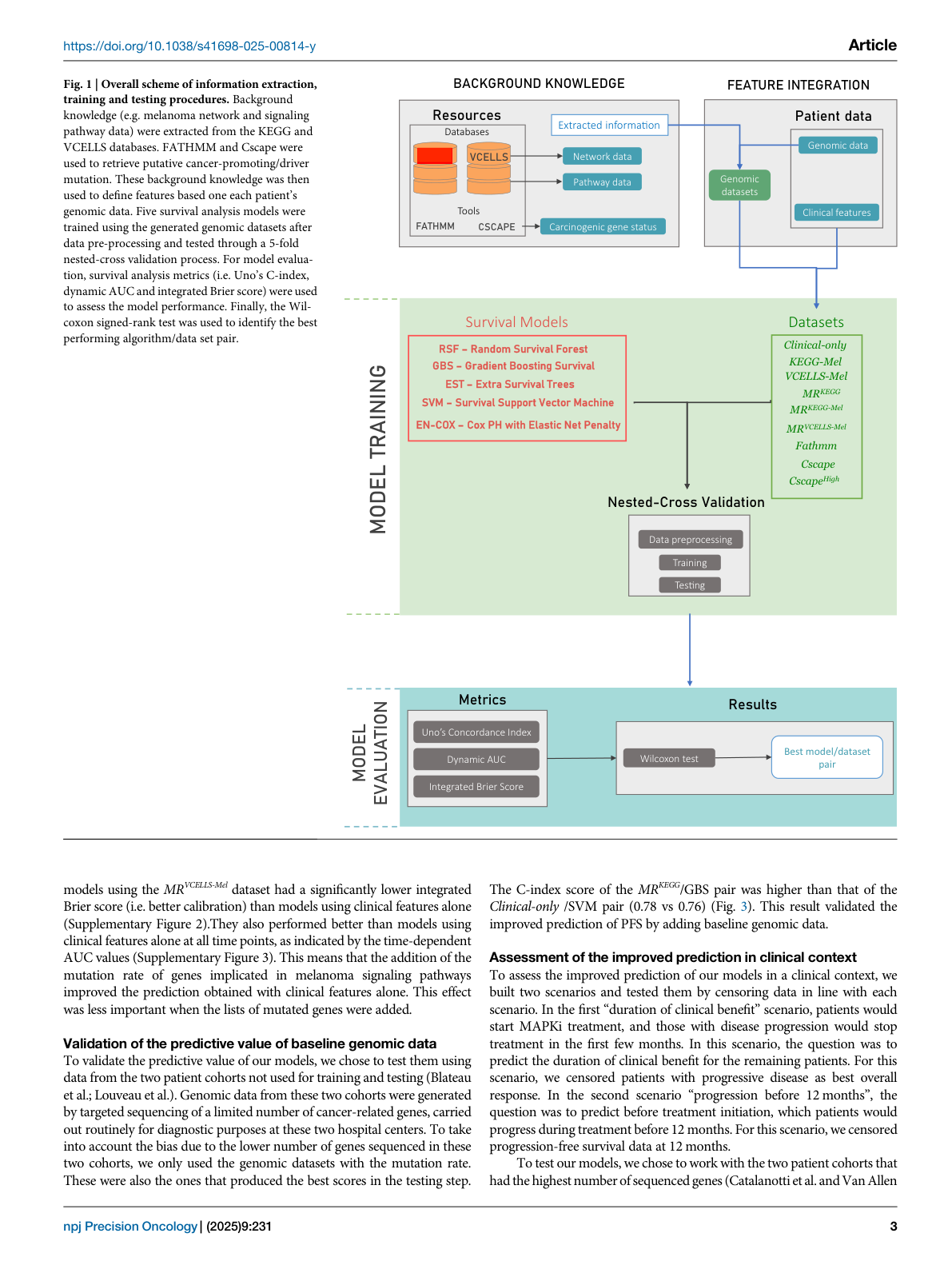}
    \caption{A comparison case for thinking and non-thinking variants on a scientific figure embedded in a document page. The masked target is \textit{KEGG}. The non-thinking variant fails to even recognize the existence of the masked region, whereas the thinking variant correctly recovers the hidden label by grounding the black box inside the ``Resources'' submodule of the diagram.}
    \label{fig:thinking_demo_3}
\end{figure}

\paragraph{Case 3: Reasoning helps recover task focus in document figures.}
Figure~\ref{fig:thinking_demo_3} compares the thinking and non-thinking variants on a scientific document page where the masked target is \textit{KEGG}.

\textbf{Ground truth:} \texttt{KEGG}

\begin{table}[H]
\centering
\small
\begin{tabular}{lp{0.28\linewidth}p{0.28\linewidth}}
\toprule
\textbf{Model Family} & \textbf{Non-thinking} & \textbf{Thinking} \\
\midrule
Qwen3.5-397B-A17B &
\texttt{No masked text regions were identified...} &
\texttt{KEGG} \\
\bottomrule
\end{tabular}
\caption{Comparison between thinking and non-thinking variants for the case shown in Figure~\ref{fig:thinking_demo_3}.}
\label{tab:thinking_case_3}
\end{table}

\textbf{Analysis.}
This example reveals a failure mode that is different from both the scoreboard and infographic cases. Here, the non-thinking variant does not merely predict the wrong word; instead, it fails at an earlier stage and incorrectly concludes that no masked text is present. In other words, the error arises before lexical reconstruction, at the level of region selection and task grounding. By contrast, the thinking variant successfully identifies the black box within the diagram and reconstructs the hidden label as \textit{KEGG}.

The difficulty of this example comes from the page composition. The masked target is embedded inside a relatively small subregion of a complex scientific figure, while the surrounding page also contains dense caption text, paragraph text, and multiple other labeled boxes. Under this setting, a shallow pass can easily be distracted by the document as a whole and miss the local masked area entirely. The thinking variant appears to recover performance by explicitly narrowing attention to the figure, then to the ``Resources'' module, and finally to the pair of database labels, one of which remains visible as \textit{VCELLS}. This makes the missing companion label recoverable as \textit{KEGG}.

This case suggests that explicit reasoning can help not only with semantic or structural completion, but also with \emph{task focusing}. Before reconstructing the hidden text, the model must first determine \emph{where} the masked region is and \emph{which} surrounding evidence is relevant. In document-style pages that mix figures with long captions and body text, this intermediate focusing step can be crucial. The comparison therefore highlights an additional benefit of thinking: it can reduce failures caused by missing the target region altogether.

\textbf{Abbreviated reasoning trace (thinking variant).}
The model first localizes the black box inside the figure rather than in the surrounding body text. It then identifies the masked region as part of the ``Resources'' block containing two database labels, one of which is still visible as \textit{VCELLS}. Using the figure caption and the symmetry of the paired database icons, it infers that the hidden companion label is \textit{KEGG}.

\begin{figure}[t]
    \centering
    \includegraphics[width=0.90\linewidth]{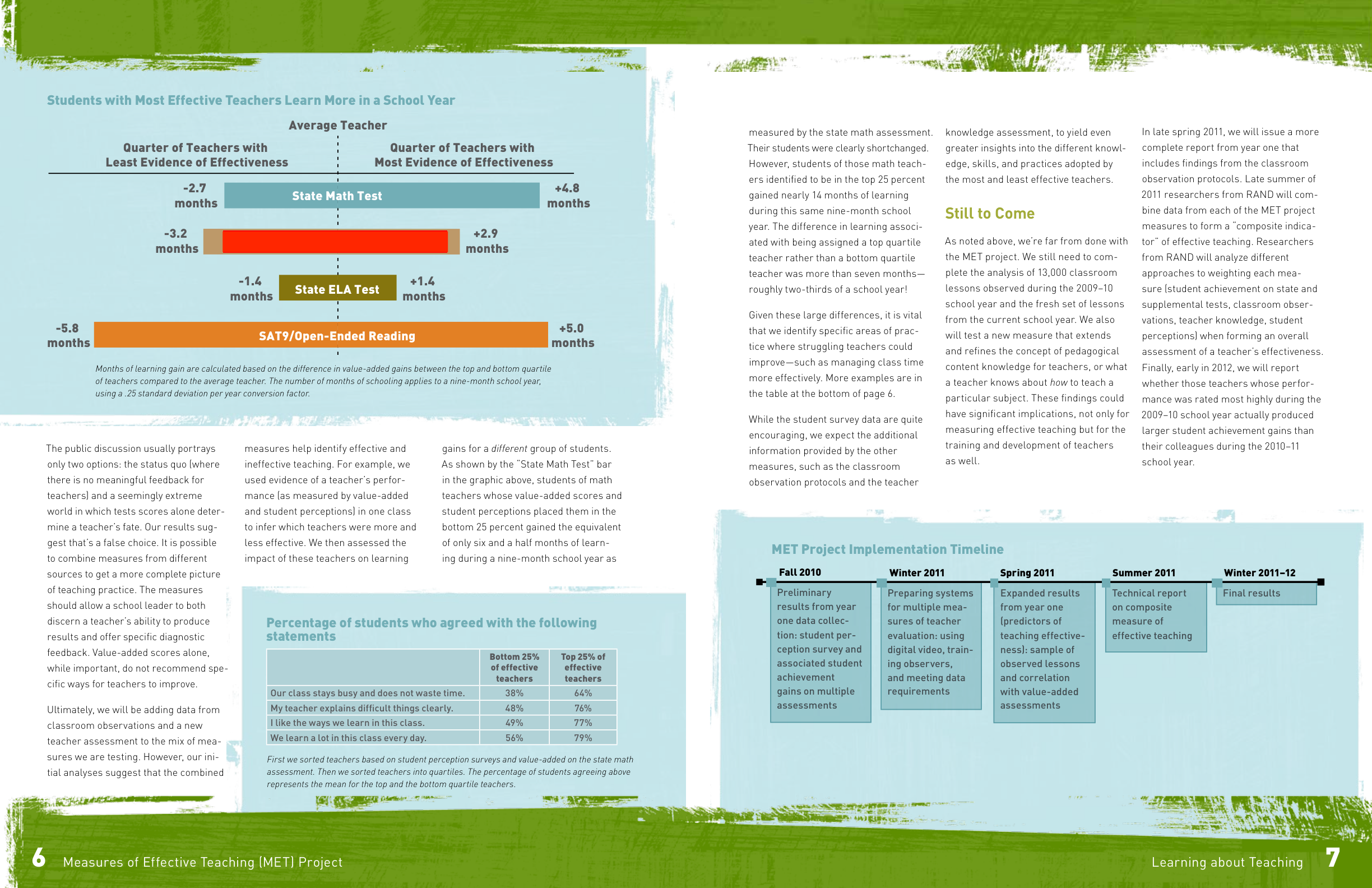}
    \caption{A comparison case for thinking and non-thinking variants on a document page containing a masked chart label. The masked target is \textit{Balanced Assessment of Mathematics}. The non-thinking variant anchors on the nearby numeric annotation, while the thinking variant identifies the hidden bar label by integrating evidence from the chart layout and the surrounding explanatory text.}
    \label{fig:thinking_demo_4}
\end{figure}

\paragraph{Case 4: Reasoning helps disambiguate label reconstruction from nearby numeric evidence.}
Figure~\ref{fig:thinking_demo_4} compares the thinking and non-thinking variants on a document page where the masked target is \textit{Balanced Assessment of Mathematics}.

\textbf{Ground truth:} \texttt{Balanced Assessment of Mathematics}

\begin{table}[H]
\centering
\small
\begin{tabular}{lp{0.24\linewidth}p{0.24\linewidth}}
\toprule
\textbf{Model Family} & \textbf{Non-thinking} & \textbf{Thinking} \\
\midrule
Qwen3.5-397B-A17B &
\texttt{-3.2 months} &
\texttt{Balanced Assessment of Mathematics} \\
\bottomrule
\end{tabular}
\caption{Comparison between thinking and non-thinking variants for the case shown in Figure~\ref{fig:thinking_demo_4}.}
\label{tab:thinking_case_4}
\end{table}

\textbf{Analysis.}
This example shows another setting in which explicit reasoning is beneficial, but here the main challenge is disambiguating the \emph{type} of missing content. The non-thinking variant outputs \textit{-3.2 months}, which is a nearby visible numeric annotation associated with the masked row. This suggests that it correctly localizes the approximate region of interest, but fails to determine whether the hidden content is a label, a value, or another graphical element. By contrast, the thinking variant correctly reconstructs the hidden test name \textit{Balanced Assessment of Mathematics}, indicating that it is able to infer the functional role of the masked span within the chart.

The figure contains multiple competing textual elements: bar labels, left and right month values, section headings, and long explanatory paragraphs below and to the right. Under such conditions, shallow local matching is insufficient because the closest visible evidence includes both the masked test label and the adjacent numeric values. The non-thinking variant appears to latch onto the most immediately available number, whereas the thinking variant uses the overall bar-chart schema and the neighboring named rows such as \textit{State Math Test}, \textit{State ELA Test}, and \textit{SAT9/Open-Ended Reading} to infer that the hidden row should also be a test label rather than a measurement.

This case also shows that successful reasoning may require combining \emph{visual structure} with \emph{document context}. The hidden row corresponds to a supplemental mathematics assessment, and the full phrase becomes recoverable only when the model connects the chart organization with the surrounding discussion of math and reading assessments. In this sense, the benefit of thinking is not just better localization, but better role assignment: it helps the model decide what kind of information is missing before attempting reconstruction.

\textbf{Abbreviated reasoning trace (thinking variant).}
The model first identifies the masked region as the title of the second bar rather than one of the nearby month values. It then uses the neighboring bar labels and the surrounding article text about supplemental mathematics assessments to infer that the hidden row corresponds to \textit{Balanced Assessment of Mathematics}. By recognizing the masked span as a chart label instead of a numeric annotation, it reconstructs the correct phrase.

\section{Case Studies}
\label{app:case_studies}

In this section, we present 32 case studies from MMTR-Bench to provide a clearer view of the dataset's diversity. 

\begin{figure*}[p]
    \centering
    \includegraphics[width=1\textwidth]{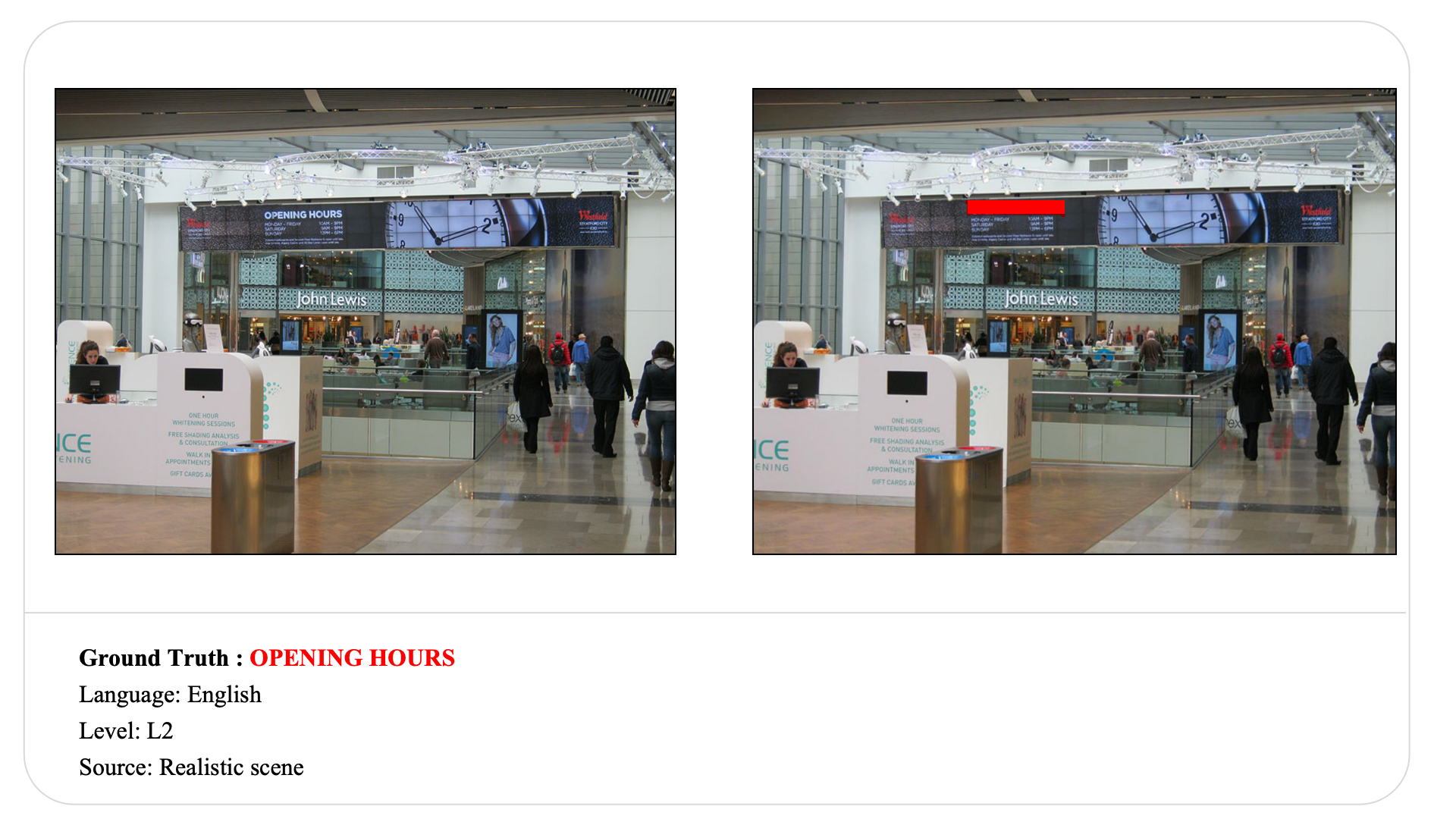}
    \caption{Case 1 from MMTR-Bench.}
    
    \vspace{1cm}
    
    \includegraphics[width=1\textwidth]{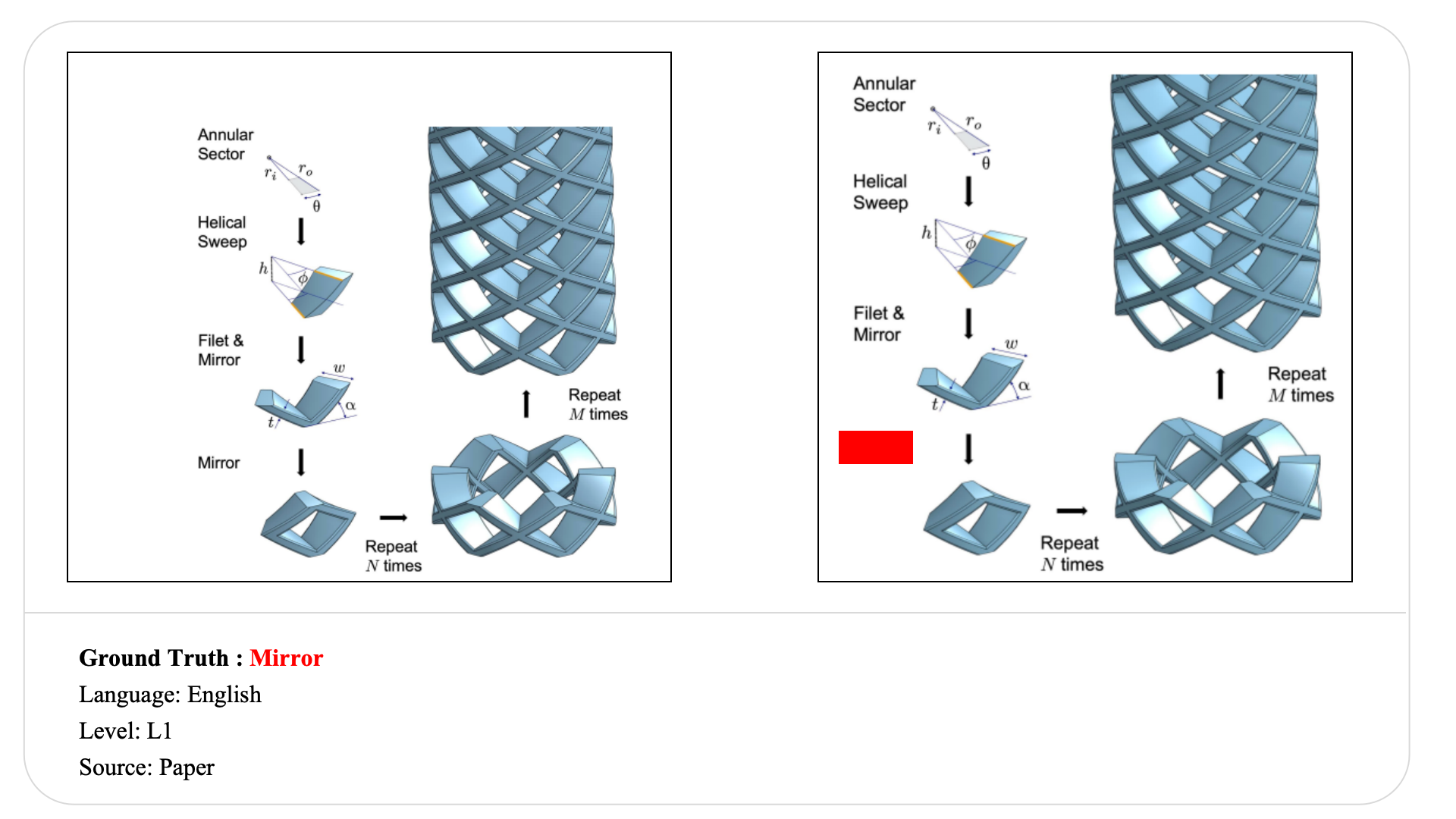}
    \caption{Case 2 from MMTR-Bench.}
\end{figure*}

\begin{figure*}[p]
    \centering
    \includegraphics[width=1\textwidth]{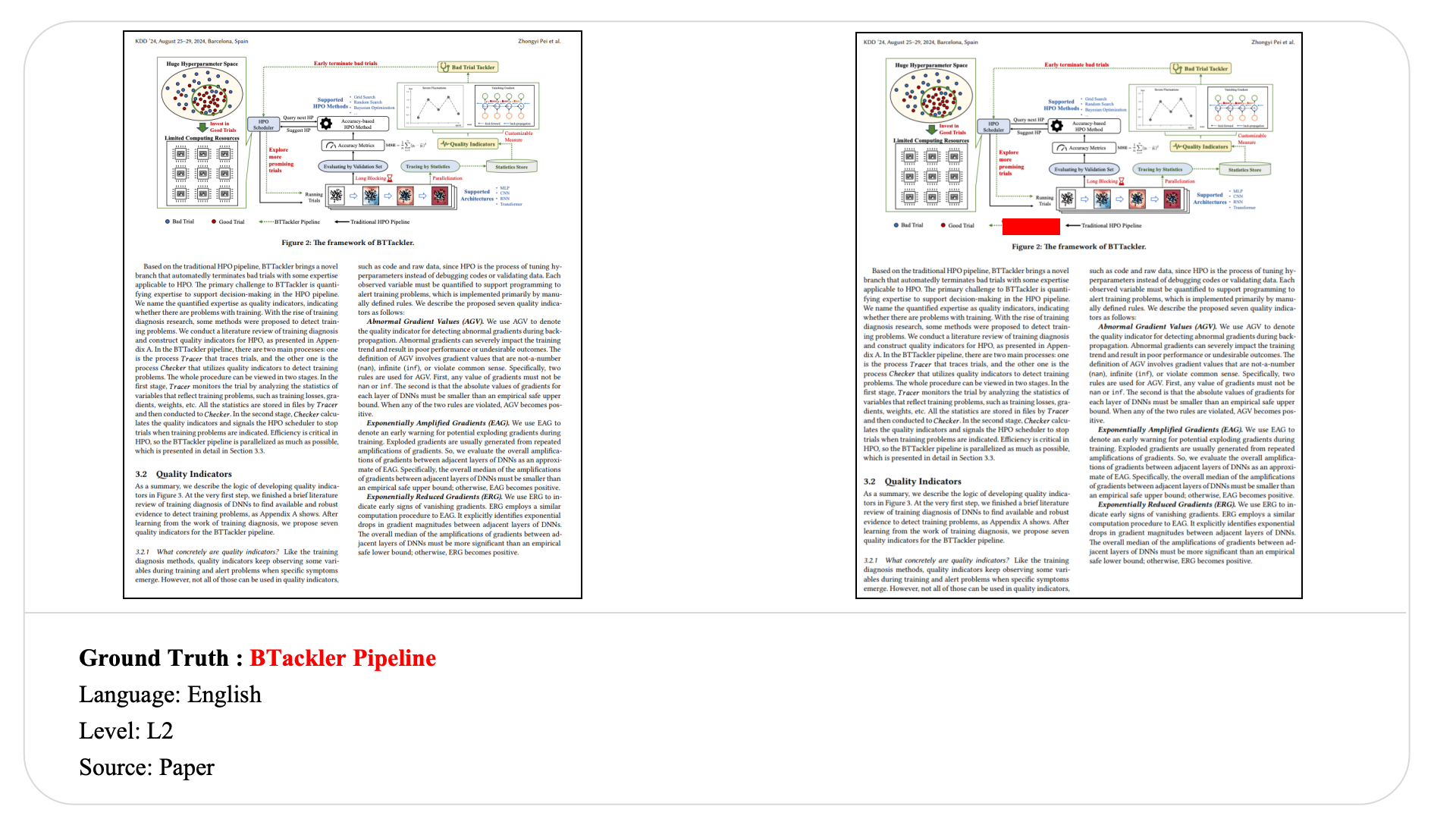}
    \caption{Case 3 from MMTR-Bench.}
    
    \vspace{1cm}
    
    \includegraphics[width=1\textwidth]{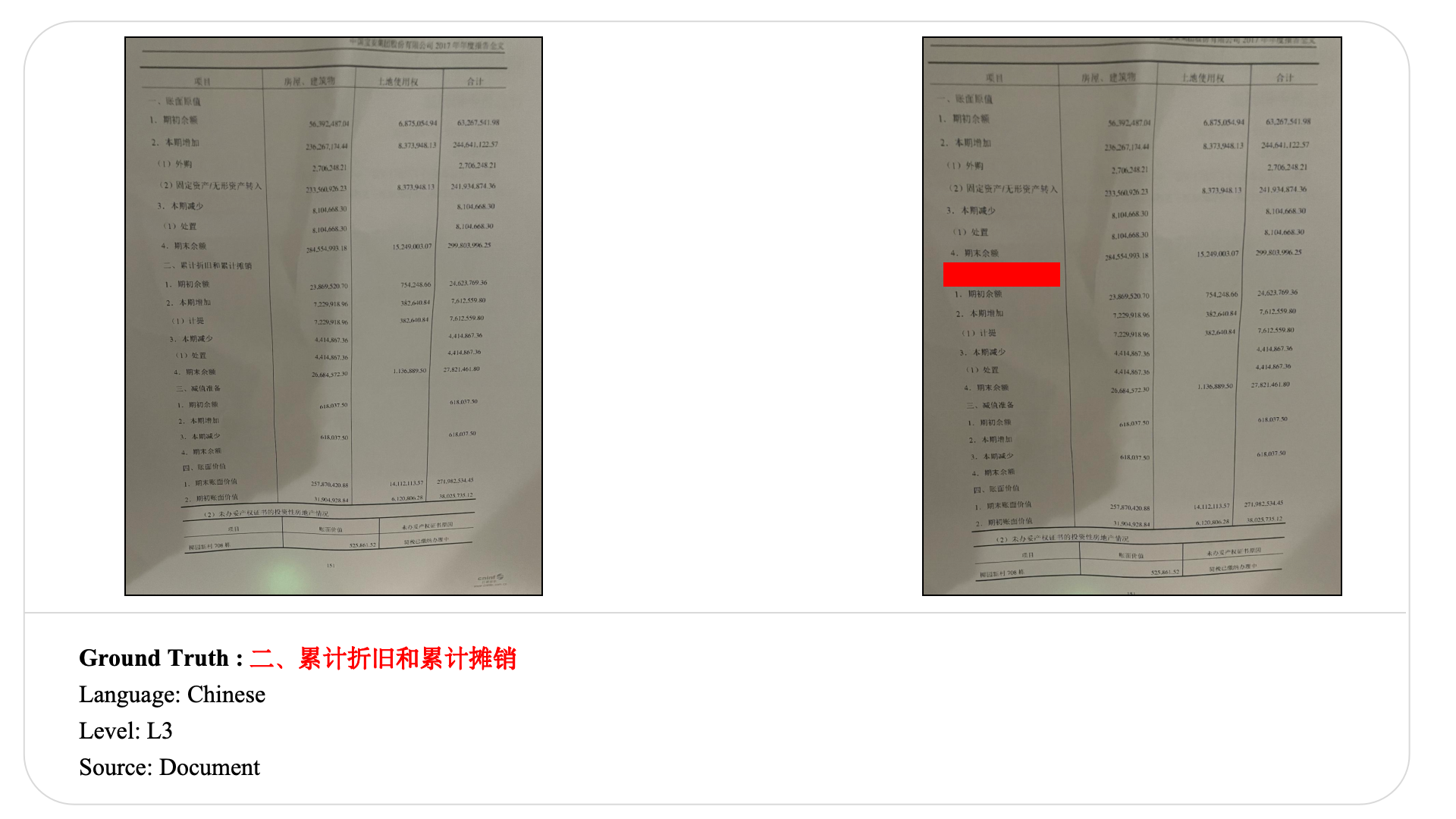}
    \caption{Case 4 from MMTR-Bench.}
\end{figure*}

\begin{figure*}[p]
    \centering
    \includegraphics[width=1\textwidth]{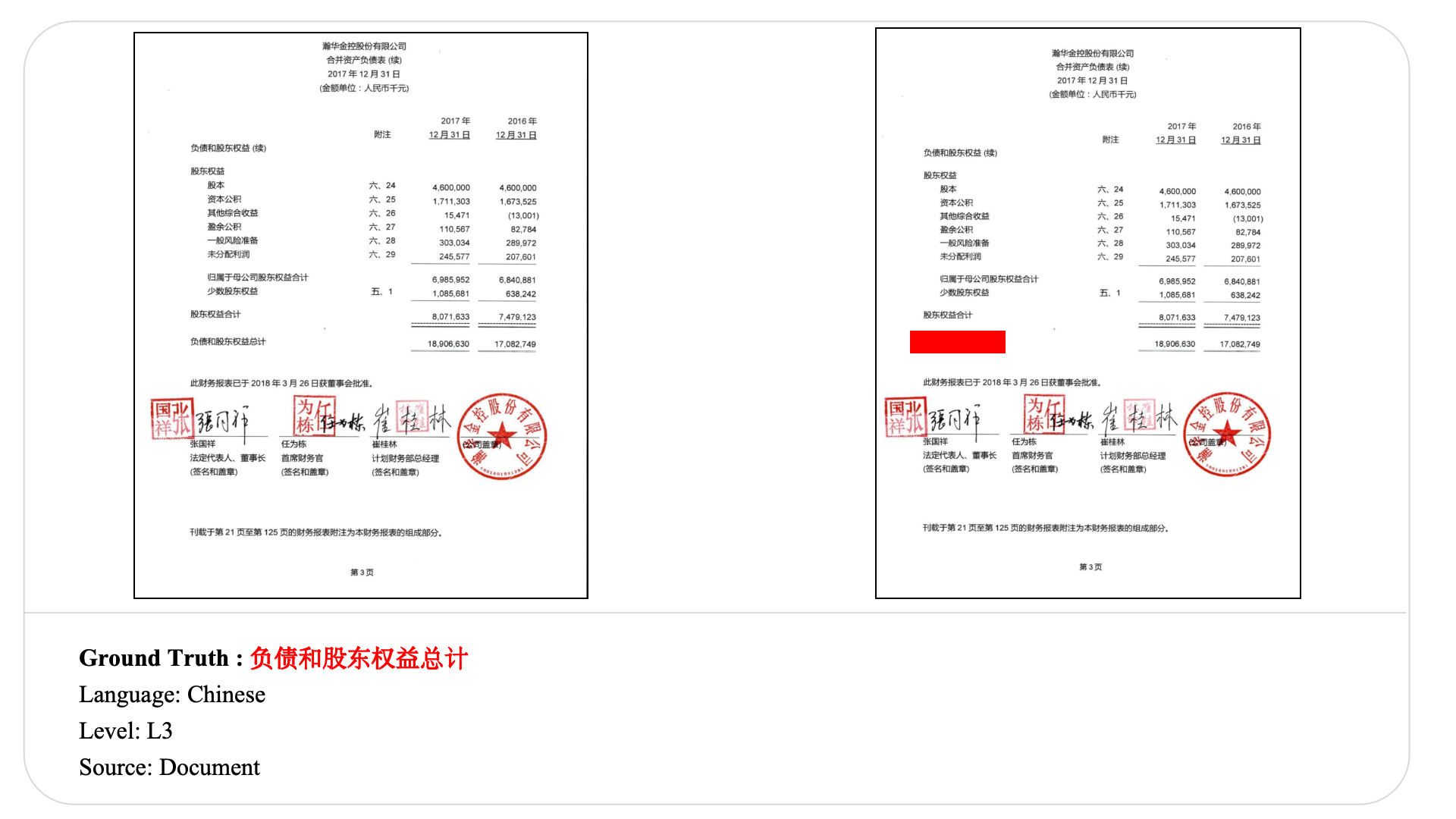}
    \caption{Case 5 from MMTR-Bench.}
    
    \vspace{1cm}
    
    \includegraphics[width=1\textwidth]{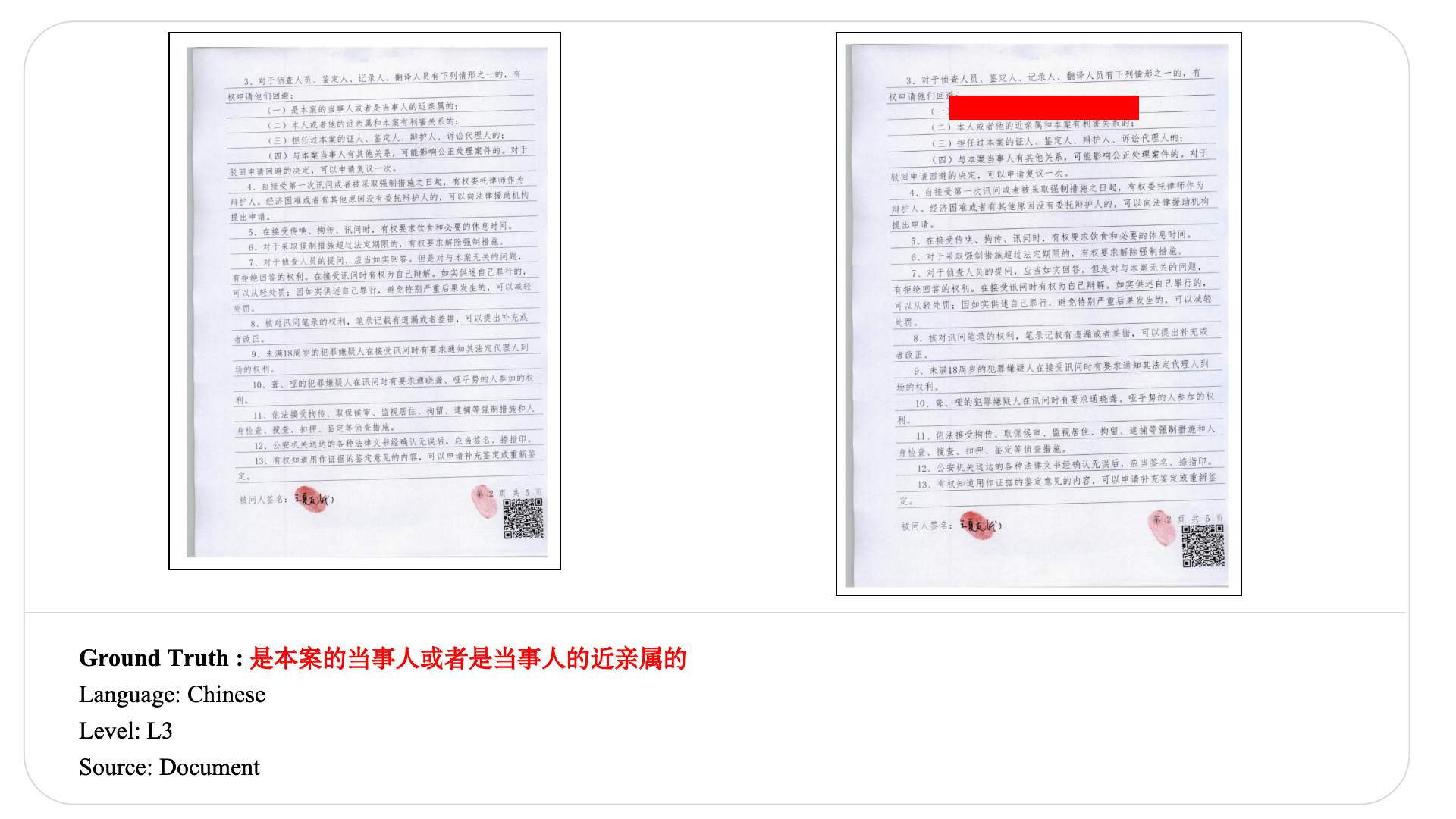}
    \caption{Case 6 from MMTR-Bench.}
\end{figure*}

\begin{figure*}[p]
    \centering
    \includegraphics[width=1\textwidth]{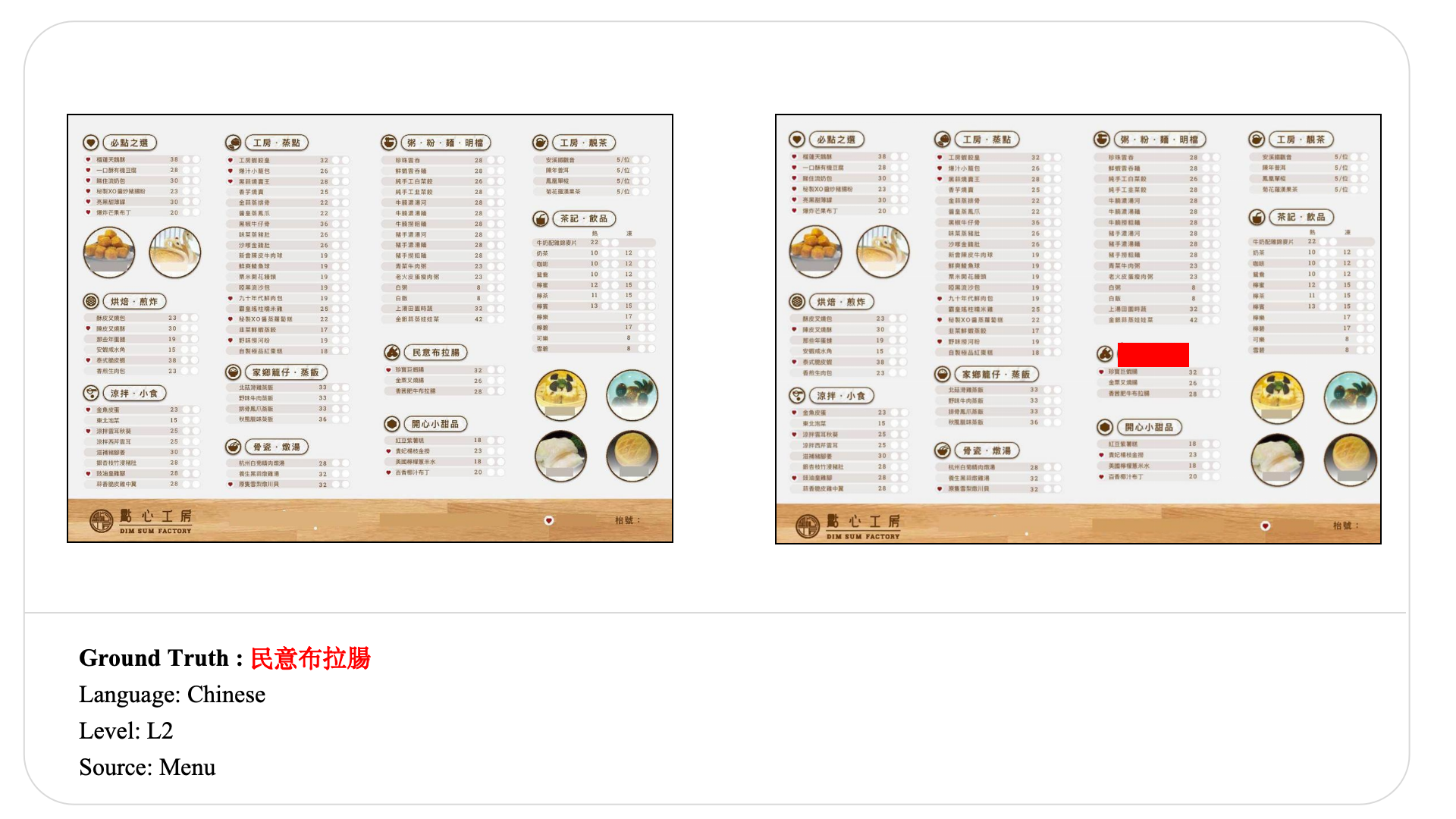}
    \caption{Case 7 from MMTR-Bench.}
    
    \vspace{1cm}
    
    \includegraphics[width=1\textwidth]{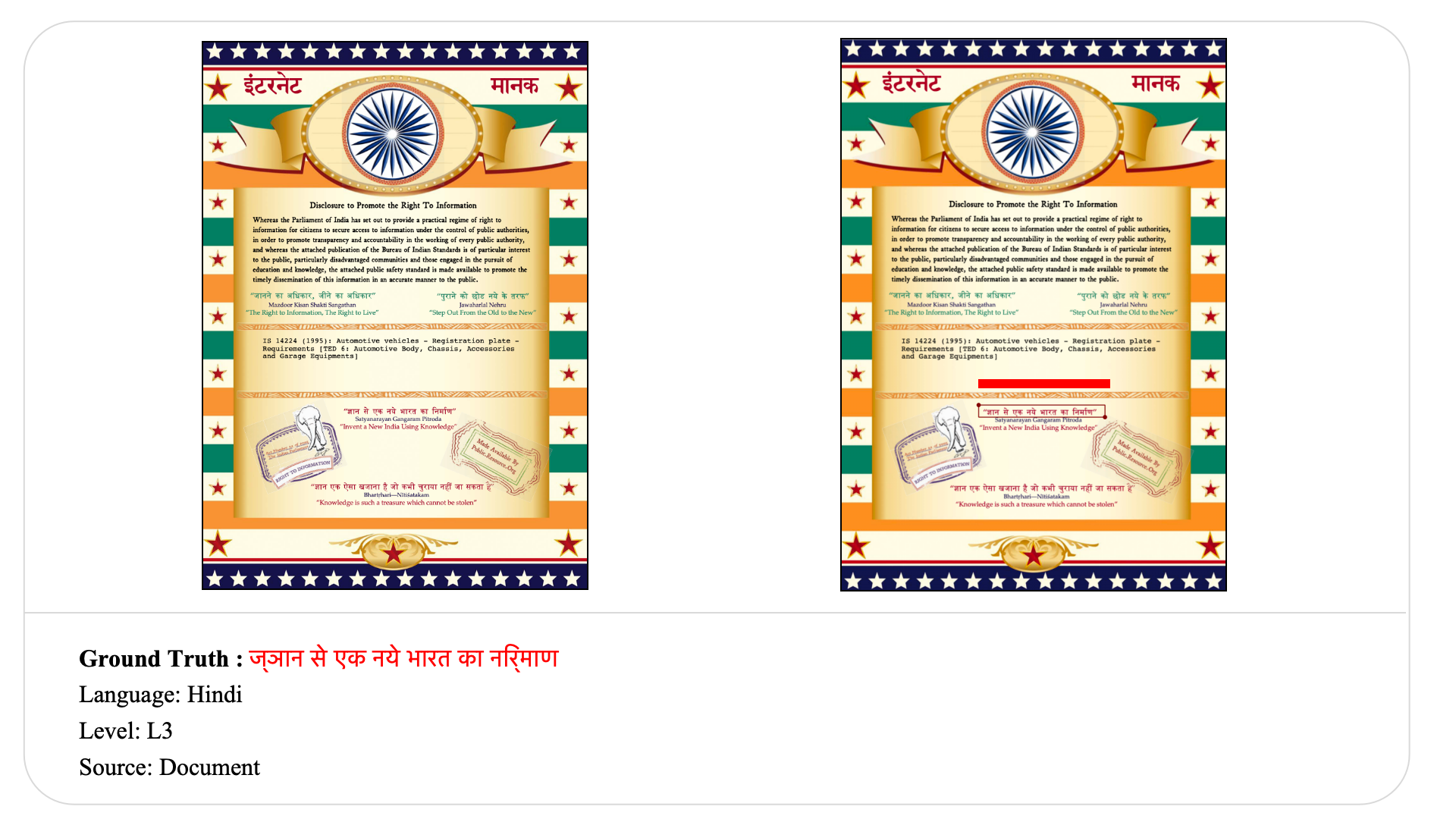}
    \caption{Case 8 from MMTR-Bench.}
\end{figure*}

\begin{figure*}[p]
    \centering
    \includegraphics[width=1\textwidth]{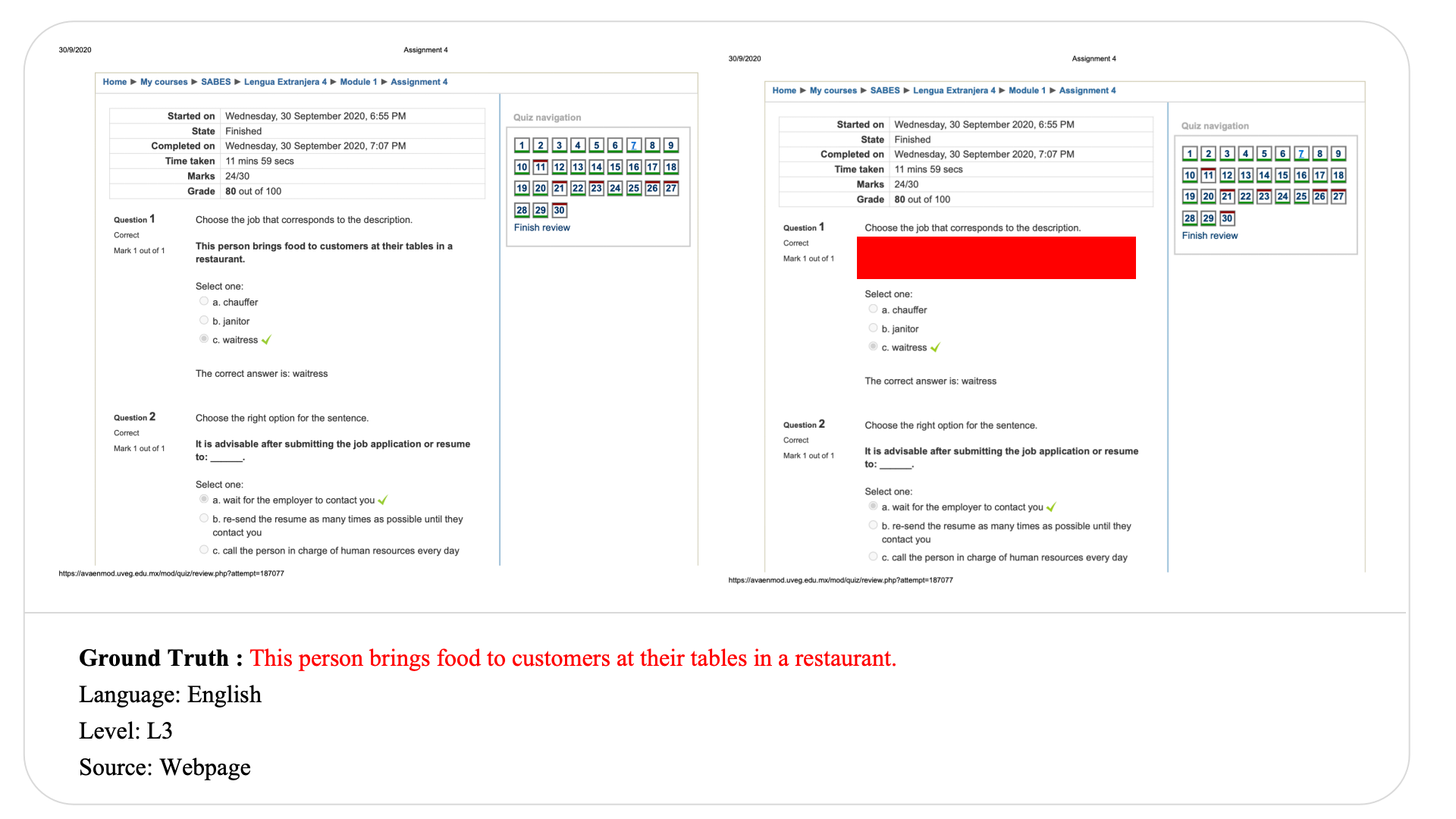}
    \caption{Case 9 from MMTR-Bench.}
    
    \vspace{1cm}
    
    \includegraphics[width=1\textwidth]{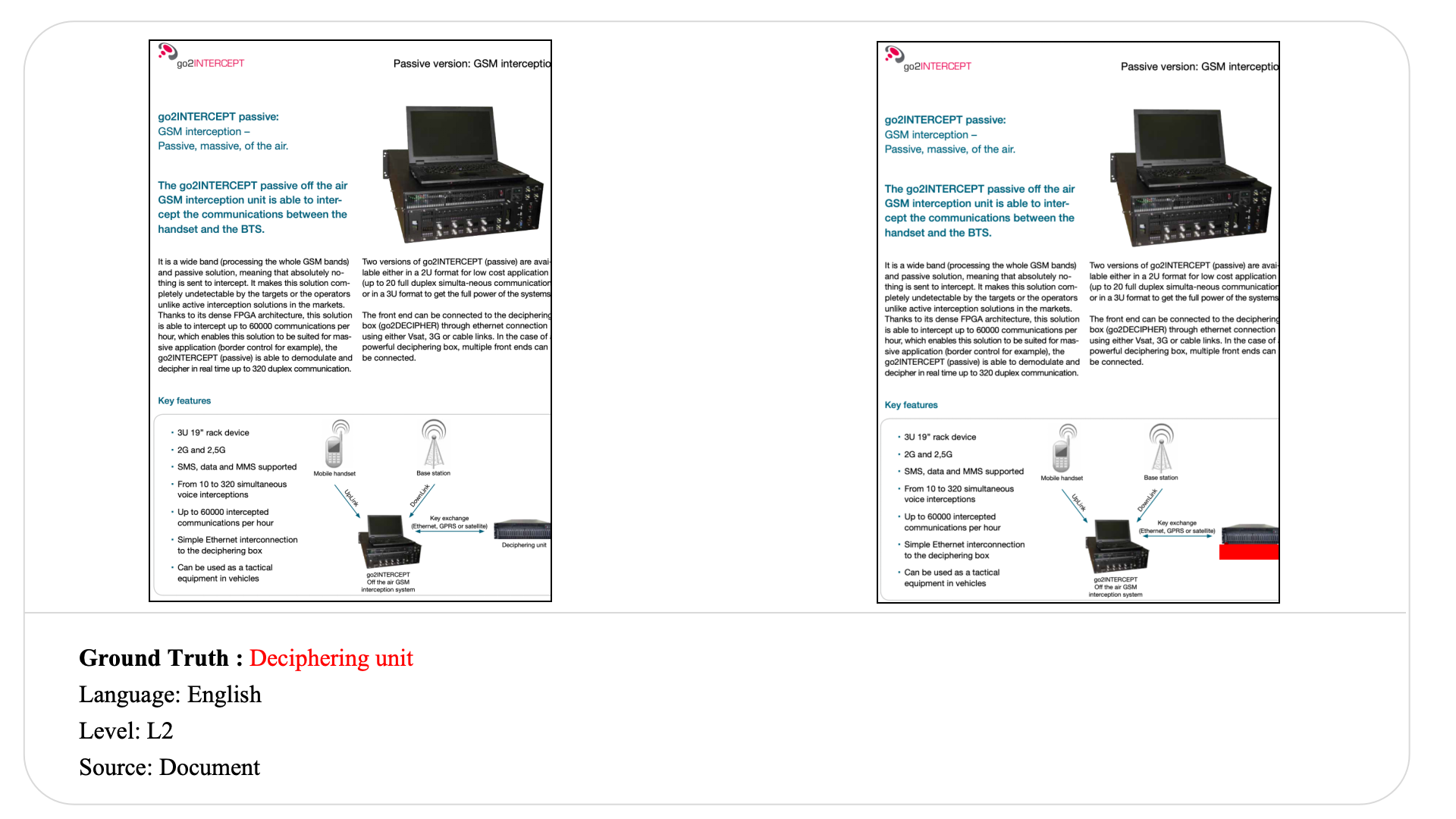}
    \caption{Case 10 from MMTR-Bench.}
\end{figure*}

\begin{figure*}[p]
    \centering
    \includegraphics[width=1\textwidth]{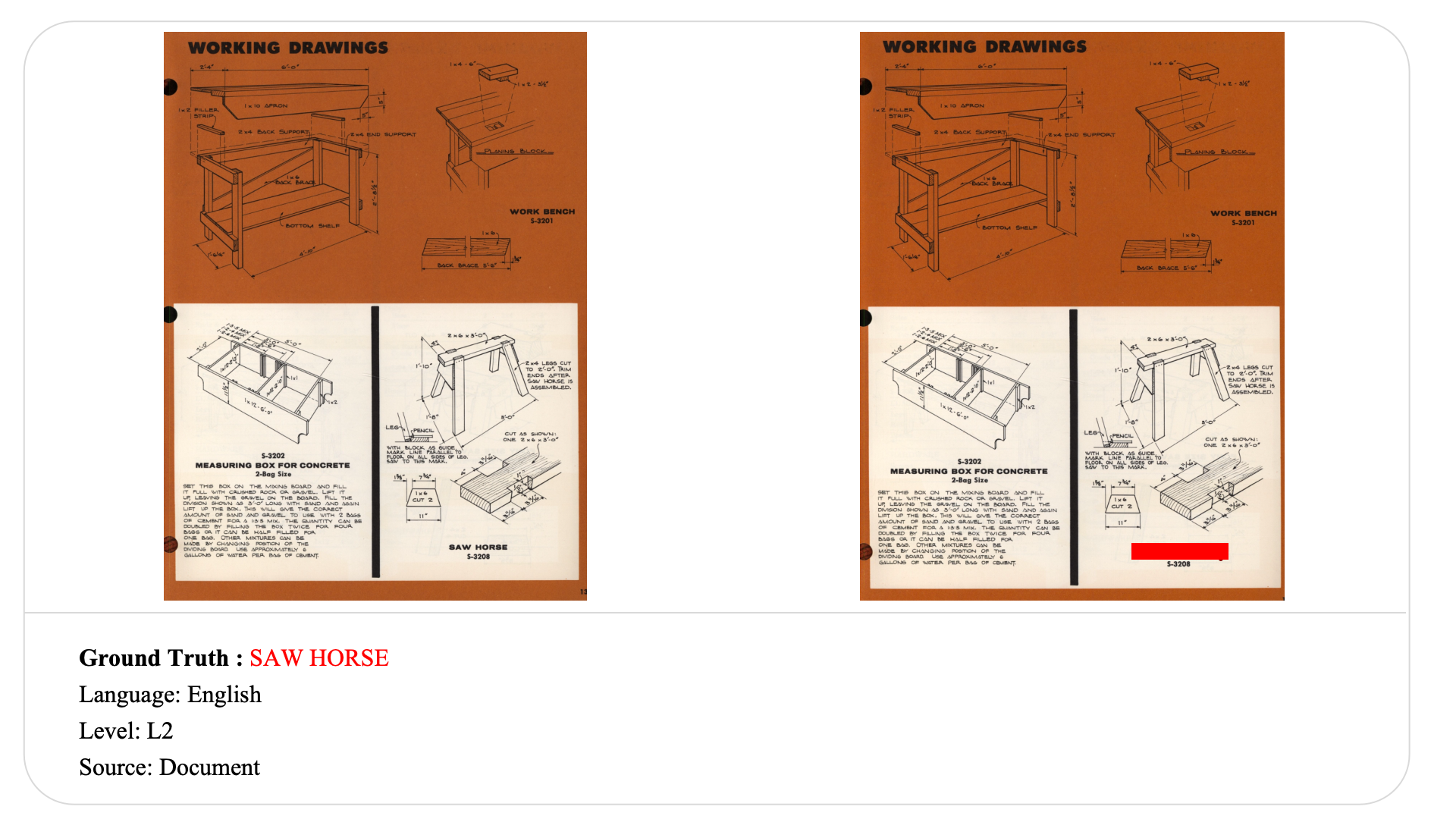}
    \caption{Case 11 from MMTR-Bench.}
    
    \vspace{1cm}
    
    \includegraphics[width=1\textwidth]{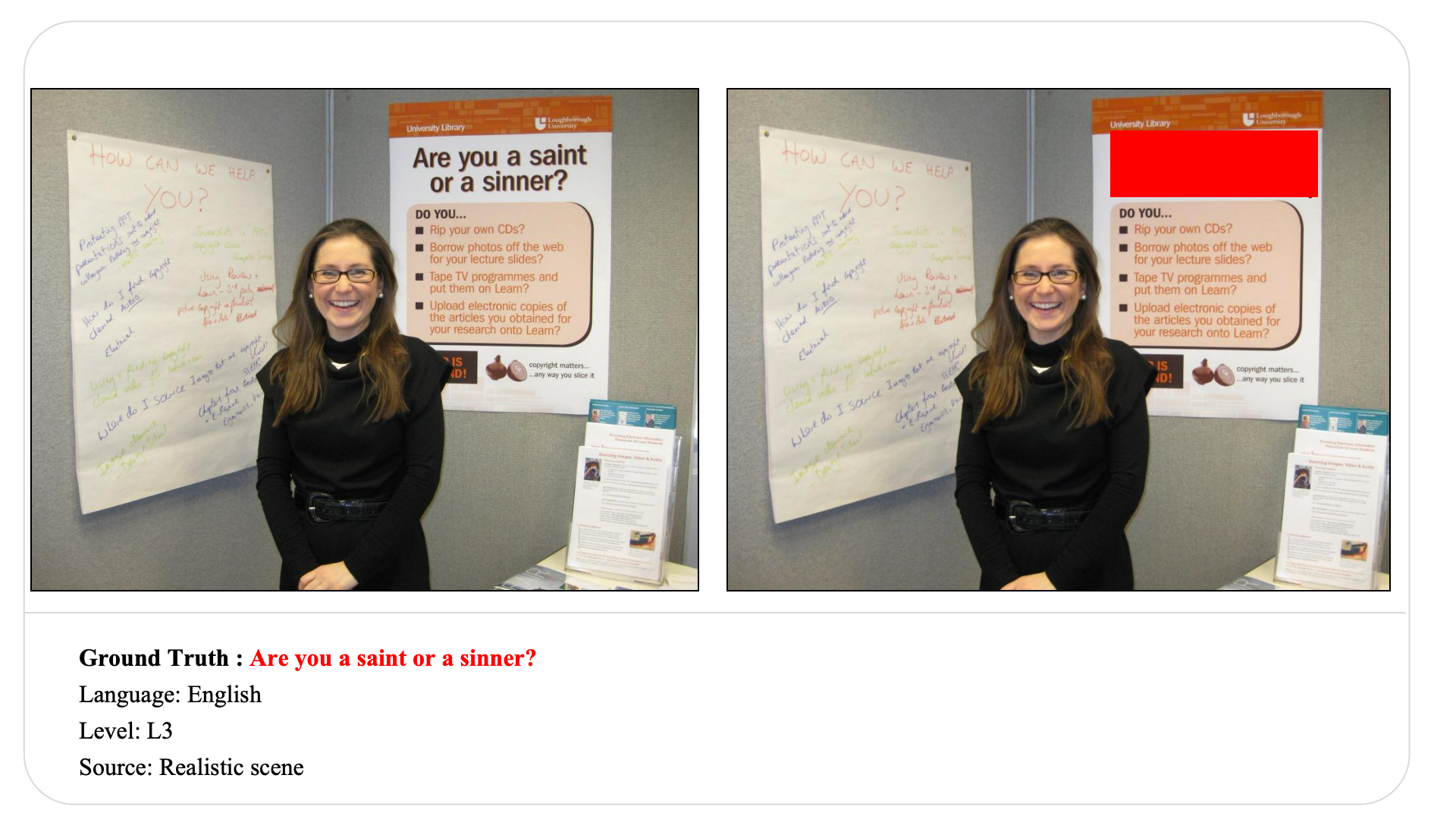}
    \caption{Case 12 from MMTR-Bench.}
\end{figure*}

\begin{figure*}[p]
    \centering
    \includegraphics[width=1\textwidth]{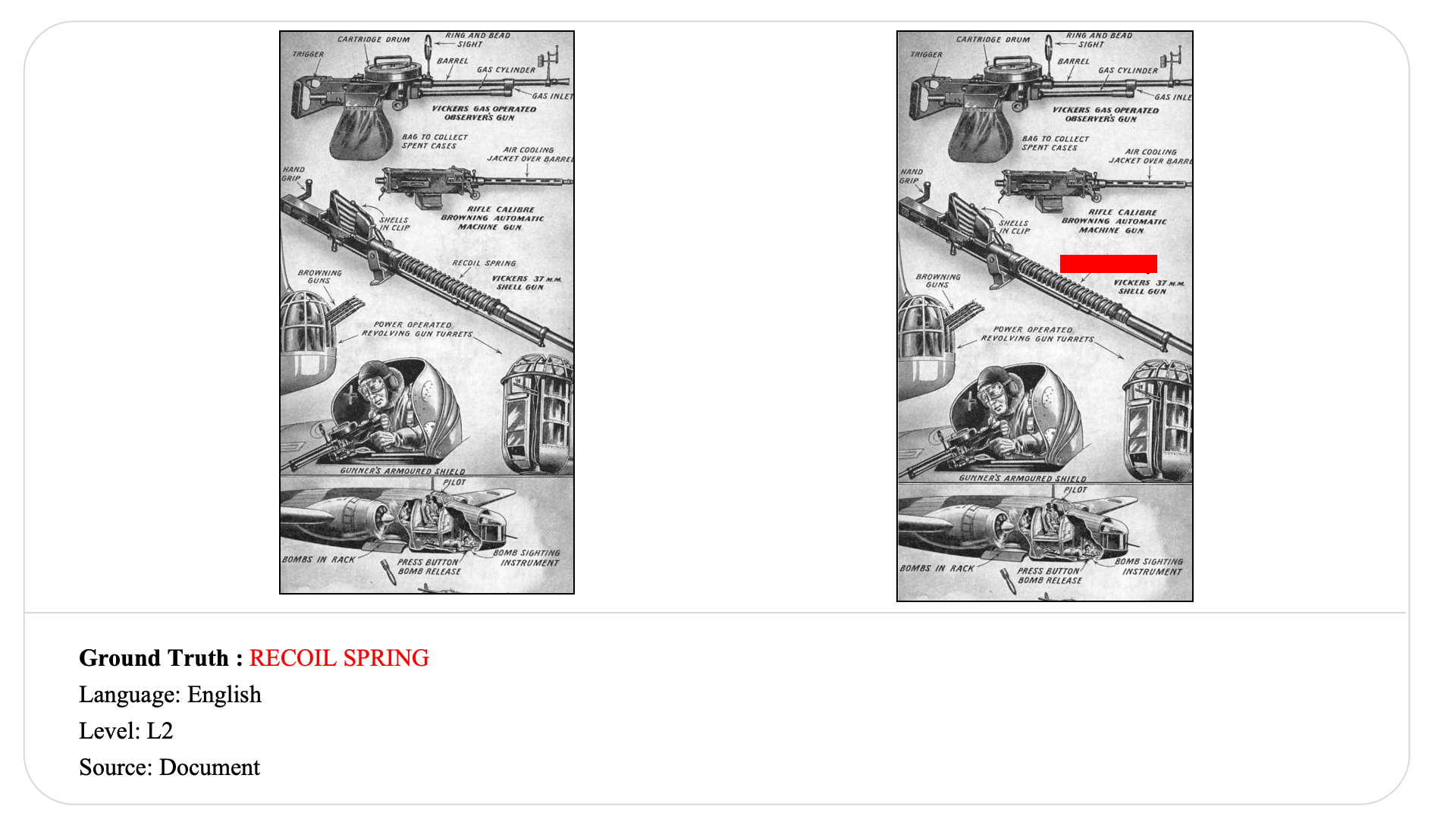}
    \caption{Case 13 from MMTR-Bench.}
    
    \vspace{1cm}
    
    \includegraphics[width=1\textwidth]{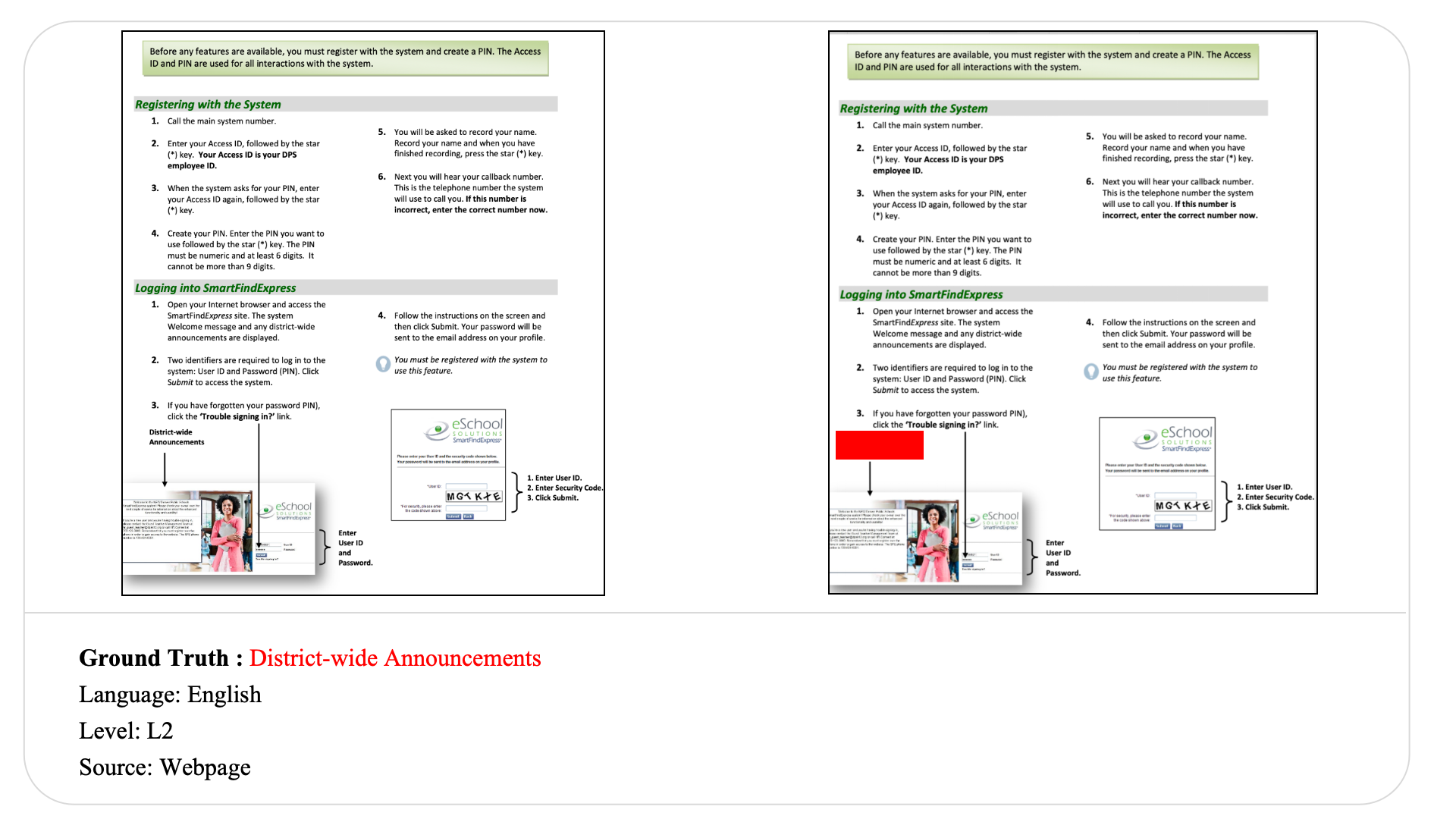}
    \caption{Case 14 from MMTR-Bench.}
\end{figure*}

\begin{figure*}[p]
    \centering
    \includegraphics[width=1\textwidth]{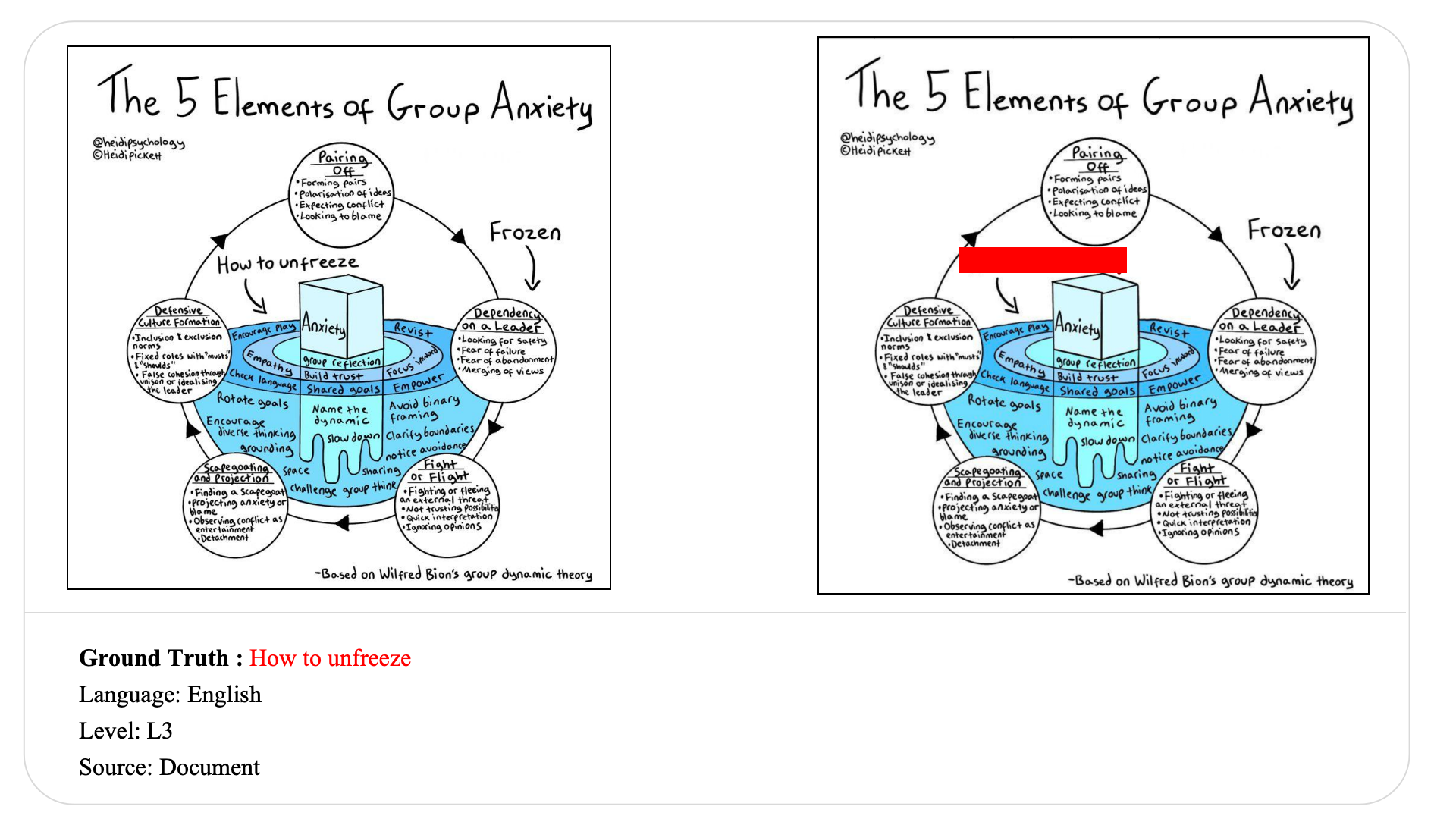}
    \caption{Case 15 from MMTR-Bench.}
    
    \vspace{1cm}
    
    \includegraphics[width=1\textwidth]{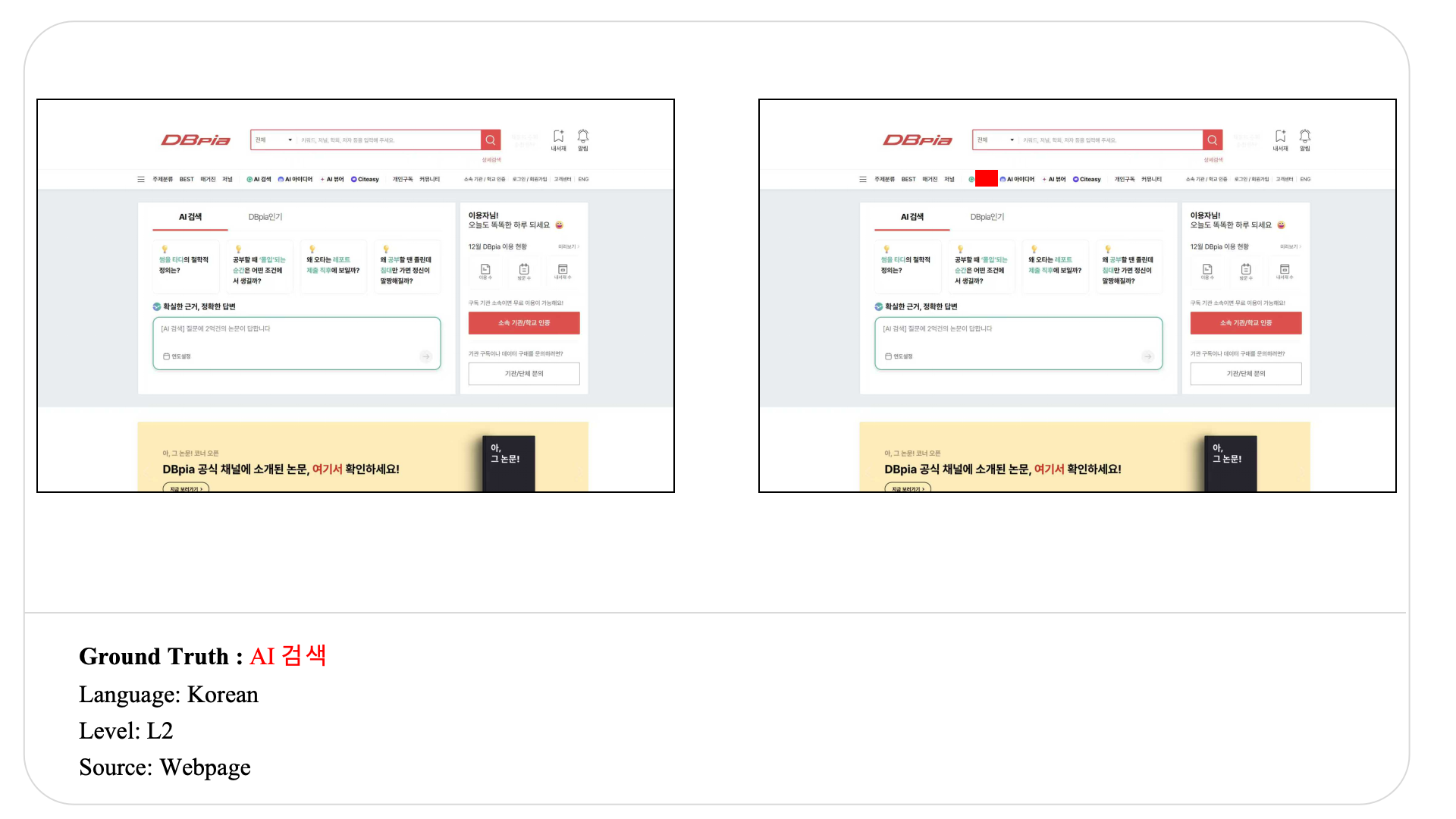}
    \caption{Case 16 from MMTR-Bench.}
\end{figure*}

\begin{figure*}[p]
    \centering
    \includegraphics[width=1\textwidth]{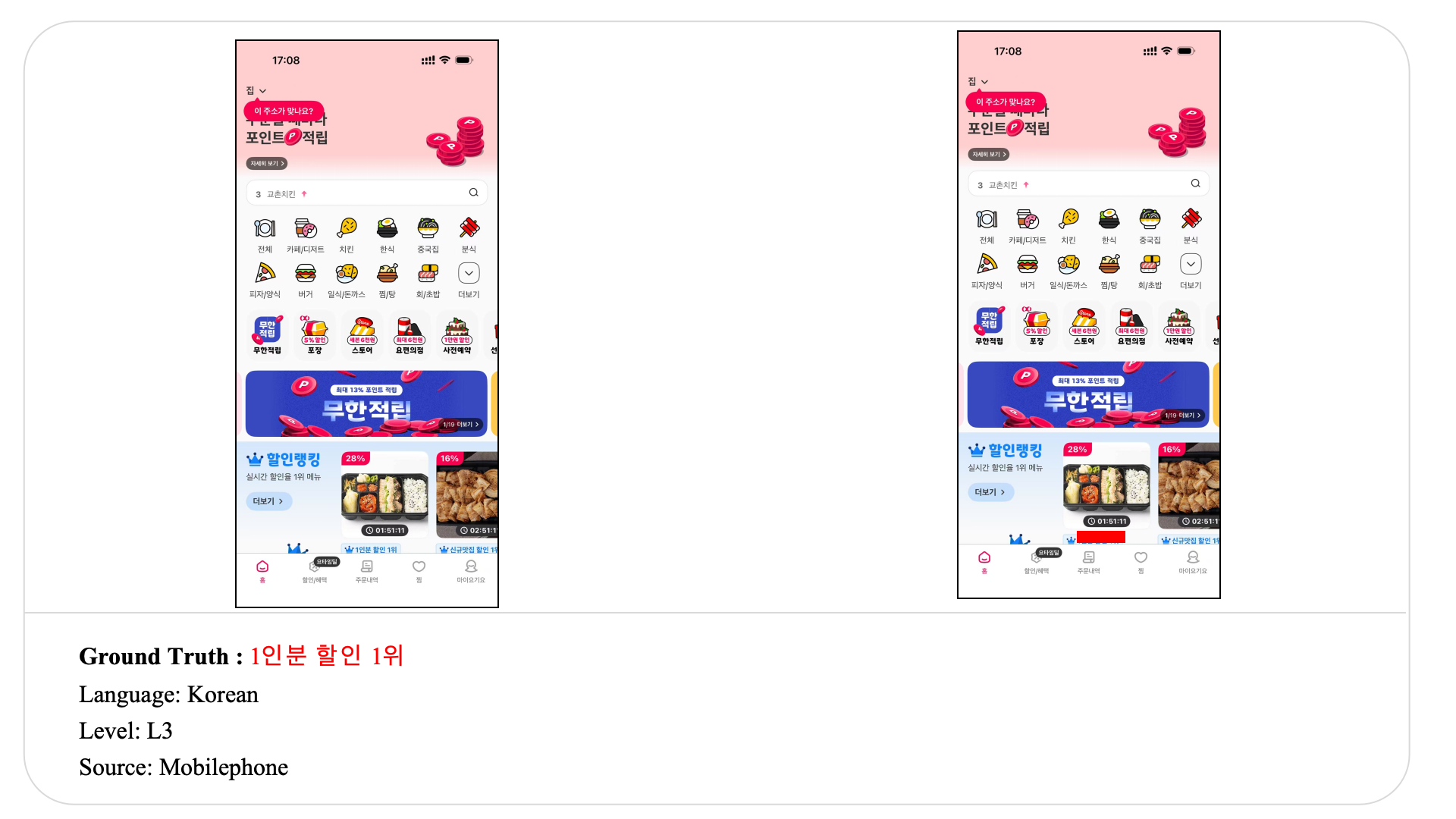}
    \caption{Case 17 from MMTR-Bench.}
    
    \vspace{1cm}
    
    \includegraphics[width=1\textwidth]{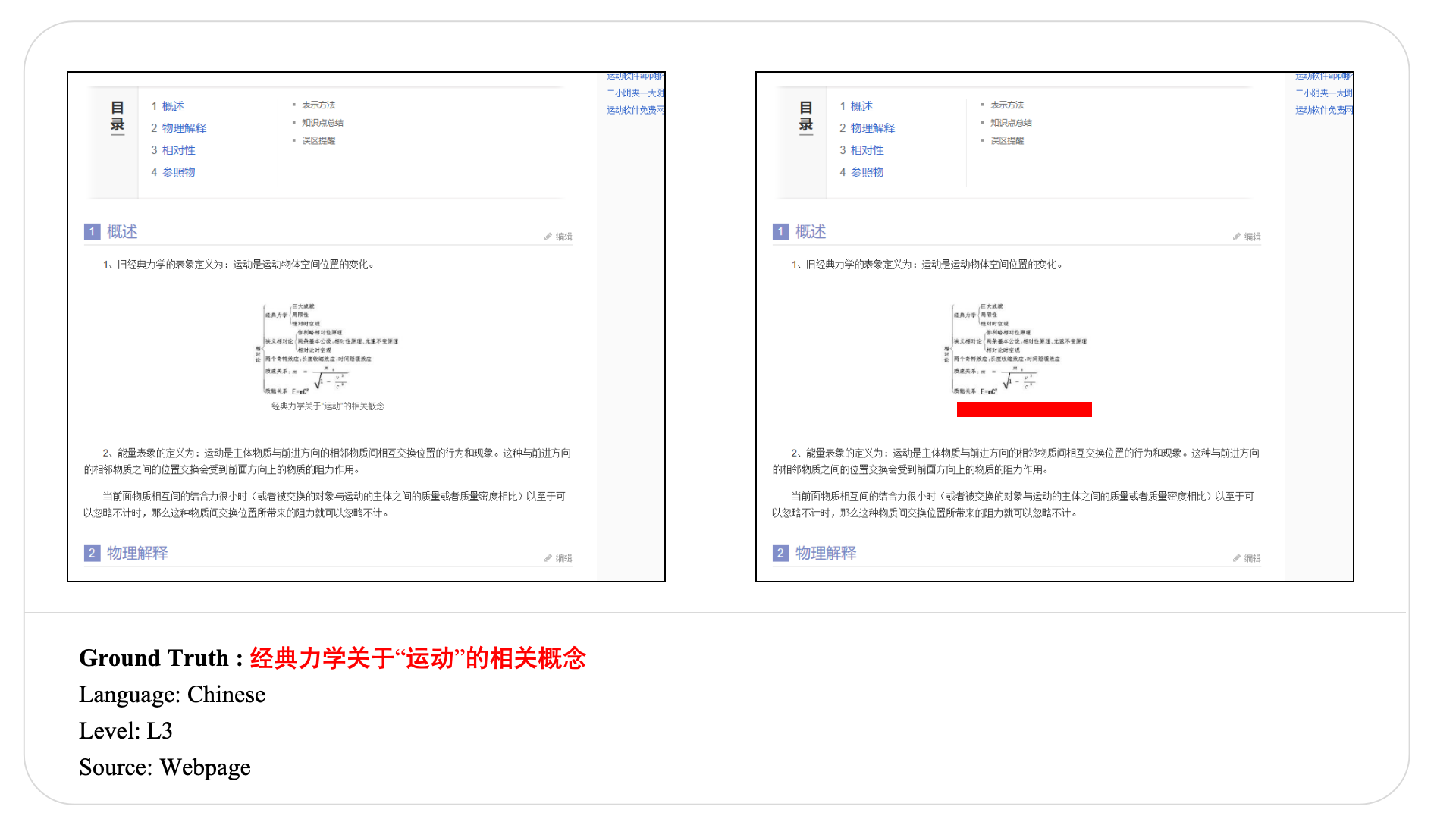}
    \caption{Case 18 from MMTR-Bench.}
\end{figure*}

\begin{figure*}[p]
    \centering
    \includegraphics[width=1\textwidth]{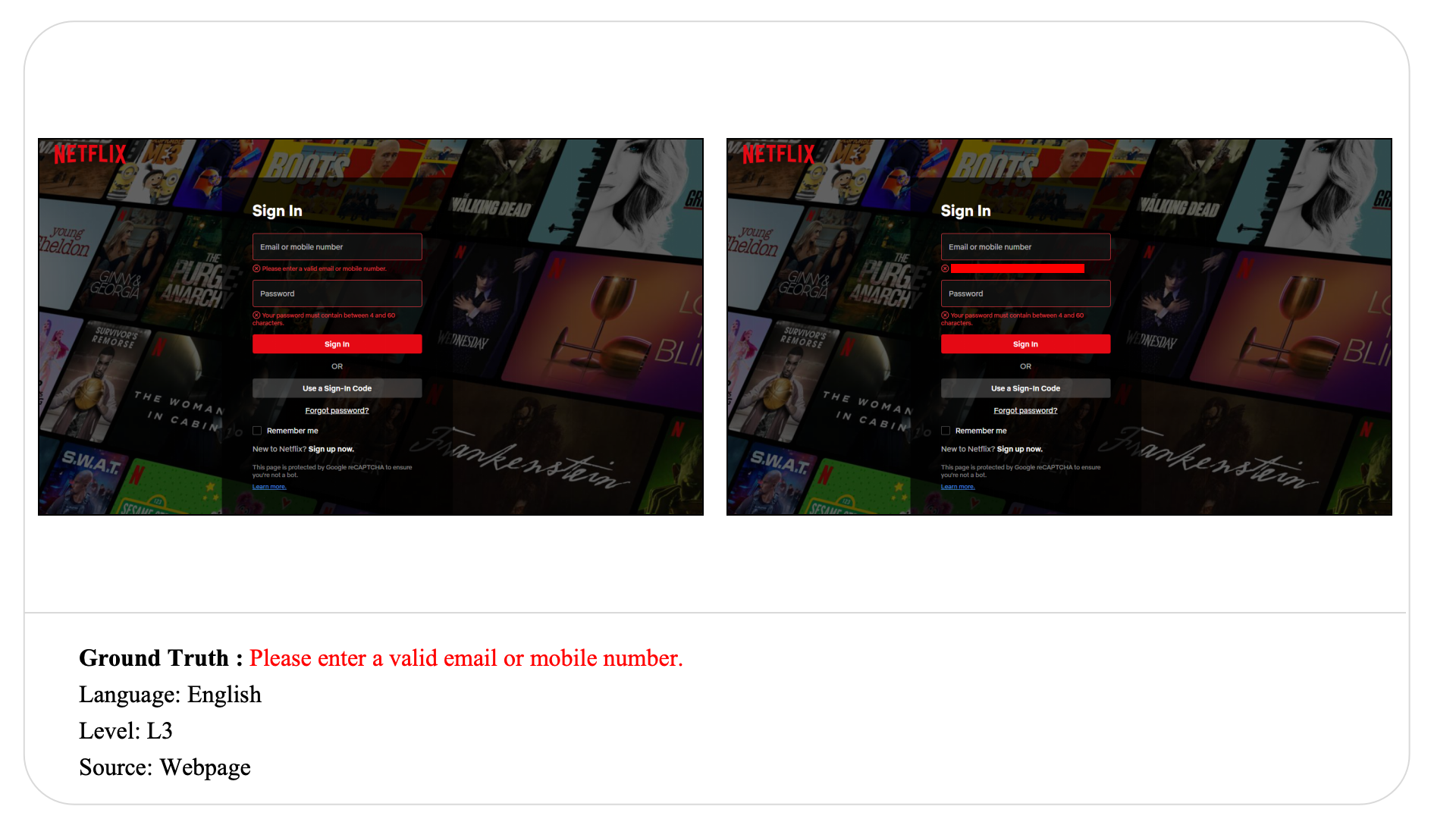}
    \caption{Case 19 from MMTR-Bench.}
    
    \vspace{1cm}
    
    \includegraphics[width=1\textwidth]{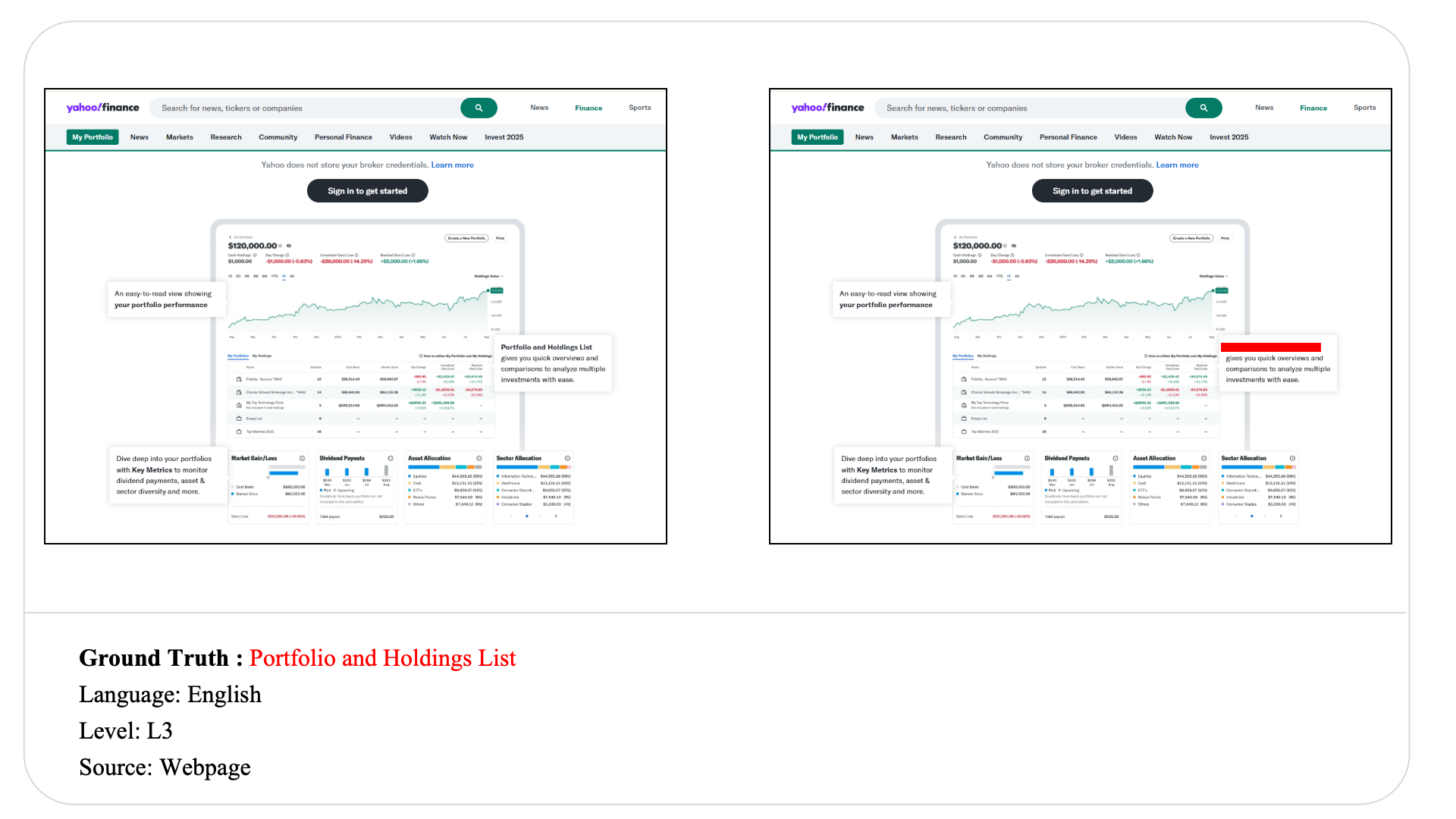}
    \caption{Case 20 from MMTR-Bench.}
\end{figure*}

\begin{figure*}[p]
    \centering
    \includegraphics[width=1\textwidth]{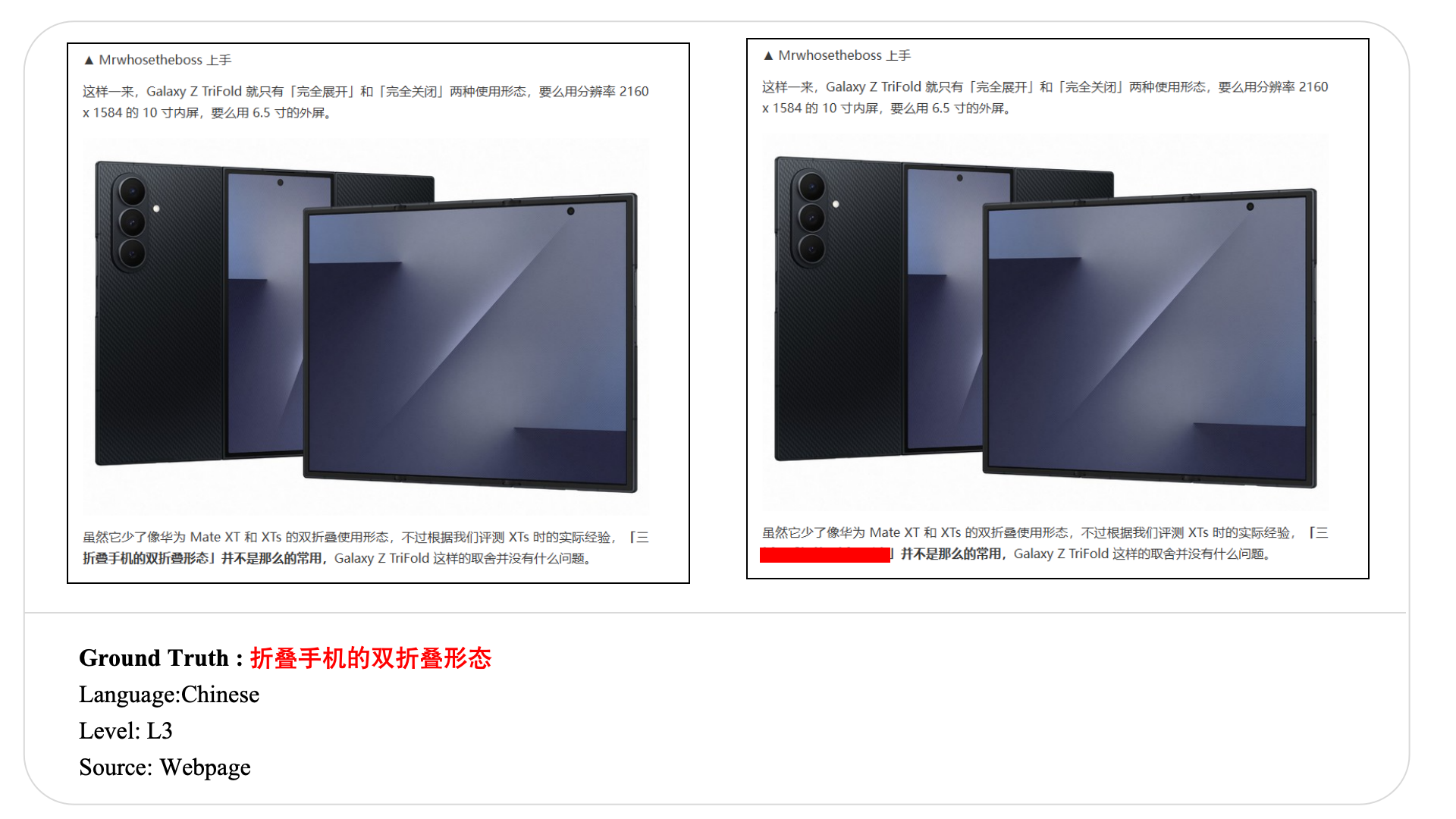}
    \caption{Case 21 from MMTR-Bench.}
    
    \vspace{1cm}
    
    \includegraphics[width=1\textwidth]{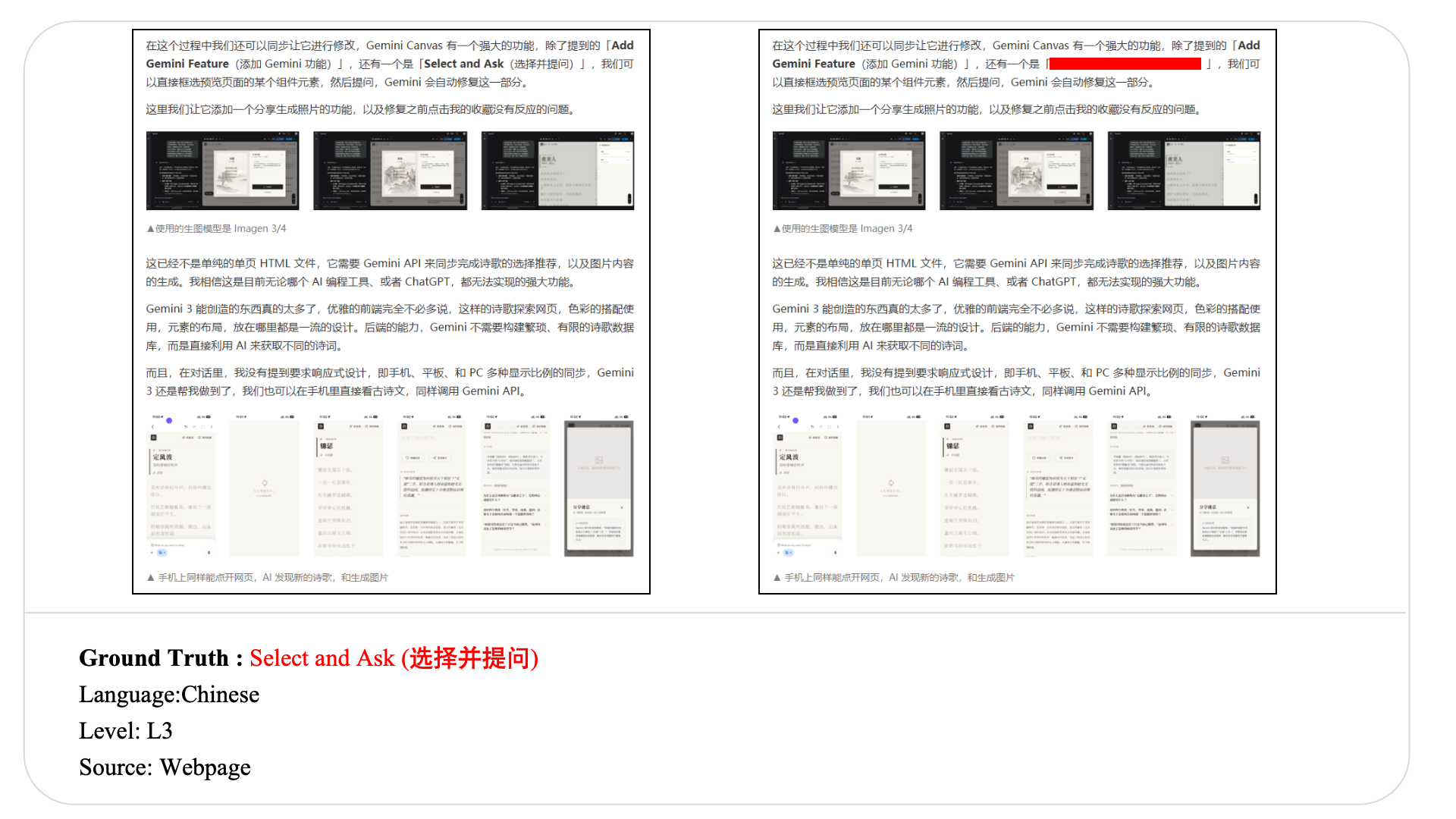}
    \caption{Case 22 from MMTR-Bench.}
\end{figure*}

\begin{figure*}[p]
    \centering
    \includegraphics[width=1\textwidth]{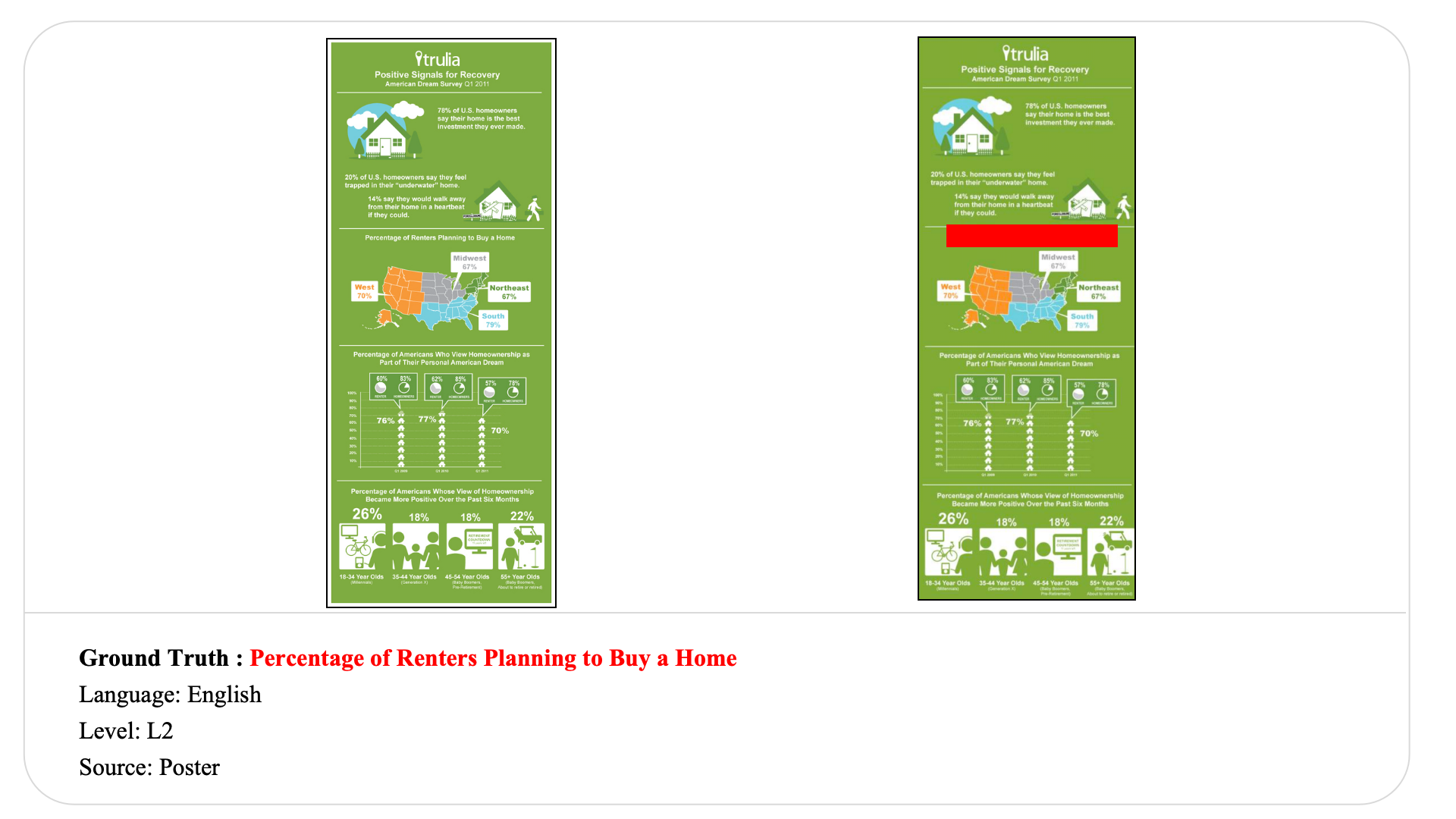}
    \caption{Case 23 from MMTR-Bench.}
    
    \vspace{1cm}
    
    \includegraphics[width=1\textwidth]{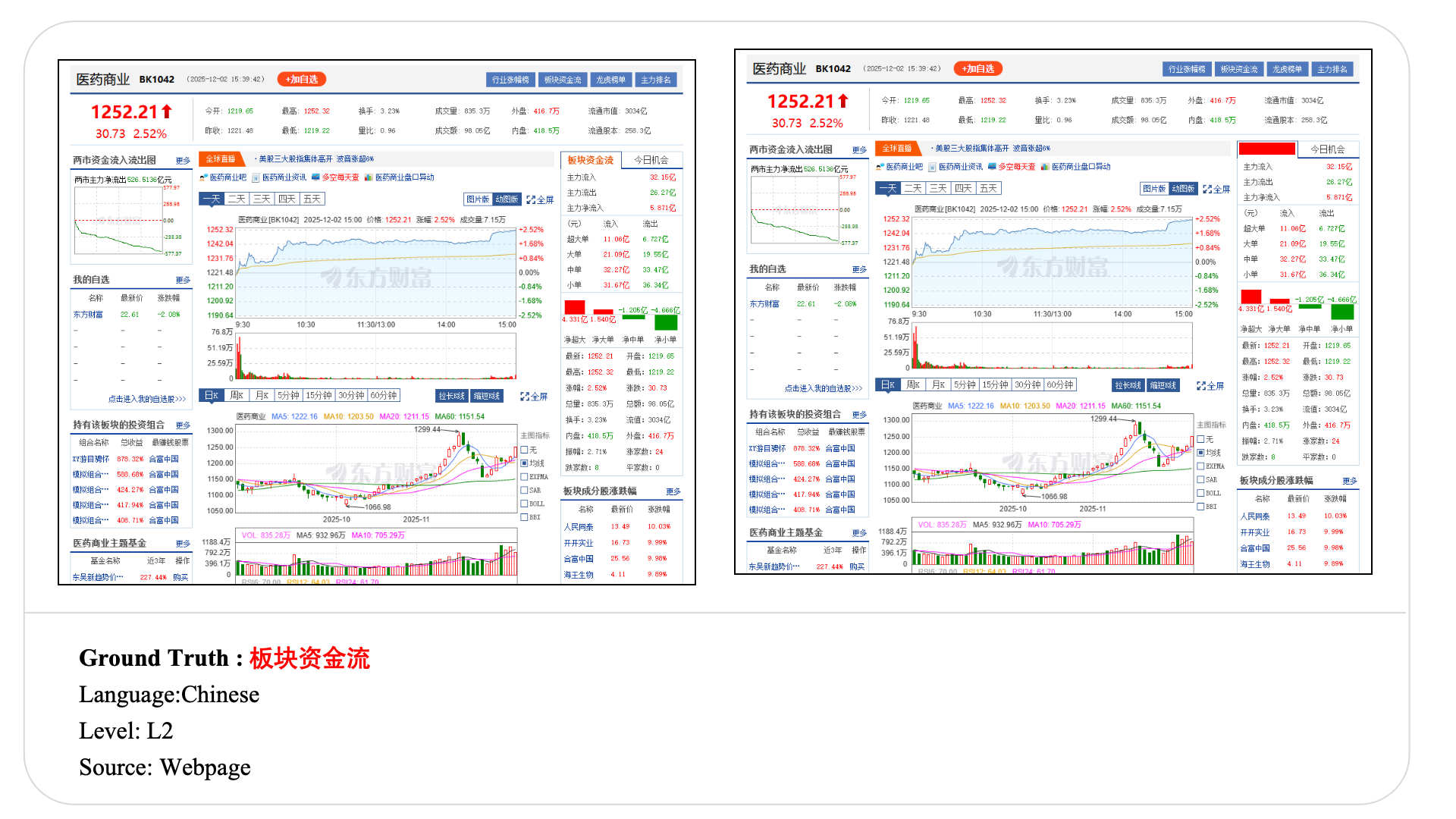}
    \caption{Case 24 from MMTR-Bench.}
\end{figure*}

\begin{figure*}[p]
    \centering
    \includegraphics[width=1\textwidth]{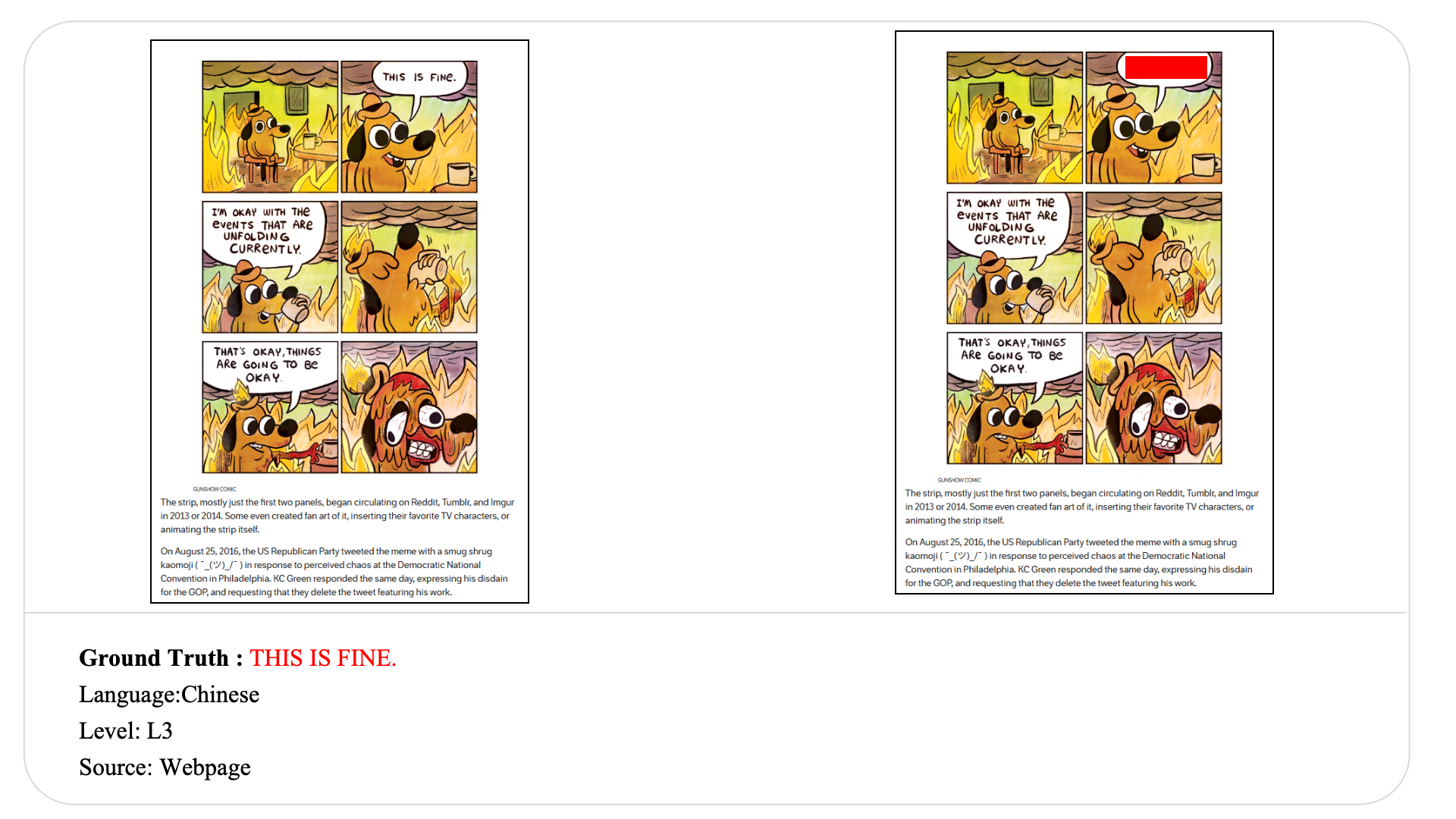}
    \caption{Case 25 from MMTR-Bench.}
    
    \vspace{1cm}
    
    \includegraphics[width=1\textwidth]{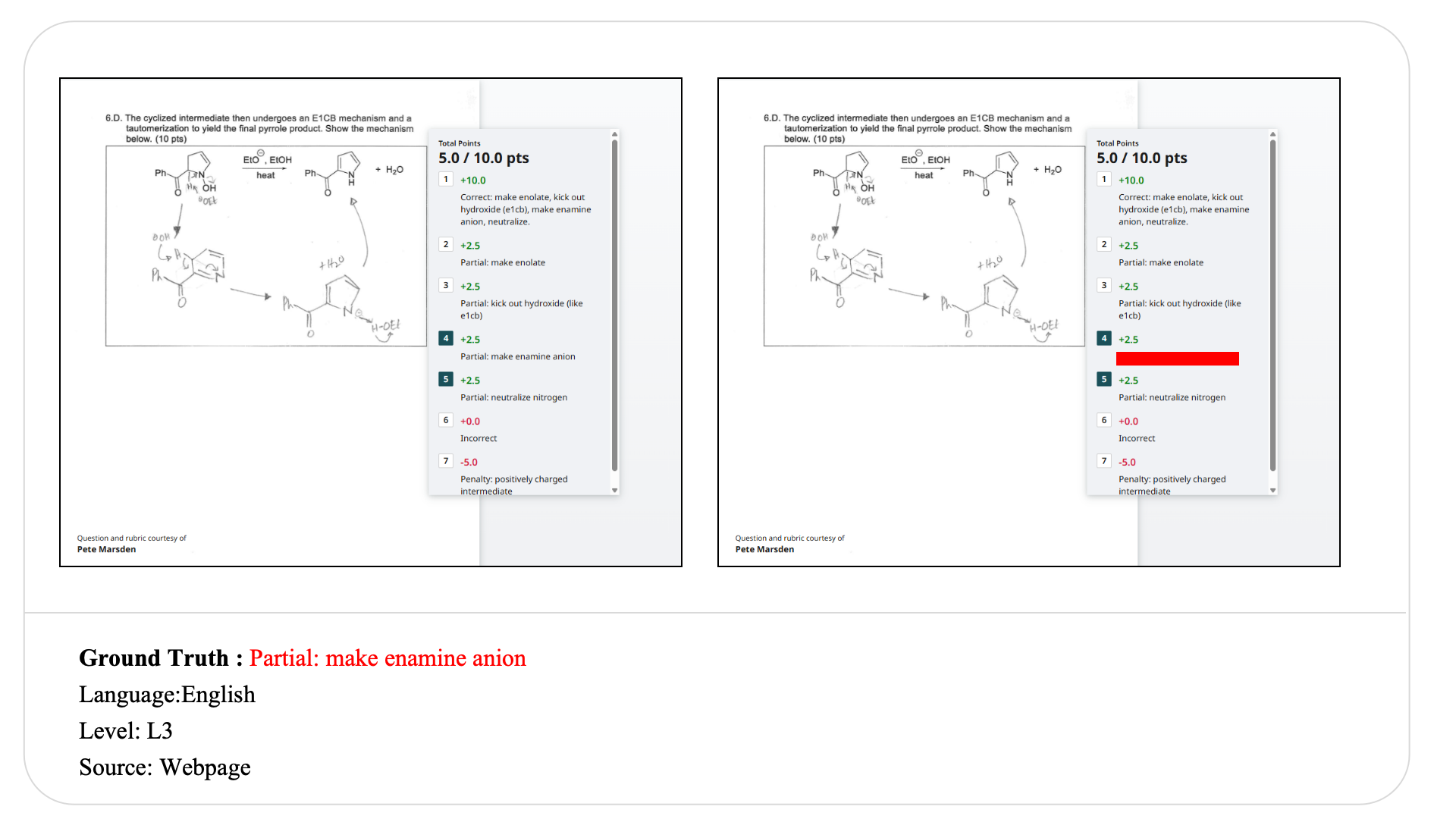}
    \caption{Case 26 from MMTR-Bench.}
\end{figure*}

\begin{figure*}[p]
    \centering
    \includegraphics[width=1\textwidth]{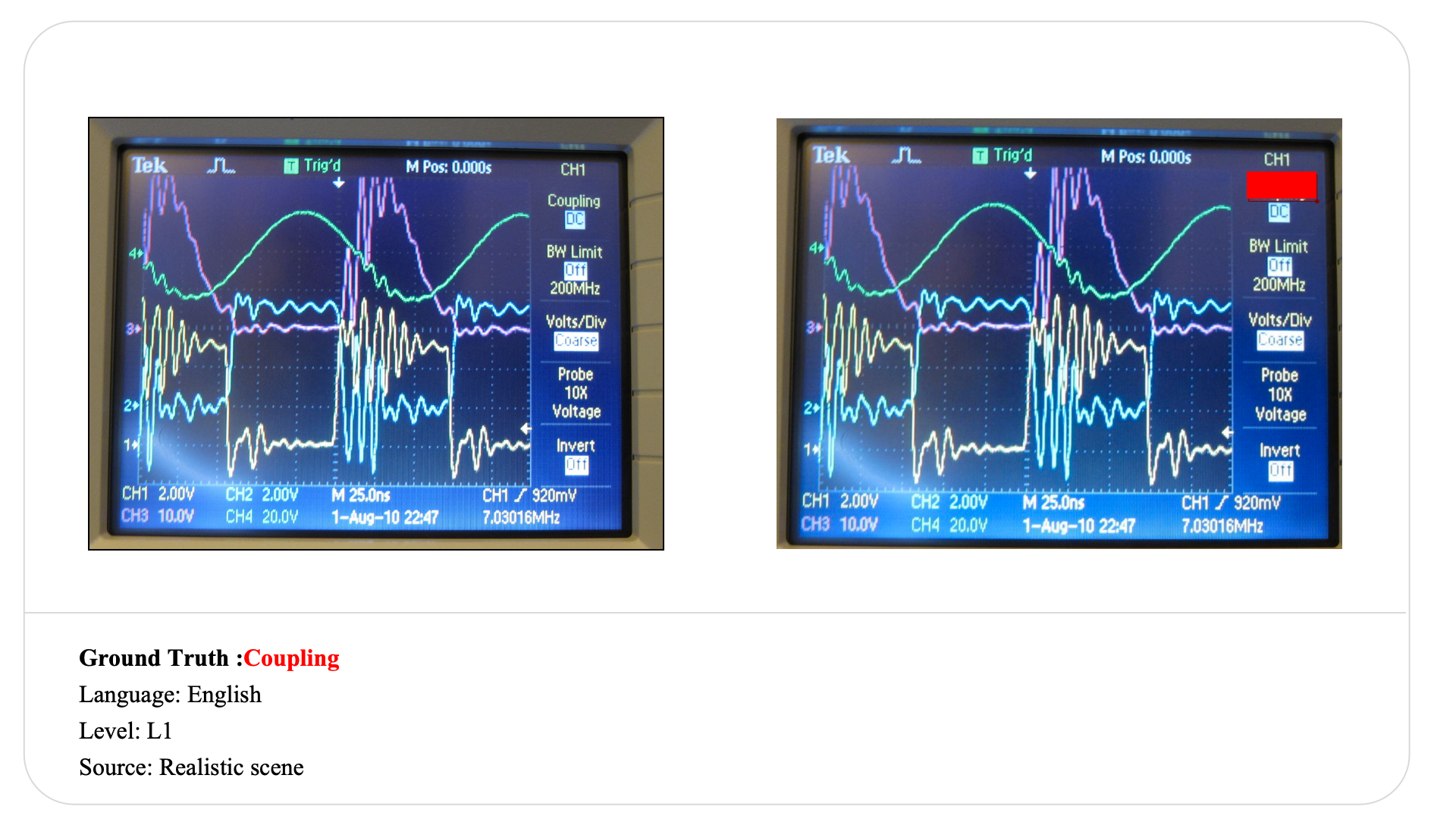}
    \caption{Case 27 from MMTR-Bench.}
    
    \vspace{1cm}
    
    \includegraphics[width=1\textwidth]{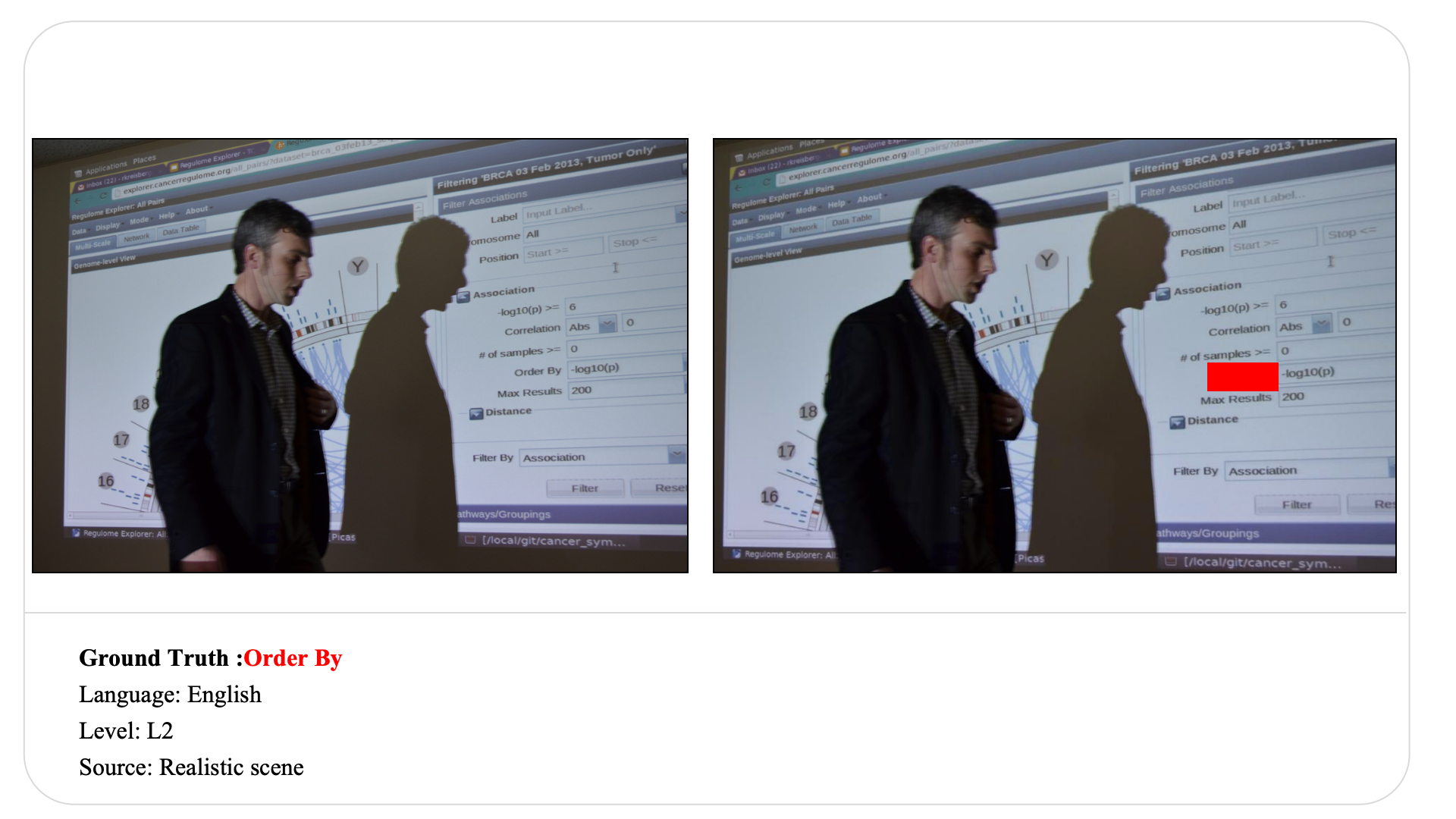}
    \caption{Case 28 from MMTR-Bench.}
\end{figure*}

\begin{figure*}[p]
    \centering
    \includegraphics[width=1\textwidth]{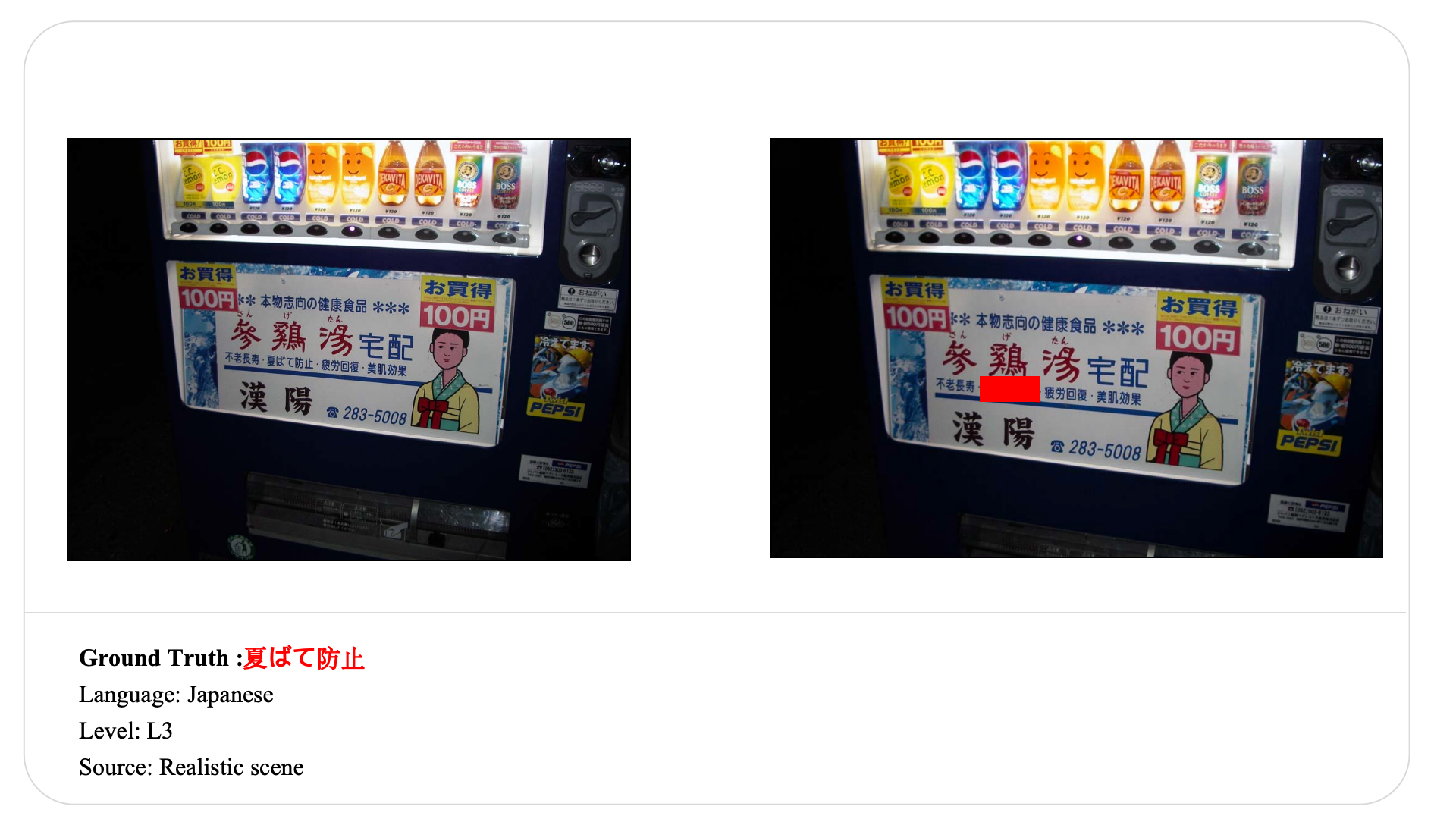}
    \caption{Case 29 from MMTR-Bench.}
    
    \vspace{1cm}
    
    \includegraphics[width=1\textwidth]{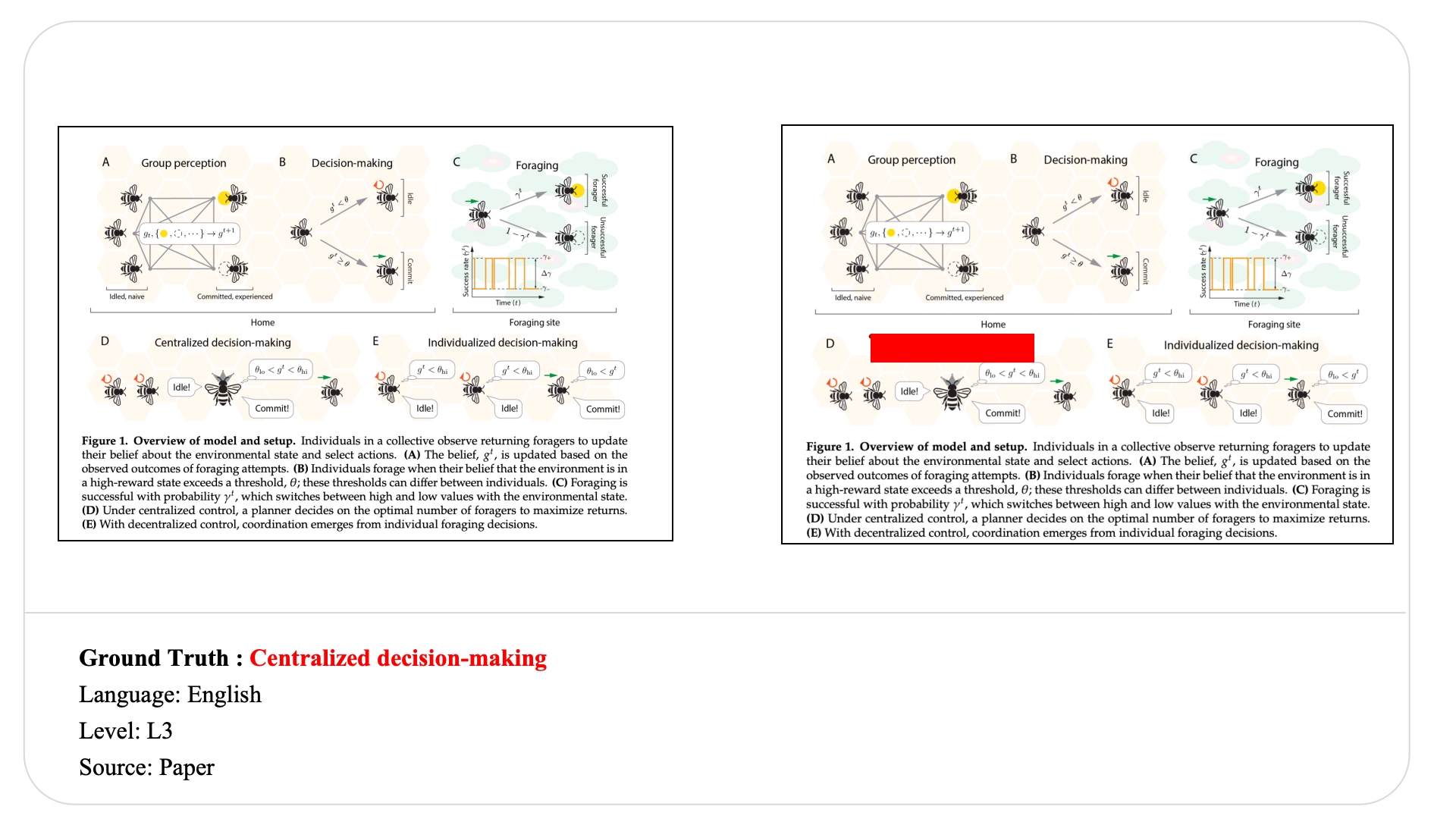}
    \caption{Case 30 from MMTR-Bench.}
\end{figure*}

\begin{figure*}[p]
    \centering
    \includegraphics[width=1\textwidth]{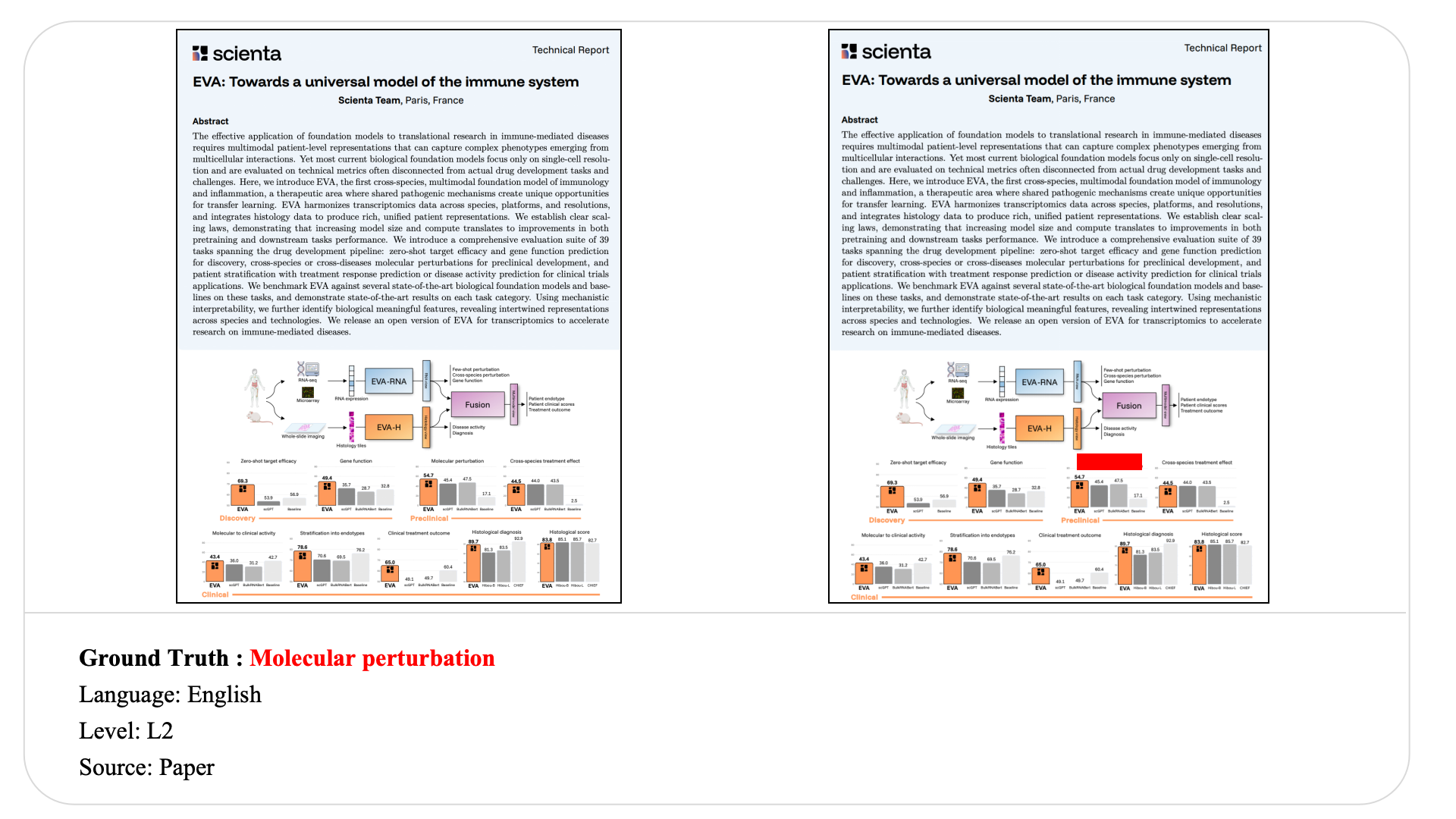}
    \caption{Case 31 from MMTR-Bench.}
    
    \vspace{1cm}
    
    \includegraphics[width=1\textwidth]{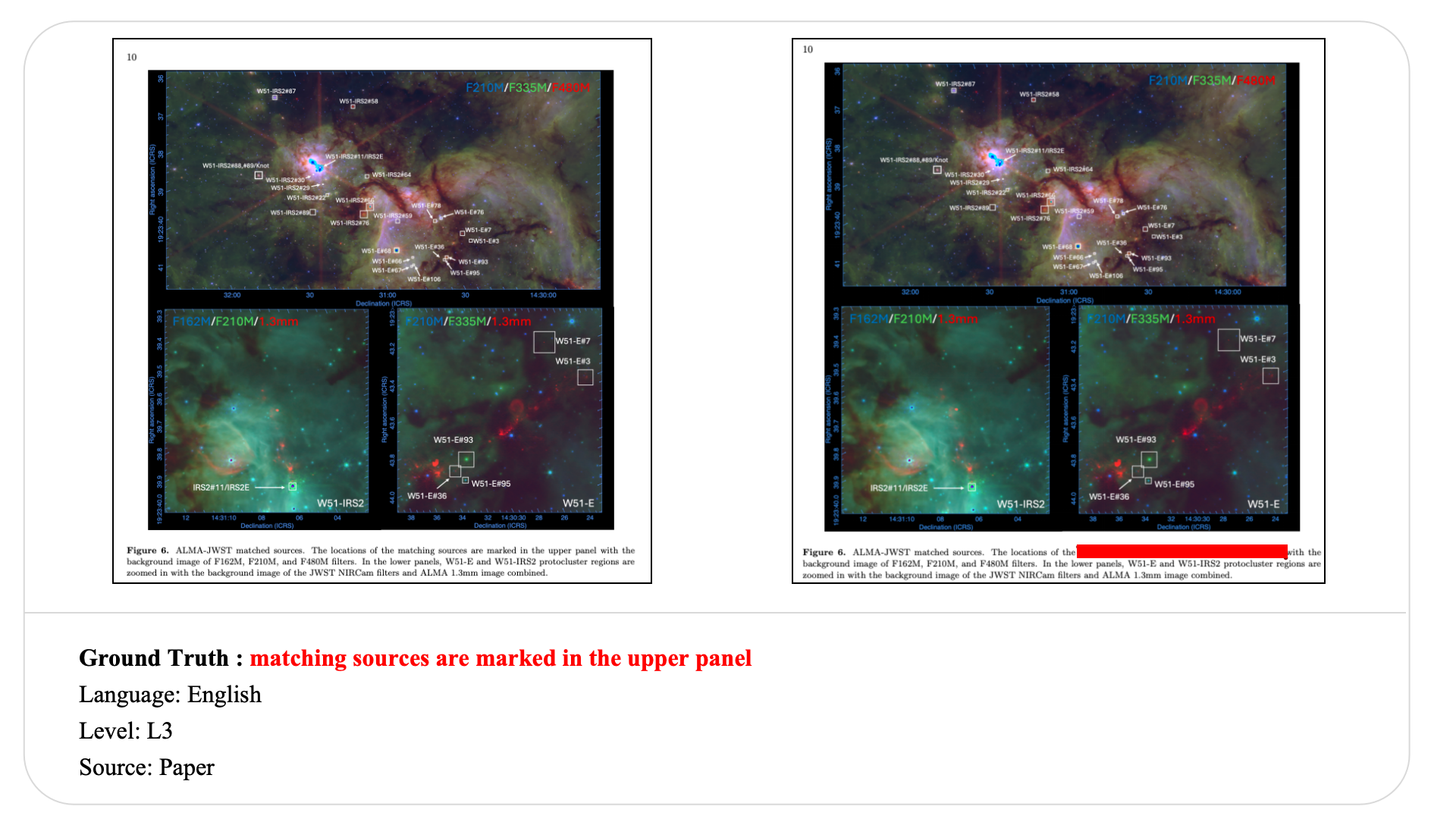}
    \caption{Case 32 from MMTR-Bench.}
\end{figure*}
\clearpage

\section{Additional Benchmark Statistics}
\label{app:benchmark_stats}

This appendix provides additional statistics of MMTR-Bench that are not shown in the main paper. We include these figures to give a more complete view of the dataset composition, target properties, and cross-factor distributions.
\newpage

\subsection{Basic dataset distributions}

\begin{figure}[h!]
    \centering
    \includegraphics[width=0.9\linewidth]{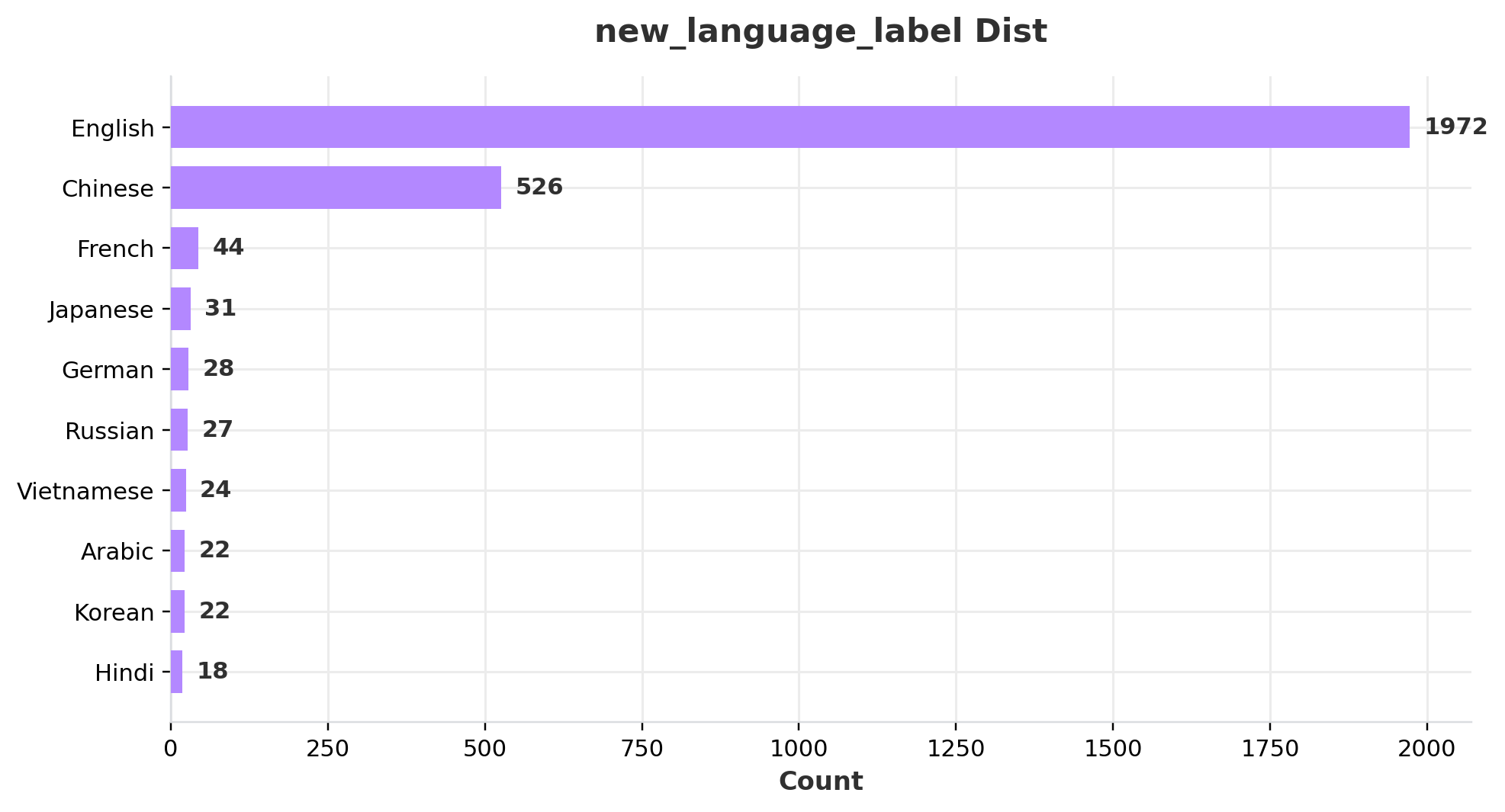}
    \caption{Language distribution of MMTR-Bench.}
    \label{fig:app_language_distribution}
\end{figure}

\begin{figure}[h!]
    \centering
    \includegraphics[width=0.9\linewidth]{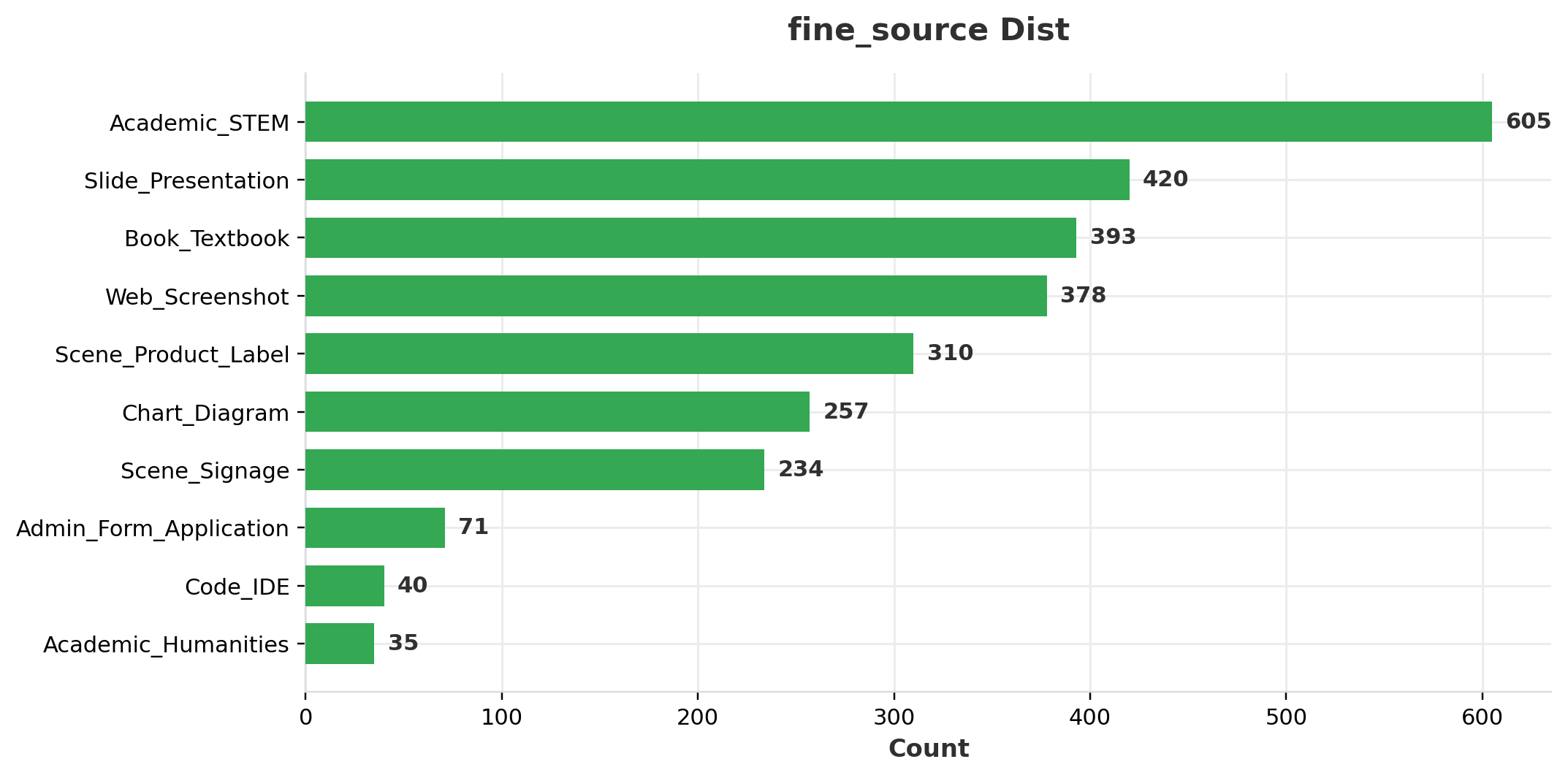}
    \caption{Fine-grained source distribution of MMTR-Bench.}
    \label{fig:app_fine_source_distribution}
\end{figure}

\newpage

\subsection{Target-length and masking statistics}

\begin{figure}[h!]
    \centering
    \includegraphics[width=0.9\linewidth]{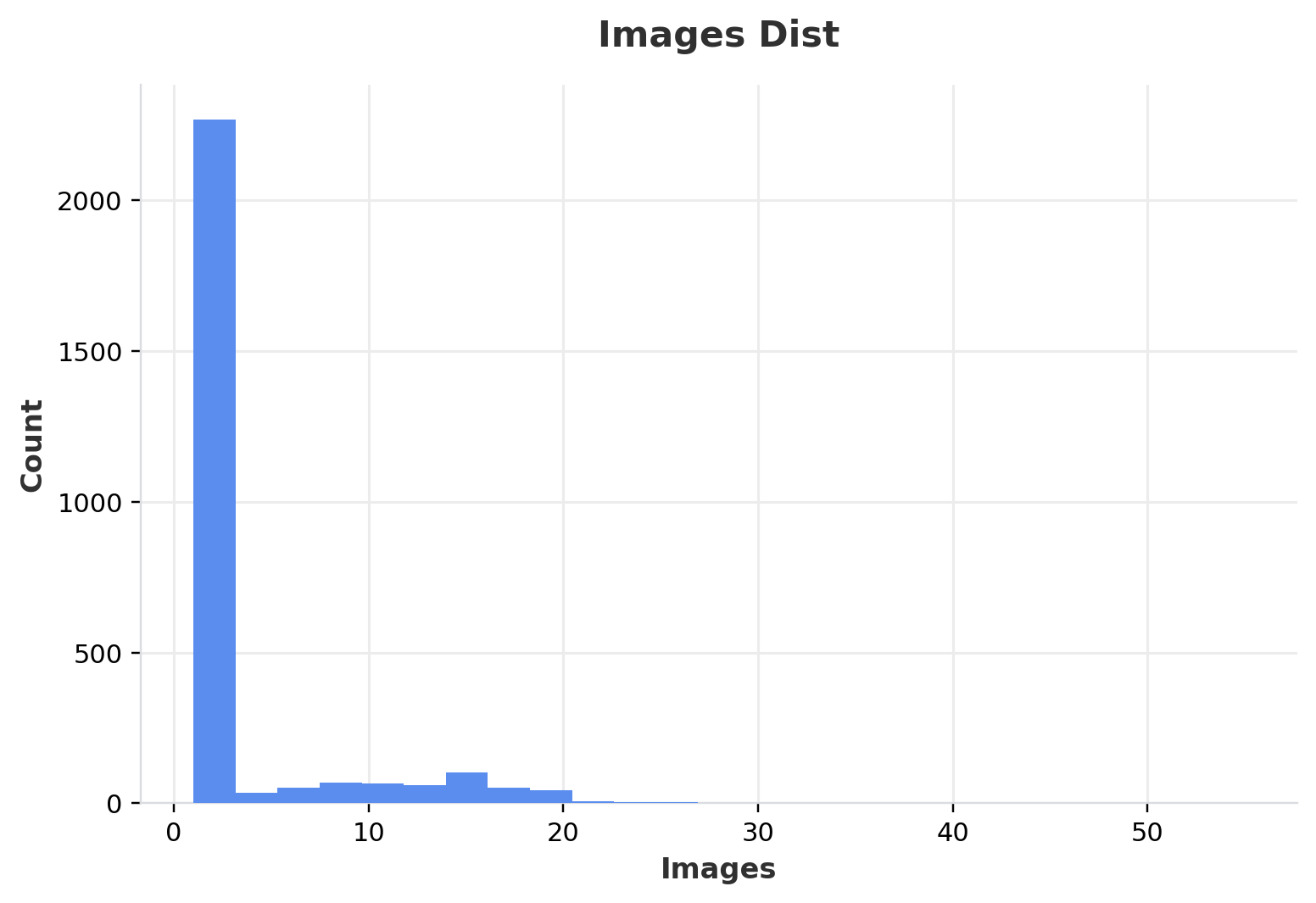}
    \caption{Histogram of the number of context images per sample.}
    \label{fig:app_num_context_images_hist}
\end{figure}

\begin{figure}[h!]
    \centering
    \includegraphics[width=0.9\linewidth]{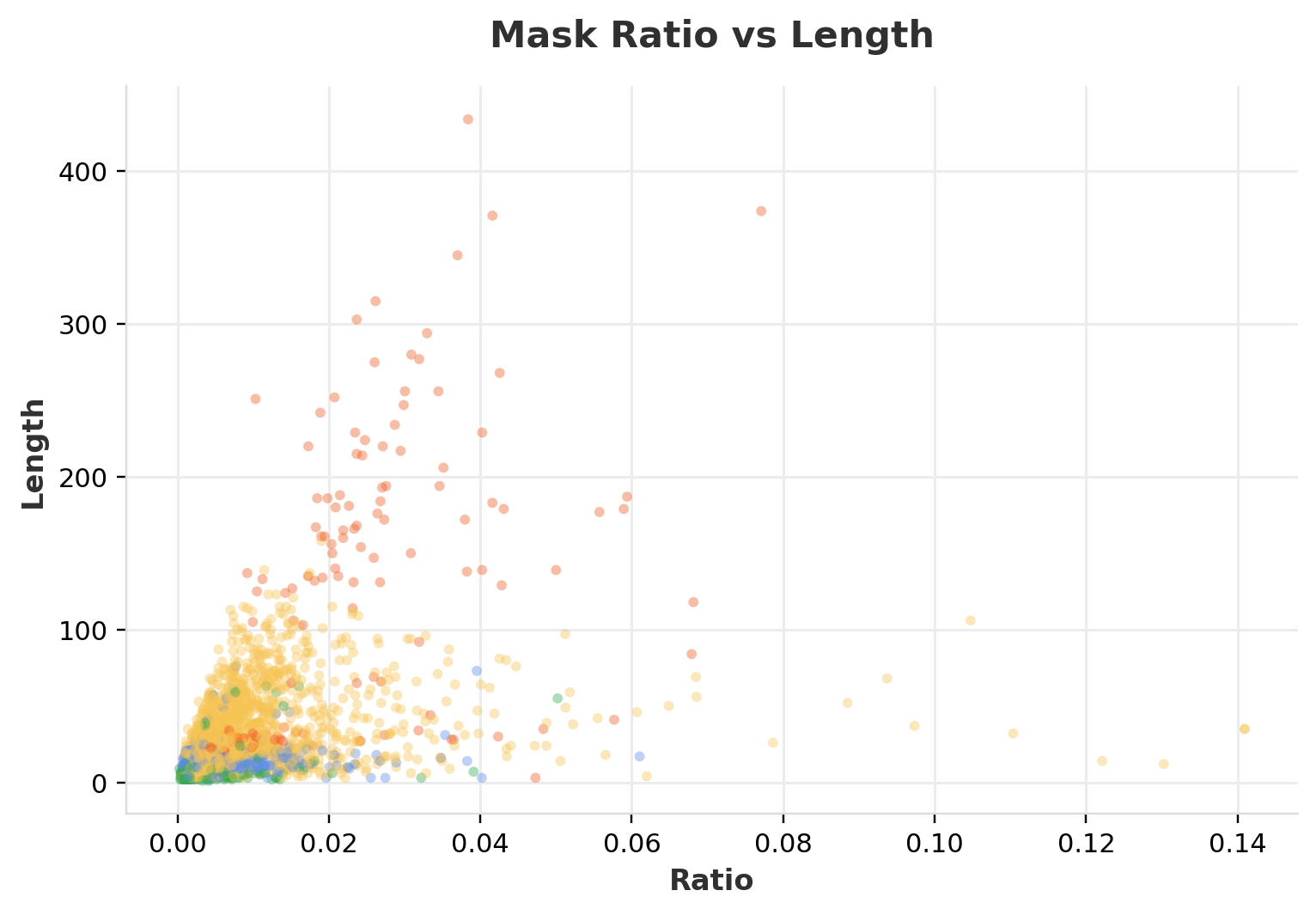}
    \caption{Relationship between mask ratio and target character length.}
    \label{fig:app_mask_ratio_vs_char_length}
\end{figure}

\newpage

\newpage

\subsection{Additional benchmark cross-slice views}

\begin{figure}[h!]
    \centering
    \includegraphics[width=0.9\linewidth]{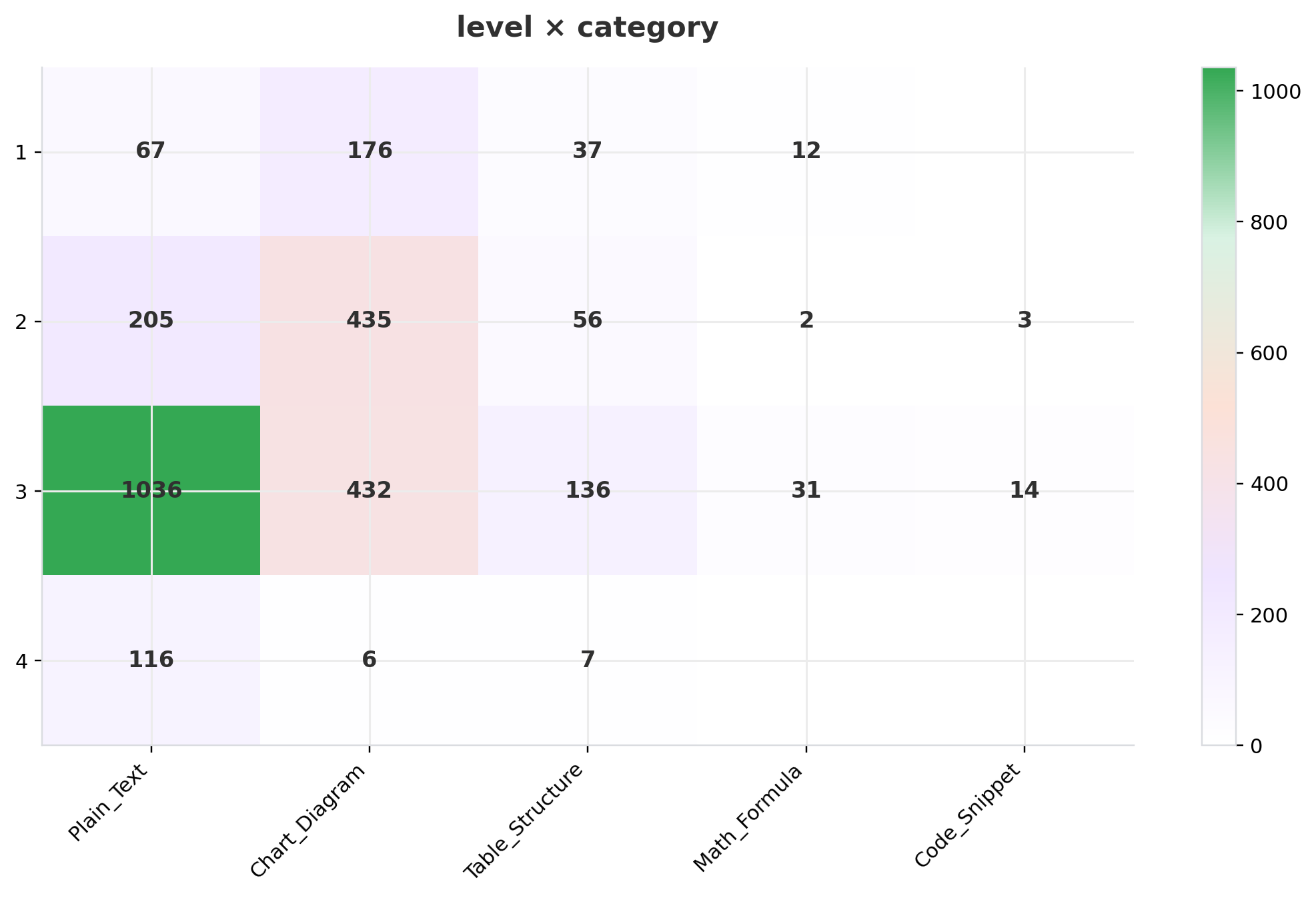}
    \caption{Heatmap of difficulty level versus semantic category.}
    \label{fig:app_level_x_category}
\end{figure}

\begin{figure}[h!]
    \centering
    \includegraphics[width=0.9\linewidth]{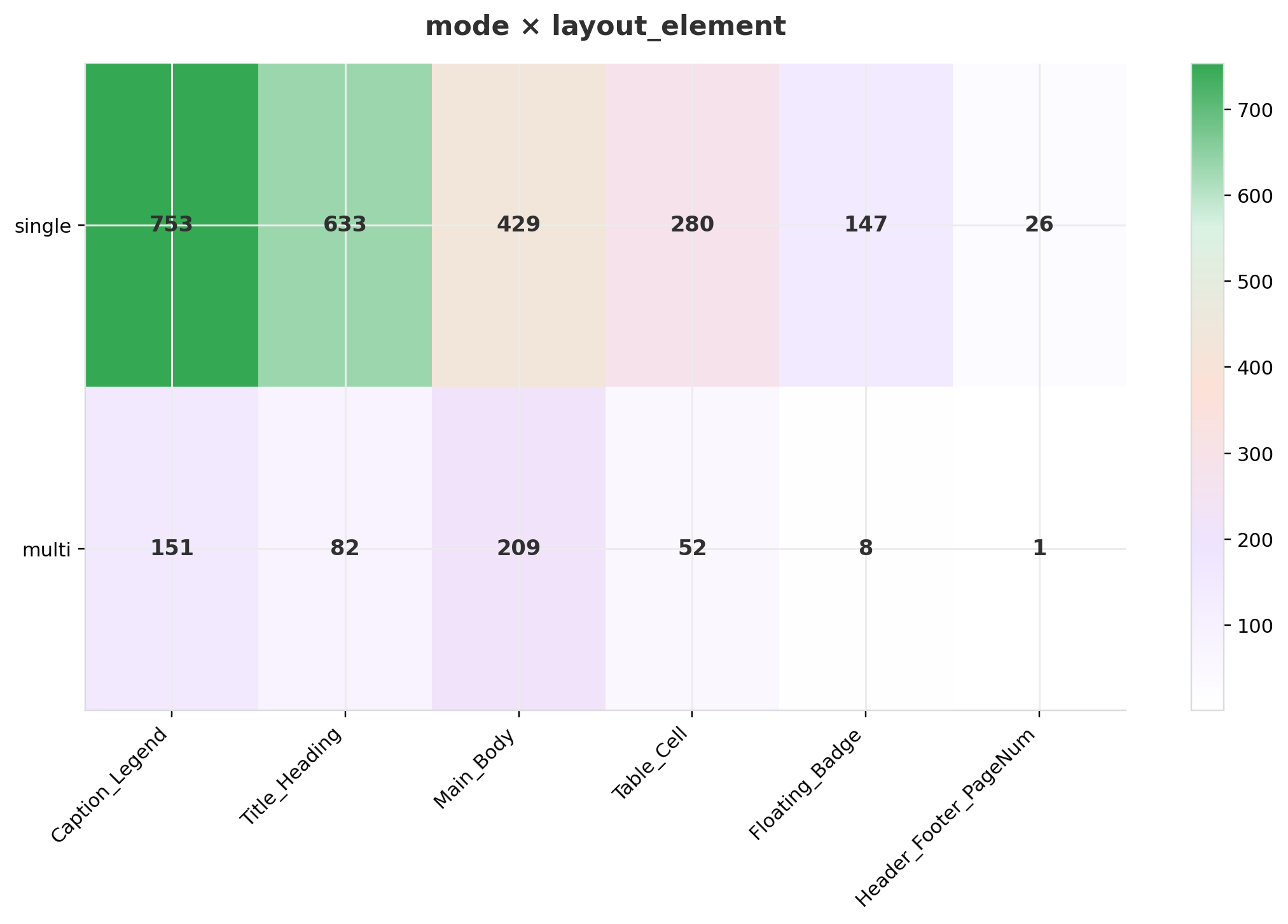}
    \caption{Heatmap of input mode versus layout element.}
    \label{fig:app_mode_x_layout_element}
\end{figure}
\newpage

\begin{figure}[h!]
    \centering
    \includegraphics[width=0.9\linewidth]{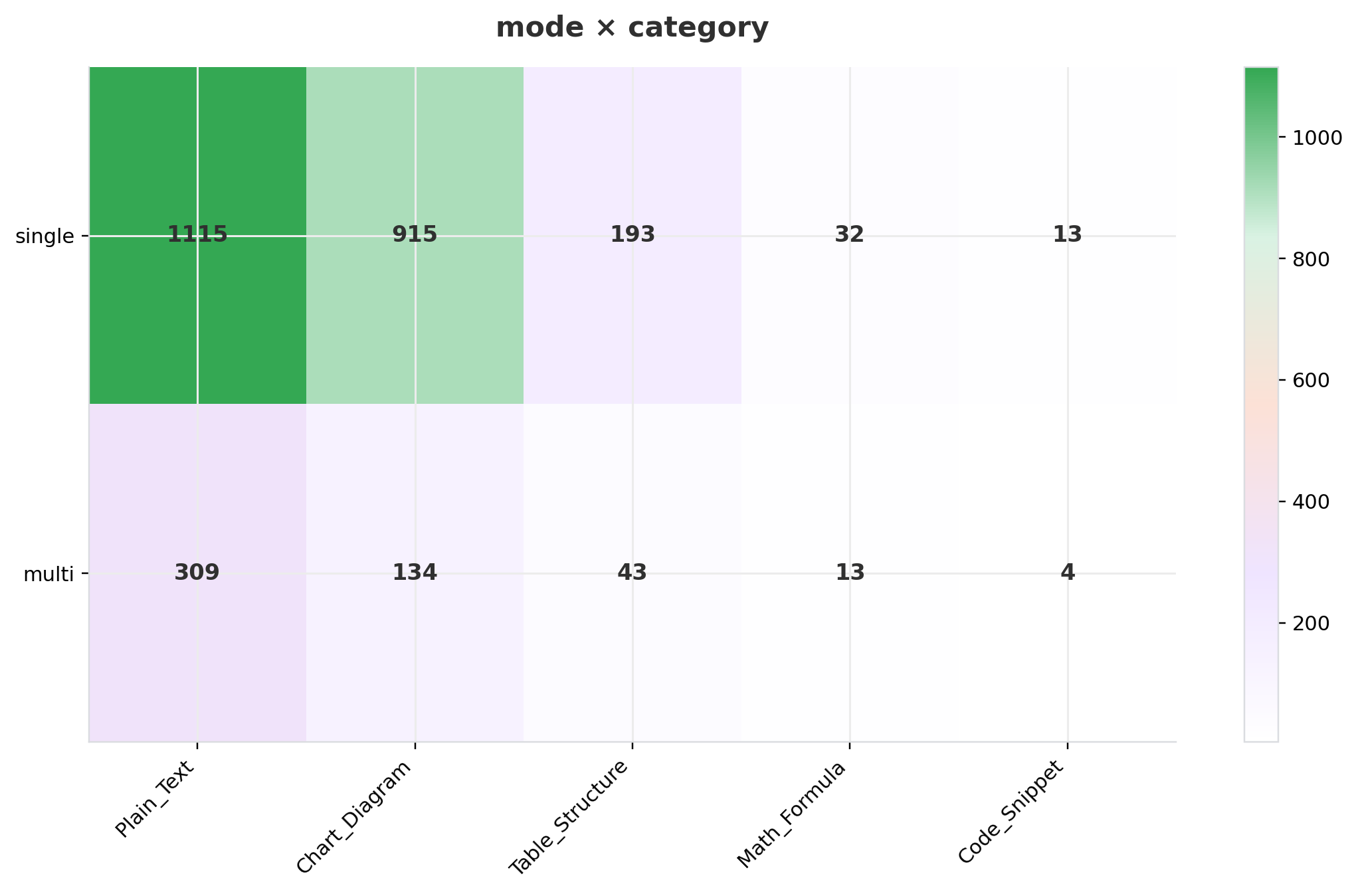}
    \caption{Heatmap of input mode versus semantic category.}
    \label{fig:app_mode_x_category}
\end{figure}

\clearpage

\section{Judge Prompt}
\label{app:judge_prompt}

For factuality gating, we use the following prompt template for the judge model.

\begin{tcolorbox}[
    colback=white,
    colframe=black!70,
    title=\textbf{\texttt{Factuality Gating Prompt (Binary Judge)}},
    arc=3mm,
    boxrule=1pt,
    fonttitle=\bfseries,
    left=3mm, right=3mm, top=3mm, bottom=3mm
]

\textbf{[Role]} \\
You are a strict, expert evaluator assessing the factual consistency of a model's prediction against a ground truth answer in a masked visual context reconstruction task. \\
\textbf{Output ONLY one character: 0 or 1. No explanation.}

\vspace{0.5em}
\textbf{[Task Overview]} \\
Span Type: \{span\_desc\} \\
Ground Truth: \{gt\} \\
Prediction: \{pred\}

\vspace{0.5em}
\textbf{[Decision Rule]} \\
Evaluate the factual alignment between the Prediction and the Ground Truth based on the specified span type.

\vspace{0.5em}
\textbf{Output 1 (Factually Consistent) IF:}
\begin{itemize}[label=--, leftmargin=1.5em, topsep=0pt]
    \item The Prediction successfully captures the core semantics and key facts of the Ground Truth.
    \item The Prediction does \textbf{not} introduce any contradictory facts, incorrect entities, wrong numbers, or opposing modifiers.
    \item \textit{Note:} Minor variations in wording, synonyms, or the presence of non-conflicting extra information are acceptable (especially for sentence and paragraph span types).
\end{itemize}

\vspace{0.5em}
\textbf{Output 0 (Factual Error) IF:}
\begin{itemize}[label=--, leftmargin=1.5em, topsep=0pt]
    \item The Prediction explicitly contradicts the Ground Truth.
    \item Essential information (e.g., specific dates, names, key modifiers like "cultural" vs. "agricultural") is missing or hallucinated.
    \item The Prediction is merely topic-related but fails to recover the specific masked meaning, or its granularity is too coarse compared to the Ground Truth.
\end{itemize}

\vspace{0.5em}
\textbf{[Output]} \\
\{0 or 1\}

\end{tcolorbox}
\newpage

\section{Additional Model Analysis}
\label{app:model_analysis}

This appendix provides additional model-side analysis that is not included in the main paper. These figures offer expanded views of score distributions, per-slice variation, and full heatmaps across benchmark slices.

\subsection{Additional score views}

\begin{figure}[h!]
    \centering
    \includegraphics[width=0.9\linewidth]{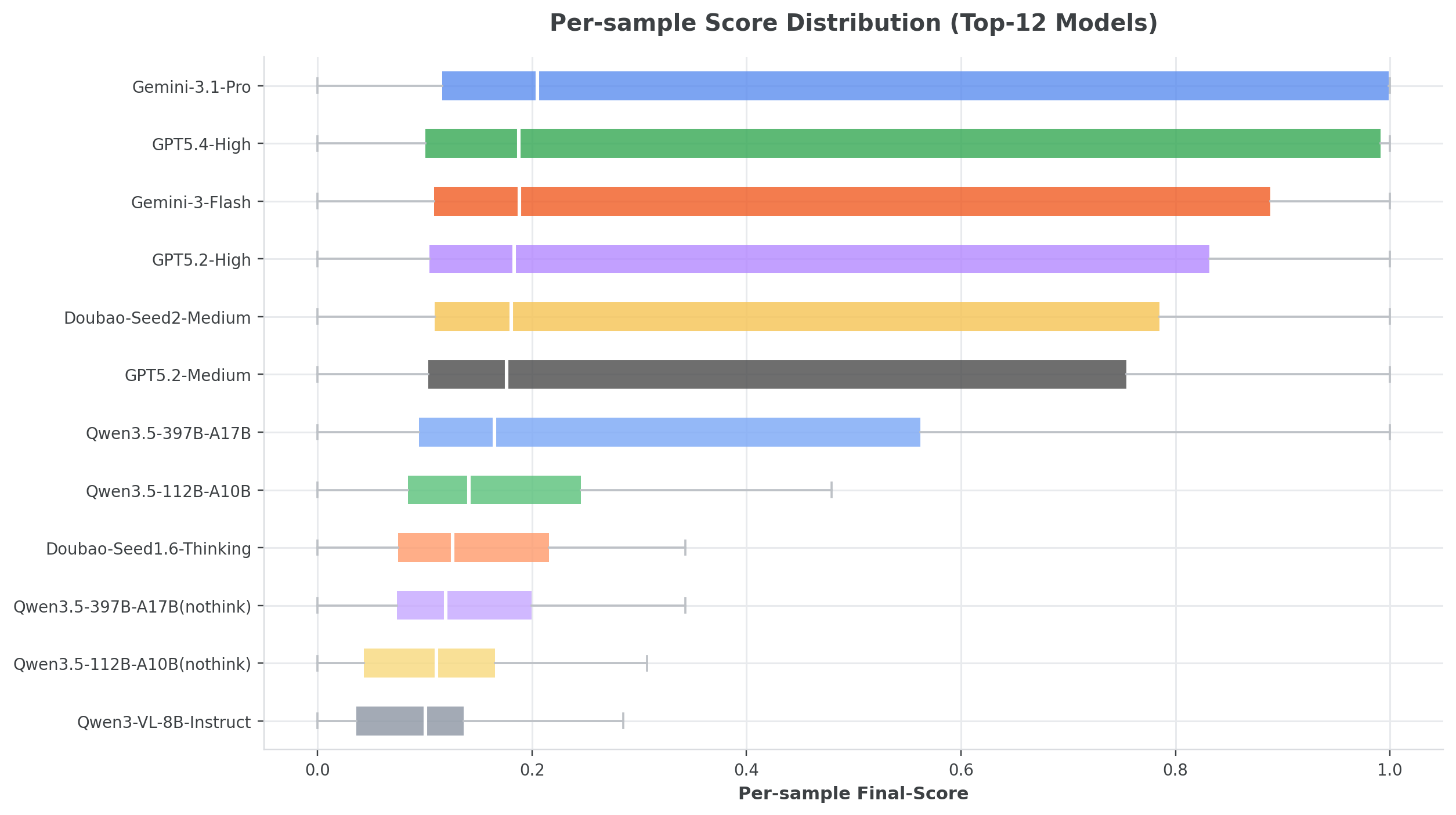}
    \caption{Per-sample score distributions for top-performing models.}
    \label{fig:app_score_distribution_top_models}
\end{figure}

\begin{figure}[h!]
    \centering
    \includegraphics[width=0.9\linewidth]{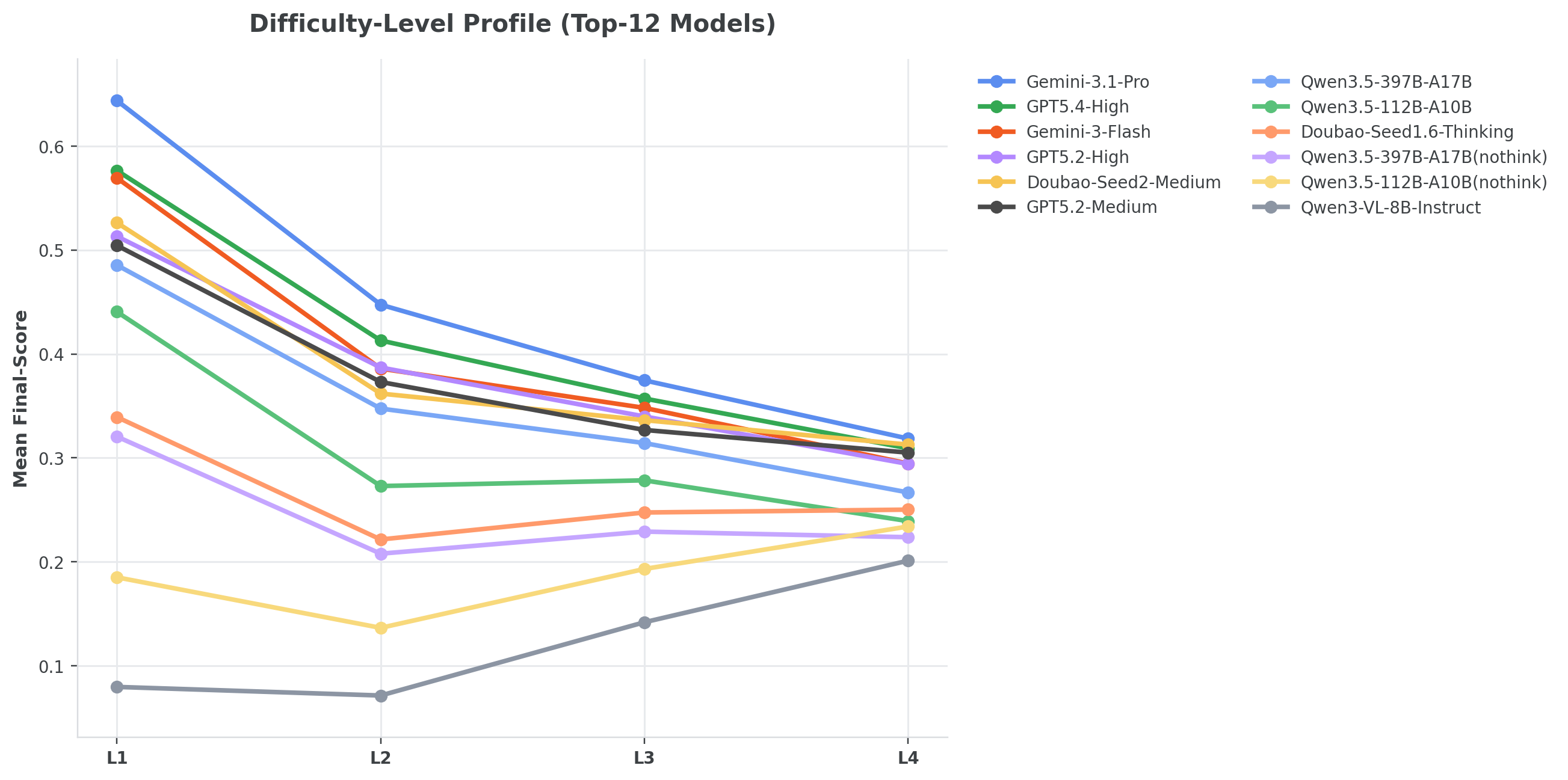}
    \caption{Difficulty-level profile for top-performing models.}
    \label{fig:app_level_profile_top10}
\end{figure}

\newpage

\begin{figure}[h!]
    \centering
    \includegraphics[width=0.9\linewidth]{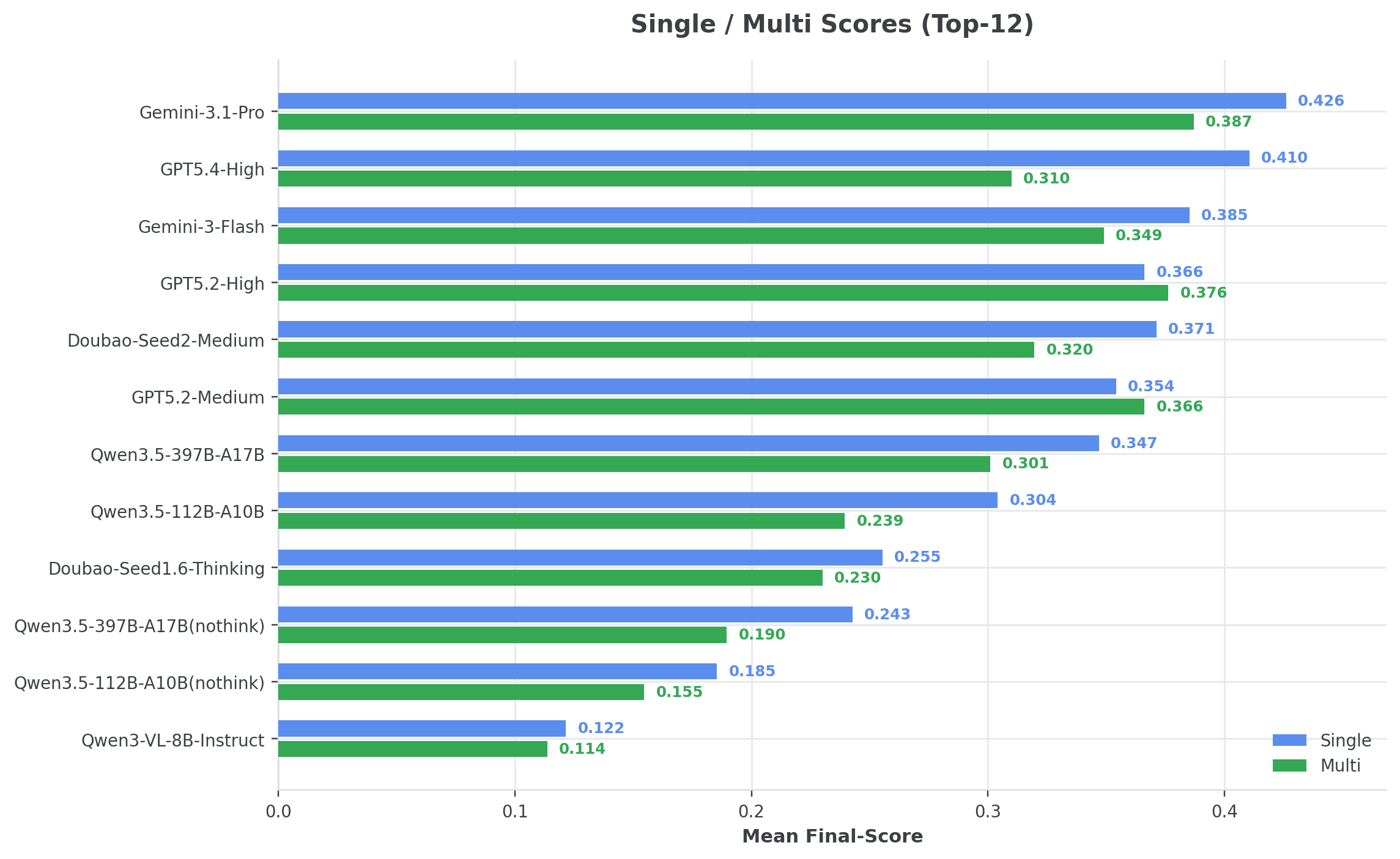}
    \caption{Performance under Single vs. Multi Context}
    \label{fig:app_single_vs_multi_top10}
\end{figure}

\begin{figure}[h!]
    \centering
    \includegraphics[width=0.9\linewidth]{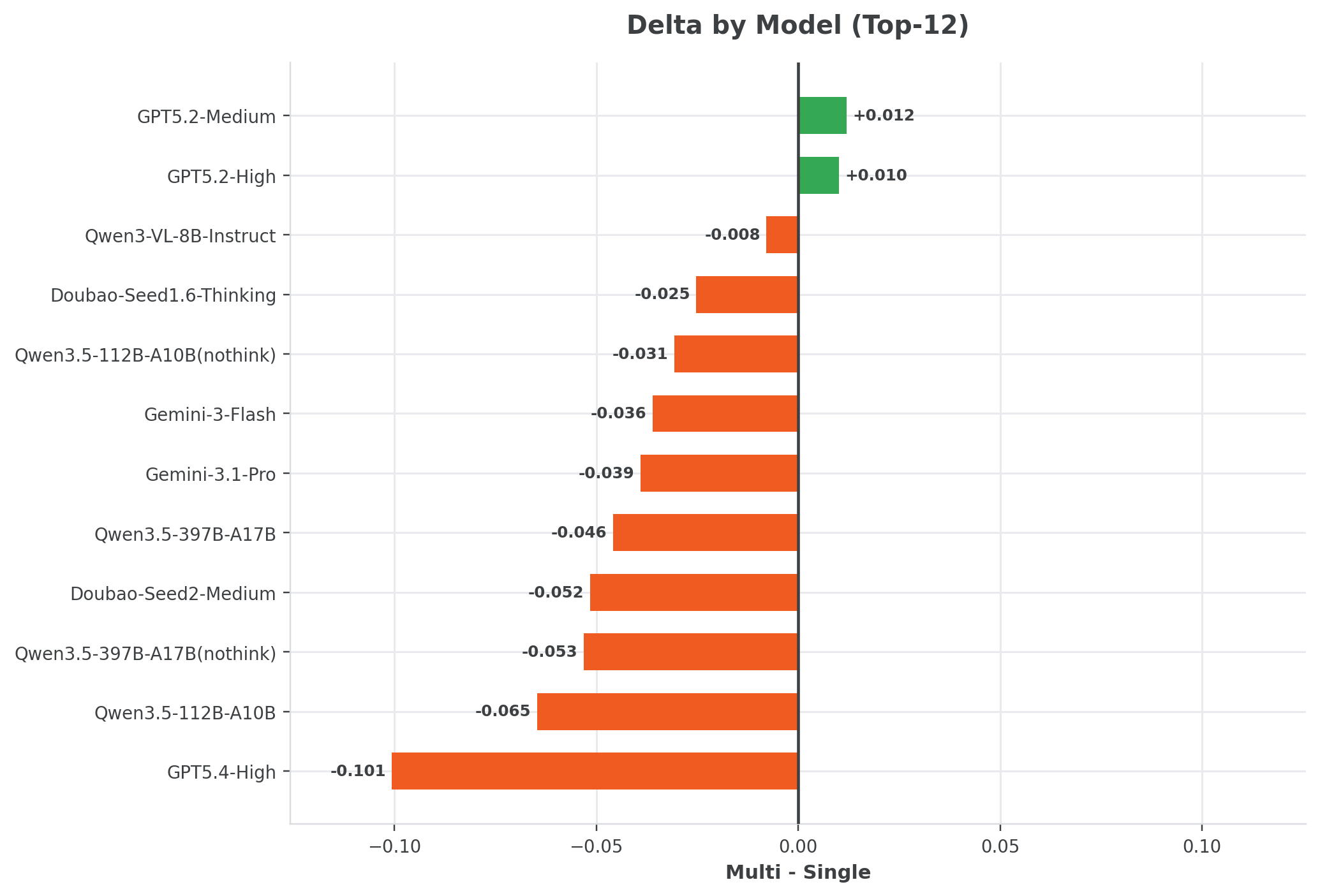}
    \caption{Performance Gain from Multi Context}
    \label{fig:app_single_vs_multi_top10_2}
\end{figure}
\clearpage

\subsection{Full semantic and structural heatmaps}

\begin{figure}[h!]
    \centering
    \includegraphics[width=0.9\linewidth]{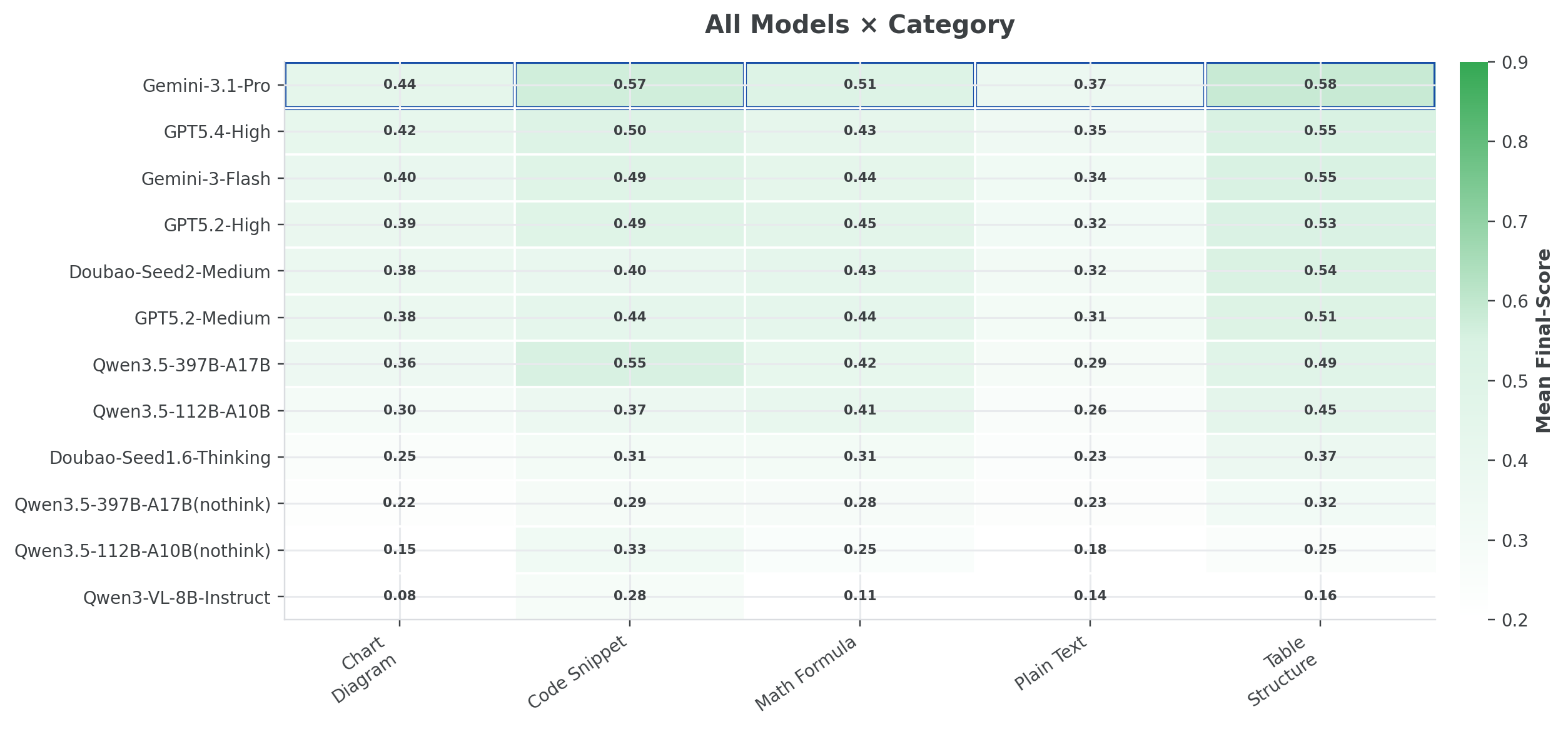}
    \caption{Full heatmap over semantic categories.}
    \label{fig:app_category_heatmap_full}
\end{figure}

\begin{figure}[h!]
    \centering
    \includegraphics[width=0.9\linewidth]{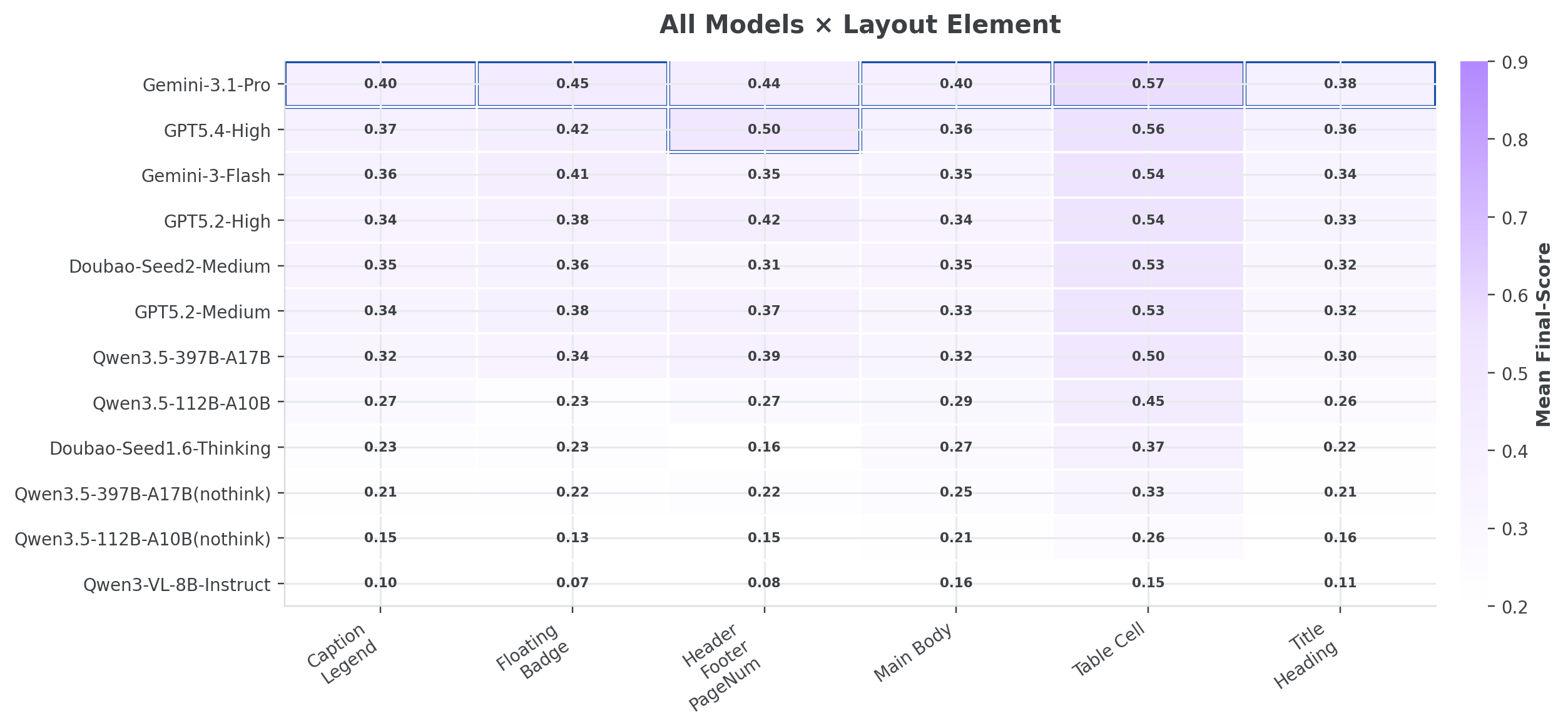}
    \caption{Full heatmap over layout elements.}
    \label{fig:app_layout_element_heatmap_full}
\end{figure}

\begin{figure}[h!]
    \centering
    \includegraphics[width=0.9\linewidth]{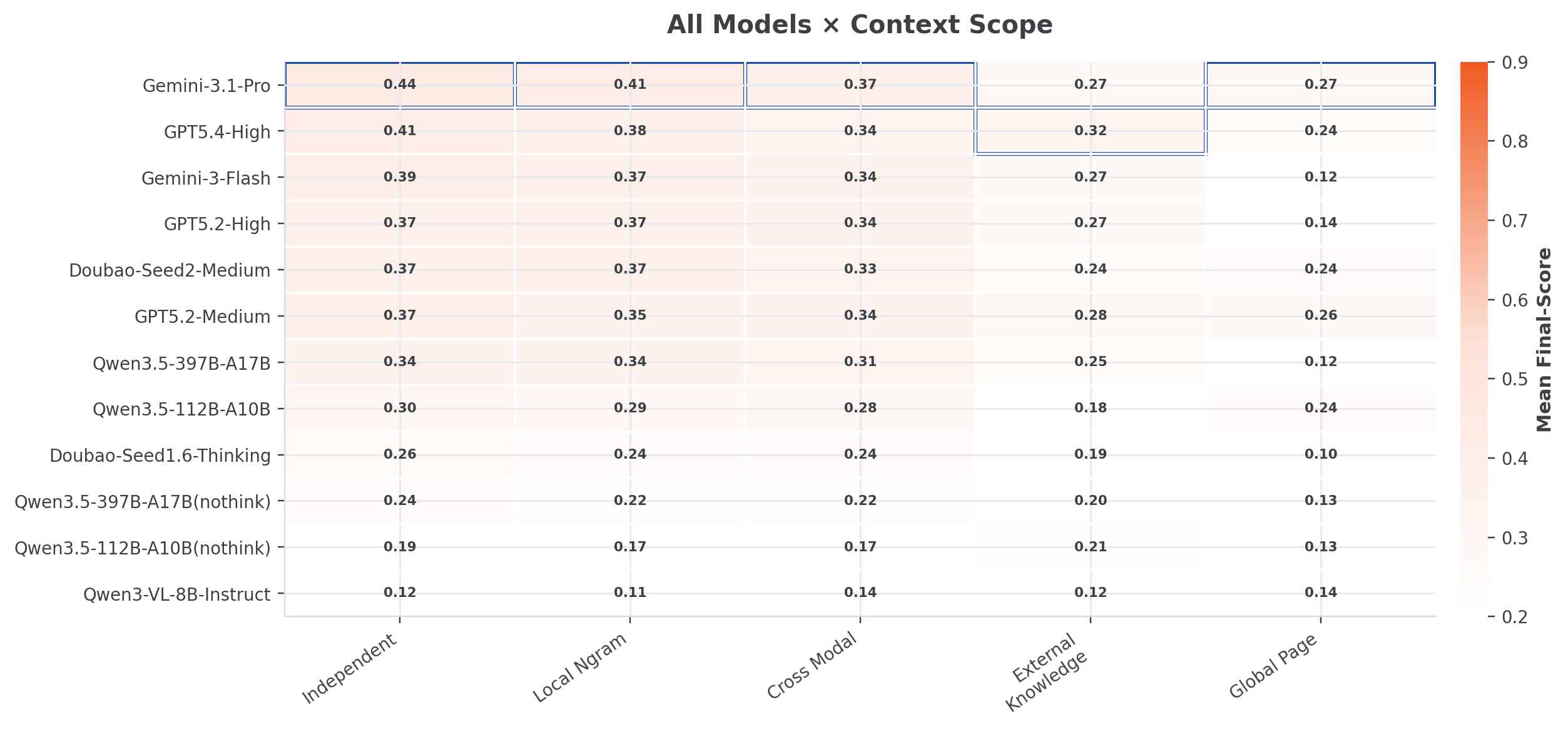}
    \caption{Full heatmap over context scope.}
    \label{fig:app_context_scope_heatmap_full}
\end{figure}

\begin{figure}[h!]
    \centering
    \includegraphics[width=0.9\linewidth]{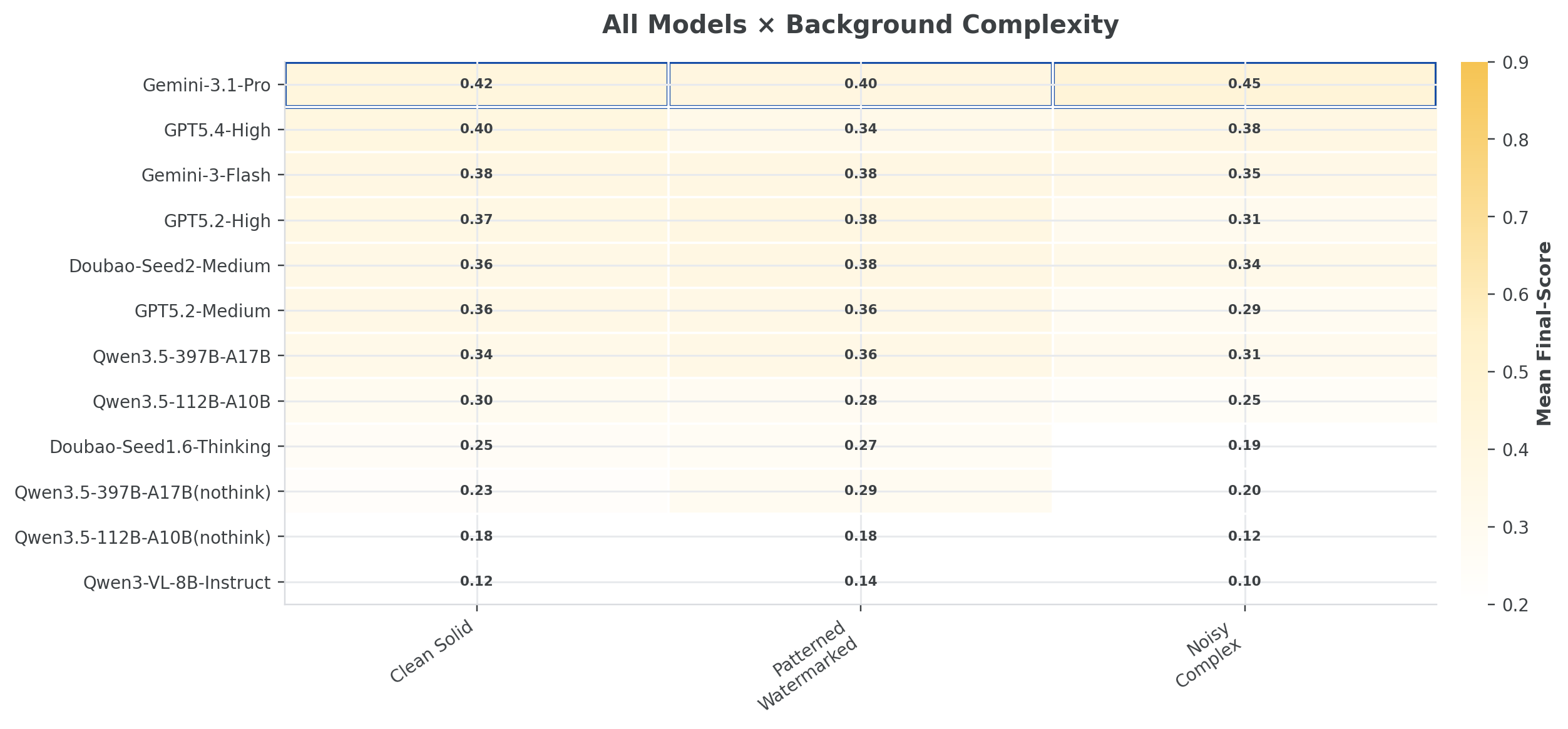}
    \caption{Full heatmap over background complexity.}
    \label{fig:app_background_complexity_heatmap_full}
\end{figure}

\begin{figure}[h!]
    \centering
    \includegraphics[width=0.9\linewidth]{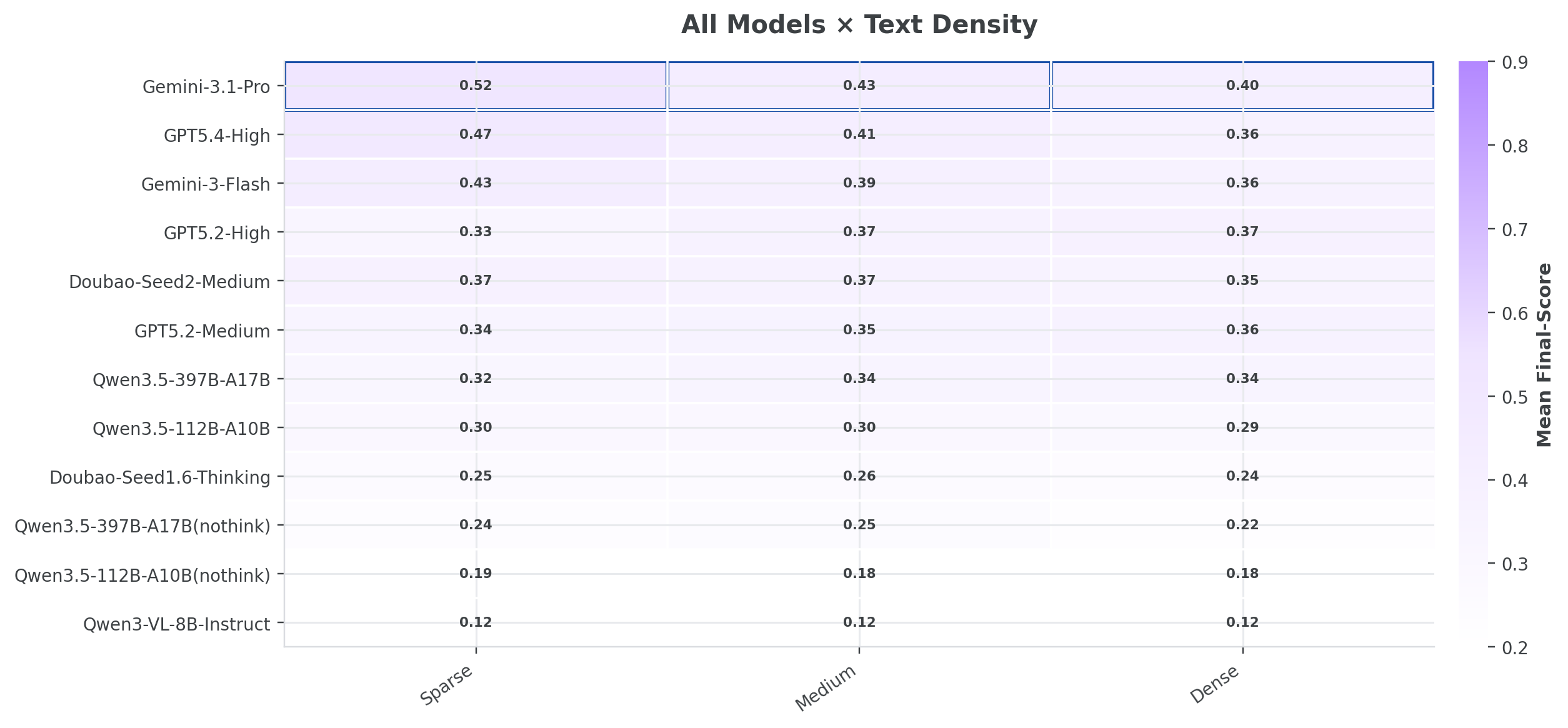}
    \caption{Full heatmap over text density.}
    \label{fig:app_text_density_heatmap_full}
\end{figure}

\begin{figure}[h!]
    \centering
    \includegraphics[width=0.9\linewidth]{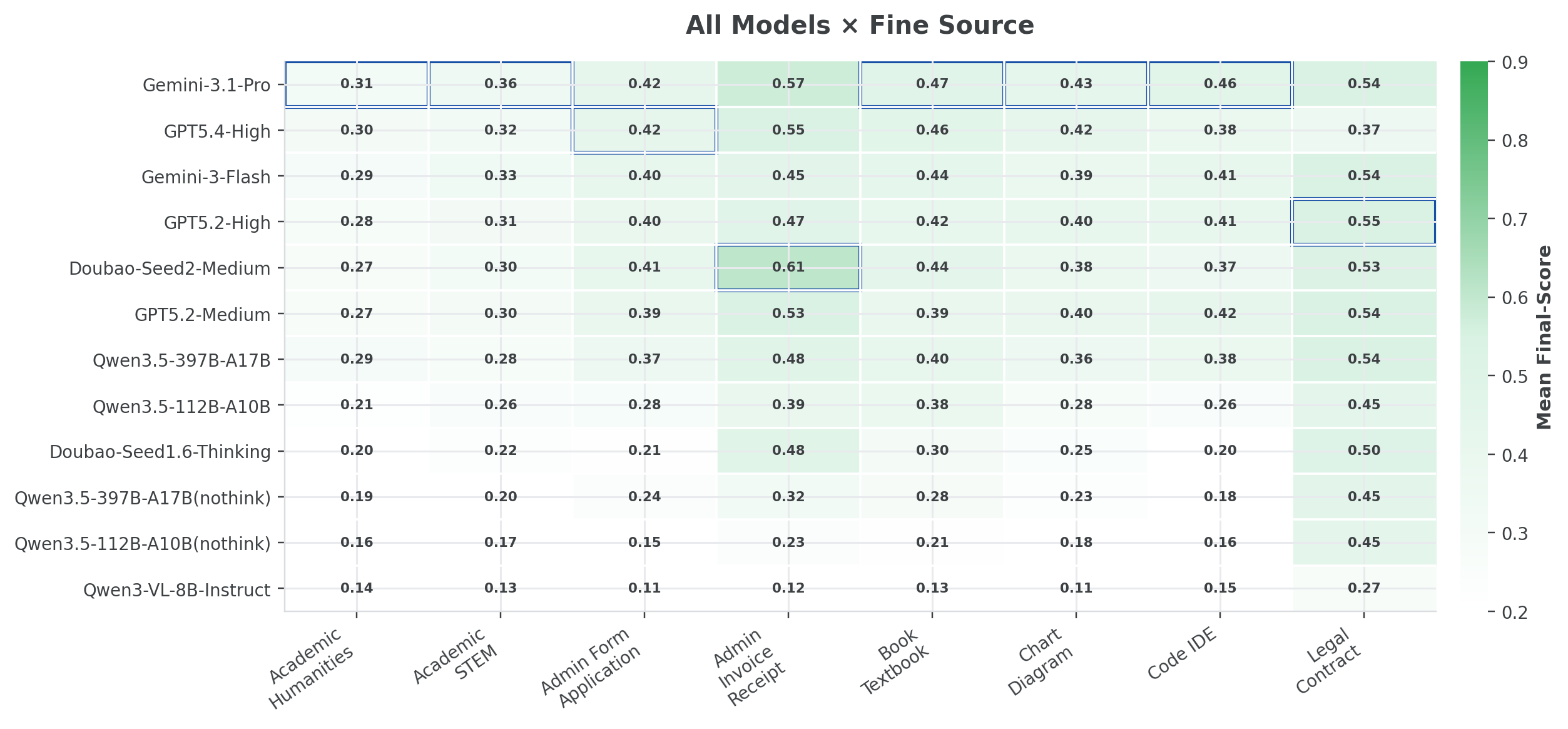}
    \caption{Full heatmap over fine-grained source types.}
    \label{fig:app_fine_source_heatmap_full}
\end{figure}

\subsection{Slice-level variation}

\newpage

\end{document}